\documentclass[10pt,twocolumn,letterpaper]{article}

\usepackage[pagenumbers]{cvpr} 

\usepackage{epsfig}
\usepackage{graphicx}
\usepackage{amsmath}
\usepackage{amssymb}
\usepackage{booktabs}
\usepackage{multirow}

\usepackage[pagebackref=true,breaklinks=true,letterpaper=true,colorlinks,citecolor=blue,linkcolor=blue,bookmarks=false]{hyperref}

\usepackage[capitalize]{cleveref}
\crefname{section}{Sec.}{Secs.}
\Crefname{section}{Section}{Sections}
\Crefname{table}{Table}{Tables}
\crefname{table}{Tab.}{Tabs.}

\def\cvprPaperID{2801} 
\def\confName{CVPR}
\def\confYear{2022}

\begin{document}

\title{Single UHD Image Dehazing via Interpretable Pyramid Network}

\author{Boxue Xiao$^{1}$, Zhuoran Zheng$^{2}$,Xiang Chen$^{2}$, Chen Lv$^{3\ast}$, Yunliang Zhuang$^{3}$, Tao Wang$^{4}$
	\\
	$^{1}$SPE,Shandong Normal University
	$^{2}$CSE,Nanjing University of Science and Technology \\
	$^{3}$SISE,Shandong Normal University 
	$^{4}$Huawei Noah's Ark Lab \\
}
\maketitle

\begin{abstract}
	Currently, most single image dehazing models cannot run an ultra-high-resolution (UHD) image with a single GPU shader in real-time. To address the problem, we introduce the principle of infinite approximation of Taylor's theorem with the Laplace pyramid pattern to build a model which is capable of handling 4K hazy images in real-time. The N branch networks of the pyramid network correspond to the N constraint terms in Taylor's theorem. Low-order polynomials reconstruct the low-frequency information of the image (e.g. color, illumination). High-order polynomials regress the high-frequency information of the image (e.g. texture). In addition, we propose a Tucker reconstruction-based regularization term that acts on each branch network of the pyramid model. It further constrains the generation of anomalous signals in the feature space. Extensive experimental results demonstrate that our approach can not only run 4K images with haze in real-time on a single GPU (80FPS) but also has unparalleled interpretability. 
	The developed method achieves state-of-the-art (SOTA) performance on two benchmarks (O/I-HAZE) and our updated 4KID dataset while providing the reliable groundwork for subsequent optimization schemes.
\end{abstract}

\begin{figure}[t] 
	\begin{center}
		\begin{tabular}{@{}c@{}}
			\includegraphics[width = 0.475\textwidth]{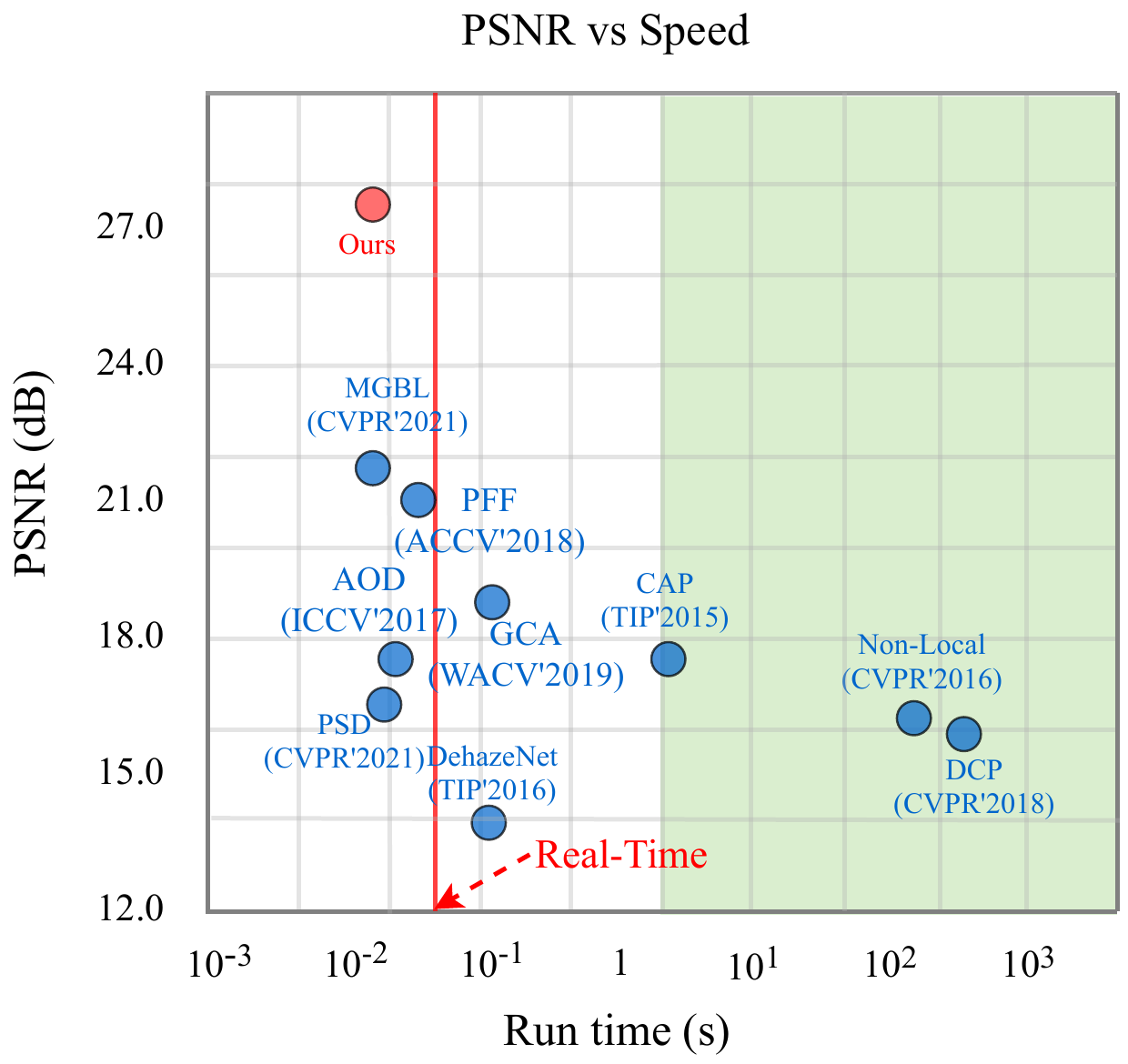}                   
		\end{tabular}
	\end{center}
	\vspace{-4mm}
	\caption{Our method has SOTA performance in the process of comparing with other models on the 4KID datasets. The methods in the green area take more than 1s to run an image. The red line in the figure indicates that the model processes images at a speed of 30fps.}
	\vspace{-5mm}
	\label{ms-psnr}
\end{figure}

\section{Introduction}
The image dehazing tasks are essential to boost the capability of target detection in harsh.
Physical model-based approaches usually require a manual definition of the hyperparameters of the model, which entails expensive labor costs.

Recently, numerous end-to-end CNN-based methods~\cite{qin2020ffa,zheng2021ultra,wu2021contrastive,li2017all,he2010single,qu2019enhanced} have been proposed to simplify the de-hazing problem by directly learning hazy-to-clear image translation via a dehazing network. Although these methods show great performance on the public datasets, some inevitable problems are as follows:

1) The physical prior-based dehazing algorithms \cite{he2010single,zhu2015fast,berman2016non} need a long time and their performance is poor when running the high-resolution hazy images. For example, the DCP \cite{he2010single}, non-local \cite{berman2016non} all require more than 100s to process a single 4K image (see Figure \ref{ms-psnr}).

2) The deep learning-based methods \cite{qin2020ffa,wu2021contrastive,he2010single,zheng2021ultra,cai2016dehazenet} adopt several convolutional operations at the cost of computing resources to obtain high performance of the model. For instance, FFA-Net \cite{qin2020ffa} has achieved remarkable performance in dehazing tasks. Unfortunately, it takes more than 0.1s and cannot be trained at scale using full-resolution size images on 24G RAM.

3) The recent approaches \cite{zheng2021ultra,chen2021psd} have made some effective attempts at the high-resolution image dehazing task, but they still lack sufficient interpretability and can only prove their validity by limited qualitative analysis of the modules.

To address the above problems, we introduce the principle of infinite approximation of Taylor's theorem with the Laplace pyramid pattern to build a model which can handle 4K hazy images in real-time.
Taylor's theorem considers that an ordinary polynomial can be composed of low-order polynomials and higher-order polynomials expanded around a point (a pixel). 
Importantly, it is beneficial for computers to deal with high-dimensional data. Eli Schwartz et al. \cite{schwartz2018deepisp} took Taylor's theorem as the core for image noise reduction and achieved good results.
This shows that it is reasonable to integrate Taylor's theorem into image processing. 
In addition, our developed model learns the attention sharing tensor $K$ which can make the network focus on high-order Taylor terms via a low-rank U-Net. 
The $K$ acts on the high-resolution image by interpolation, which avoids the overhead associated with stacking convolution kernels on high-resolution branches to extract higher-order features.
In addition, we propose a Tucker reconstruction-based regularization term which can further constrain the abnormal signal.
As shown in Figure \ref{ms-psnr}, the PSNR of our method can reach 27.79 dB on the 4KID dataset, and the running time of an image of 3840 $\times$ 2160 resolution is only 14 ms.
Our main contributions are summarized as follows:
\begin{itemize}
	\vspace{-2mm}
	\item  We introduce the infinite approximation of Taylor's theorem with the Laplace pyramid structure. This approach not only has extreme performance (run a 4K image in real-time) but also has strong interpretability.
	\vspace{-2mm}
	\item  We propose a Tucker reconstruction-based regularization term that further constrains the generation of anomalous signals in the feature space.
	\vspace{-2mm}
	\item
	We propose the attention sharing tensor $K$ which not only focus on realistic texture details but also reduces the computation time of high-resolution images.

\end{itemize}

\begin{figure*}[h]
	\centering
	\includegraphics[width=1\textwidth]{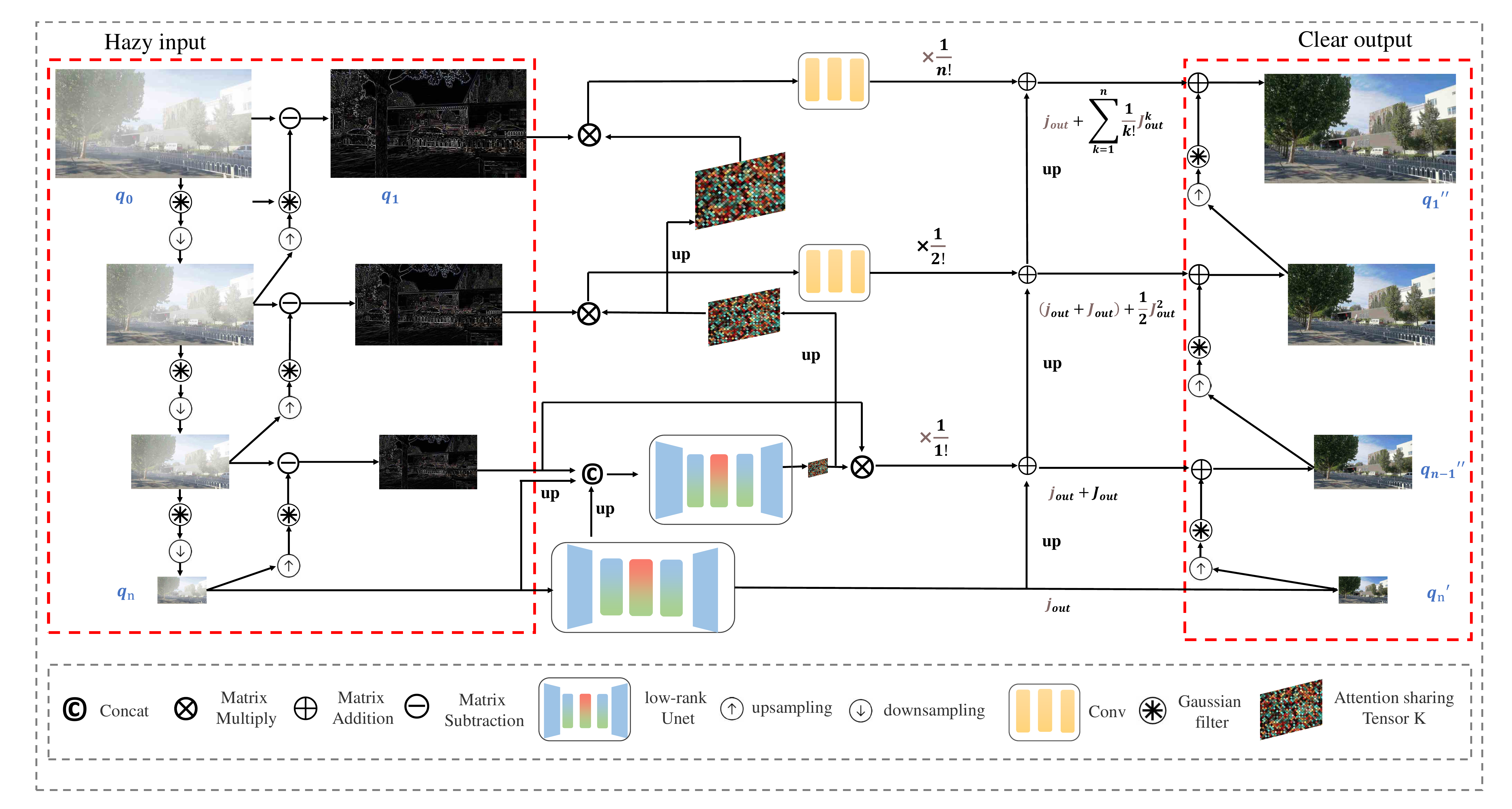}
	\vspace{-5mm}
	\caption{The framework of LapDehazeNet. 
	The red boxes are the input and output streams of the pyramid.
	The pyramid represents the first to the nth term of Taylor's theorem from the bottom to the top. 
	Each level except the bottom of the pyramid is multiplied by the attention sharing tensor to fit the derivative in Taylor's theorem. 
	The output of each level of the pyramid is an approximate Taylor term. 
	We use bilinear interpolation to up-sample the feature maps to generate a clear UHD image by the pyramid.}
	\label{network}
\end{figure*}
\section{Related work}

\subsection{Single Image Dehazing}
\noindent 
\textbf{Prior-based image dehazing methods.}
Prior-based image dehazing methods usually reconstruct the images by estimating atmospheric scattered light and transmittance such as dark channel prior (DCP)~\cite{he2010single}, color attenuation prior, CAP~\cite{zhu2015fast} and non-local prior~\cite{berman2016non}. These methods have made great progress in different degrees. However, their performance is limited by the accuracy of assumptions used in complex scenarios. In addition,  the approaches need a lot of reasoning time to run a 4K image.

\noindent 
\textbf{Learning-based image dehazing methods.} Numerous end-to-end methods~\cite{qin2020ffa,zheng2021ultra,wu2021contrastive,li2017all,qu2019enhanced,berman2016non} replace the atmospheric scattering model \cite{nayar1999vision} by learning the mapping relationship between the clear image and the hazy image. For instance, Xu et al. \cite{qin2020ffa} propose a method that combines a feature attention module with a pixel attention mechanism. It expands CNN's ability to express itself well. Zheng et al.\cite{zheng2021ultra} propose a dehazing network with three CNNs. The CNN's extract haze-relevant features and learn multiple full-resolution guidance maps corresponding to the leaned bilateral model. 
However, the above methods are difficult to establish theoretical guidance for effective optimization models.
\subsection{Taylor Approximation Theorem}
Taylor approximation is one of the most popular methods in numerical approximation. 
The previous methods~\cite{leung2004general,kobayashi2011analytical,helton2003latin,lai2017deep} have been made and based on the first-order Taylor’s expansion. 
Xue et al. \cite{xue2015high} develop the above methods by high-order Taylor series expansion, which further improve the performance. 
Recently, Zhou et al. \cite{zhou2021unfolding} use the Taylor theorem to construct an image restoration framework. 
It learns the high-level contextualized information via the mapping function. 
The derivative function combined with the degraded input restore local high-order spatial details gradually. 
Further, we propose to fuse the infinite approximation of Taylor's theorem with Laplace's pyramid. The $N$ branch networks of the pyramid network correspond to the $N$ terms in Taylor's theorem.
\subsection{Laplacian Pyramid}
Pyramid structures have been widely used~\cite{huang2021fapn,hu2021pyramid,liang2021high,lin2021drafting,he2019foreground,dong2020spatial} because of great multi-scale fusion characteristics. 
Laplacian pyramid \cite{liang2021high,lin2021drafting} has a long history in digital image processing. For instance, LapStyle first transfers global style patterns in low-resolution via a Drafting Network \cite{lin2021drafting}. It then revises the local details in high-resolution via a Revision Network, which hallucinates a residual image. 
Higher-resolution details can be easily generated by stacking Revision Networks with multiple Laplacian pyramid levels. 
Liang et al.~\cite{liang2021high} came up with a Laplacian translation network, it refines the high-frequency components effectively via the gradual masking strategy.
Inspired by these methods, 
we use the structure of the pyramid to combine with the use of Taylor's theorem for achieving UHD image dehazing.
\subsection{Processing Methods of UHD Images}
To transfer our laboratory models to real application scenarios effectively, UHD image research\cite{zhang2021multi,he2021inferring,verelst2021blockcopy,song2021starenhancer,wang2021real,wang2019underexposed,yuan2021hrformer,wu2021contrastive} which can be processed in real-time is becoming popular. To be able to process UHD images in real-time and get good results, Zhang et al.~\cite{zhang2021multi} propose a new Vision Transformer (ViT) architecture Multi-Scale Vision Longformer which provides image encodings with manageable computational cost. In addition,
Thomas Verelst et al. propose~\cite{verelst2021blockcopy} BlockCopy operating on the selected region and reduce the number of calculations. 


\section{Proposed Method}
We propose an end-to-end Laplacian pyramid dehazing network based on Taylor's infinite approximation theorem (LapDehazeNet). As shown in Figure \ref{network}, each level of the pyramid output represents a Taylor term.
%
%
In addition, we use Tucker reconstruction to constrain spatial abnormal signals on U-Net~\cite{ronneberger2015u} and use attention sharing tensor $K$ to fit the differential operation in each Taylor term.

\begin{figure*}[!hpt]\scriptsize 
	\begin{center}
		\tabcolsep 1pt
		\begin{tabular}{@{}cccc@{}}
			
			\includegraphics[width = 0.25\textwidth]{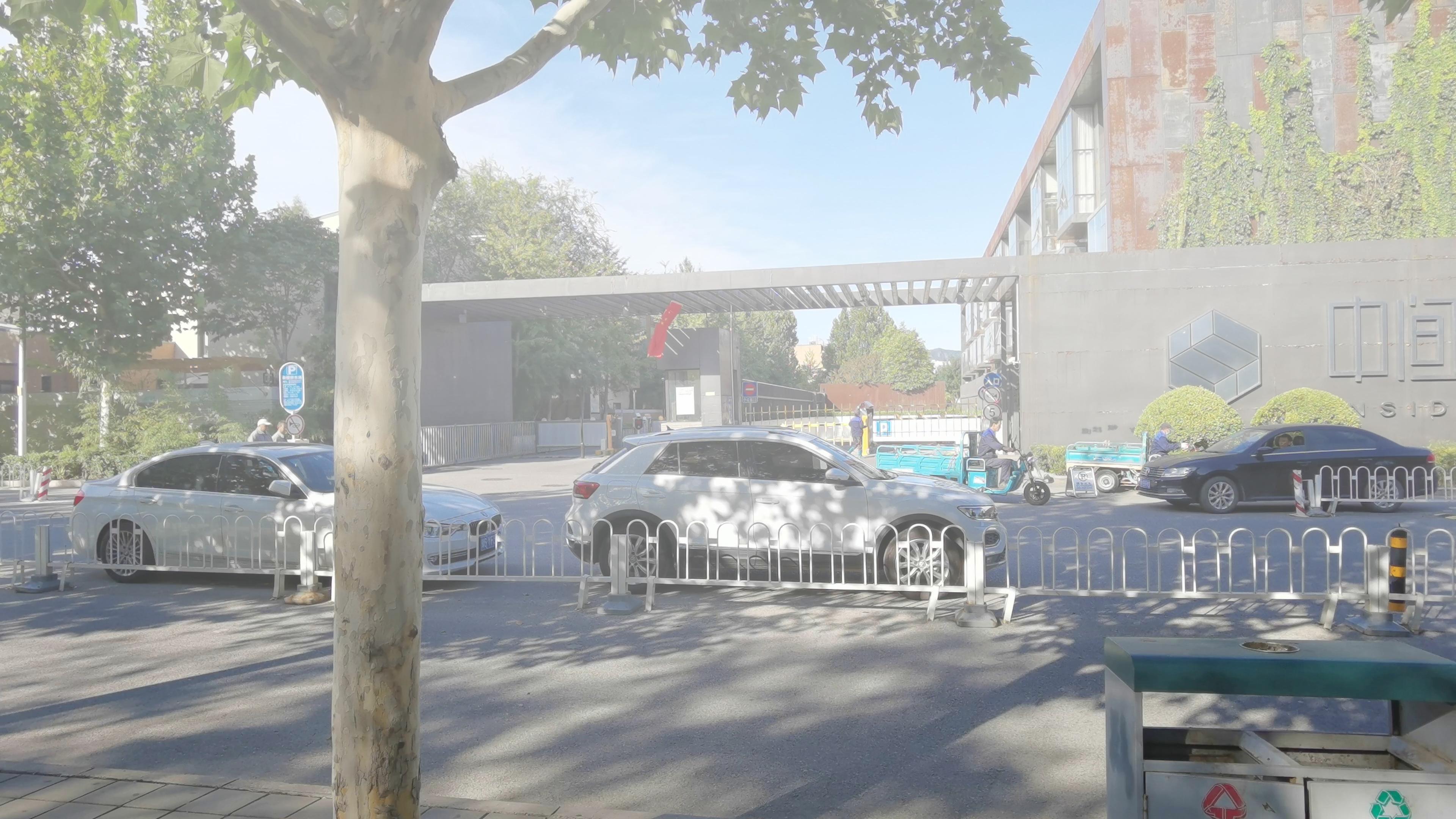}                &
			\includegraphics[width = 0.25\textwidth]{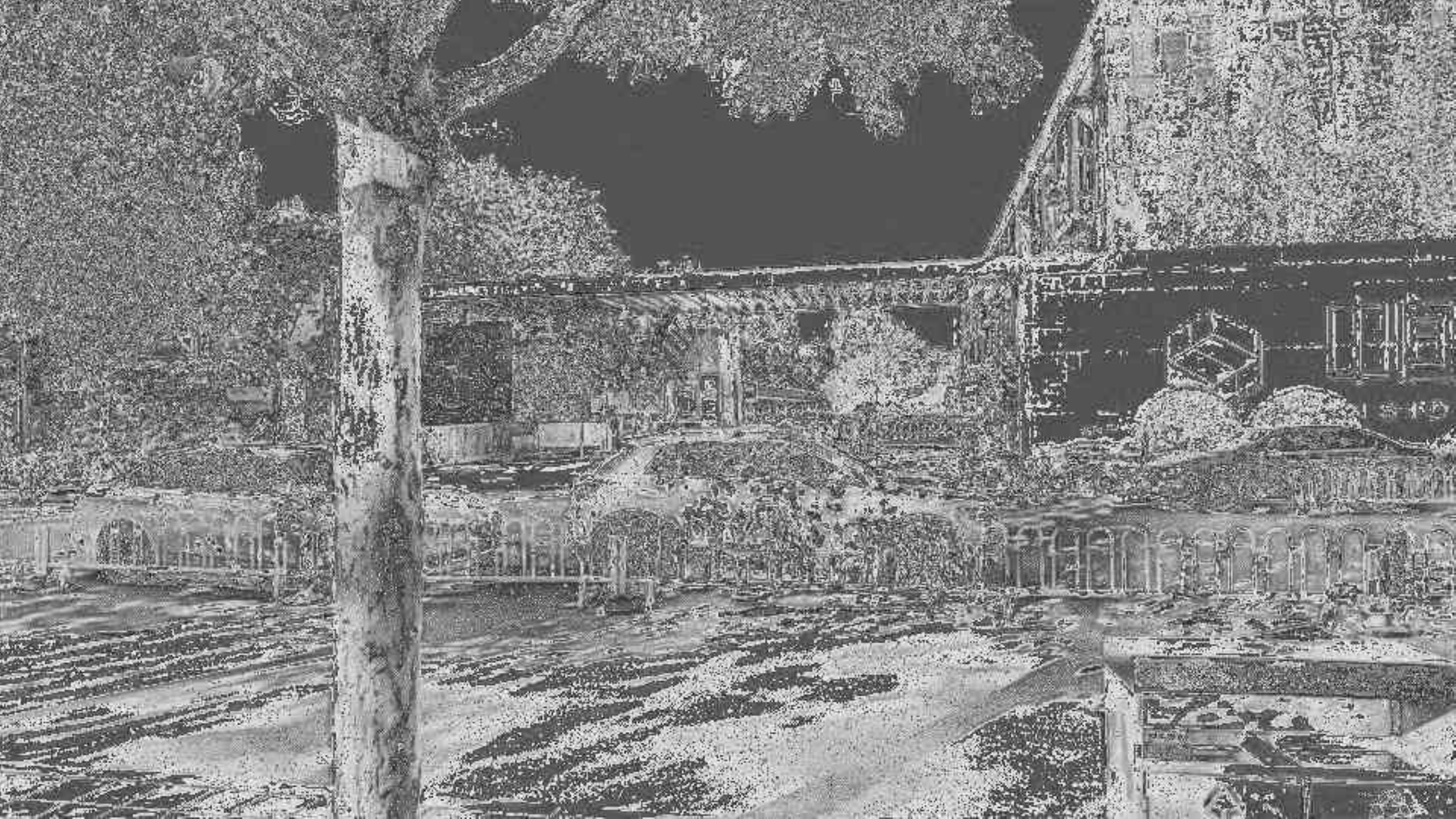}                 &
			\includegraphics[width = 0.25\textwidth]{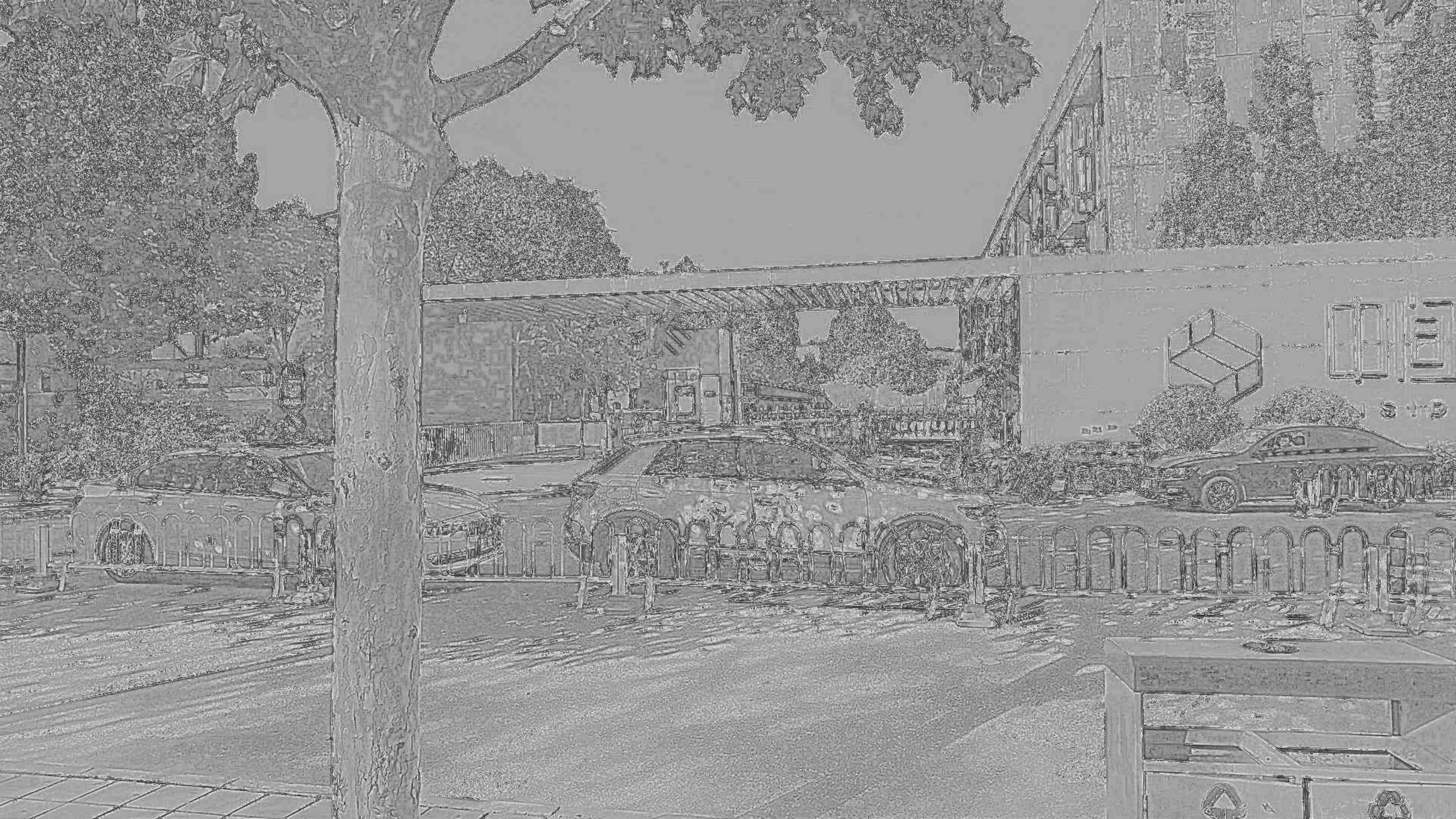}                  &
			\includegraphics[width = 0.25\textwidth]{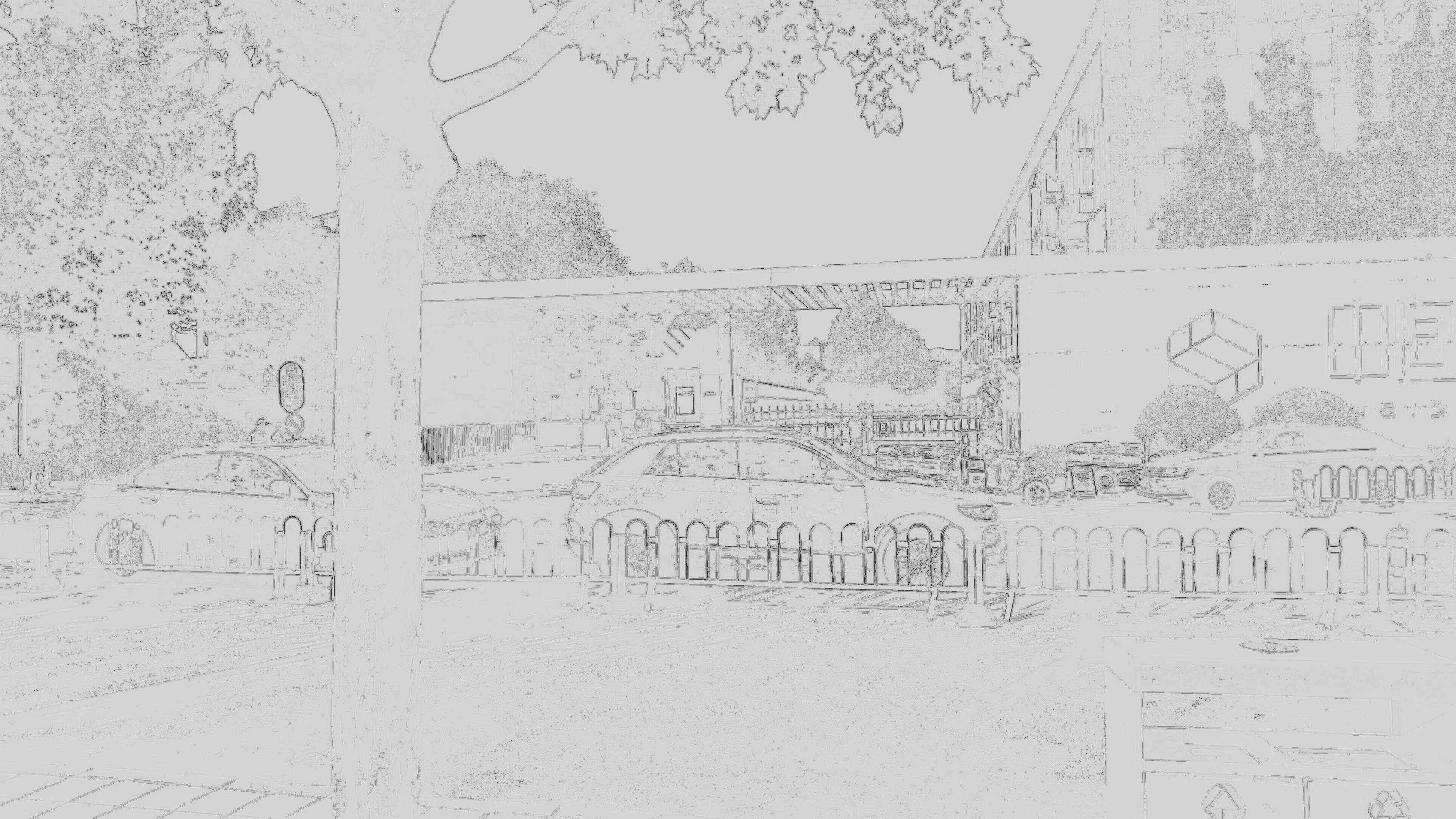}                  \\
			(a) Input                   &
			(b) Second Taylor term & 
			(c) Third Taylor term        & 
			(d) Fourth Taylor term                     \\
			
		\end{tabular}
	\end{center} 
	\caption{Visual grayscale of each Taylor term in LapDehazeNet. Low-order Taylor terms express illumination with fewer texture details, while high-order Taylor terms express texture with more texture details. } 
	\label{Lap}
\end{figure*}

\subsection{Taylor Theorem for Image Dehazing}
The UHD image dehazing can be regarded as the map from hazy images to clear images. 
We denote this map by using the function $F$:
\begin{eqnarray}
	y=F(x),
\end{eqnarray}
where $x$, $y$ denote a hazy image, a clear image. 
If the function $F(x)$ has n+1 order derivatives in the neighborhood of $x_0$, we can use an infinite Taylor series expansion to expand the above equation in the neighborhood:
\begin{small}	
	\begin{eqnarray}
		F(x)=F(x_0)+\sum_{k=1}^{\infty}\frac{1}{k!}\frac{\partial{f^k(x_0)}}{\partial{x^k}}(x-x_0)^k+R_n(x),
	\end{eqnarray}
\end{small}
\noindent
where the $R_n(x)$ is Lagrange remainder, the $x_0$ is any point. When only considering the N-order Taylor approximation, it can be simplified as:
\begin{eqnarray}
	F(x)=F(x_0)+\sum_{k=1}^{\infty}\frac{1}{k!}\frac{\partial{f^k(x_0)}}{\partial{x^k}}(x-x_0)^k.
\end{eqnarray}

The low-order (also low frequency) features reflect the primary features of color, illumination, and so on while high-order (also high frequency) features the edge information of the image.
Therefore, we use the low-order component $F(q_n)$ of the mapping function $F(x)$ to reflect the transformation process of color and illumination in the process of dehazing. $q_n$ is obtained by continuous downward sampling after Gaussian filtering. Most importantly, the resolution of $q_n$ is the lowest, which means we need to do less computation. 
This makes it easier for us to process UHD images, due to the complex operations acting on the low-resolution images.
As shown in Figure \ref{Lap}, the remaining high-order components are the high-frequency details (high-resolution image) continuously supplemented after the image is dehazed.
We unfold the mapping function $F(x)$ around the $q_n$ obtained by Laplace decomposition. it can be reformulated as:
\begin{eqnarray}
	F(x)=F(q_n)+\sum_{k=1}^{\infty}\frac{1}{k!}\frac{\partial{f^k(q_n)}}{\partial{x^k}}(x-q_n)^k.
\end{eqnarray}

We mark the $k$th term of the high-order Taylor terms as $J_{out}^k$, and make $F(q_n)=j_{out}$. 
We finally get the mapping function from hazy images to haze-free images.
\begin{small}
	\begin{eqnarray}
		F(x)=j_{out}+\sum_{k=1}^n\frac{1}{k!}J_{out}^k.
	\end{eqnarray}
\end{small}
	\vspace{-1mm}
In this paper, we use each level of the pyramid to represent each Taylor term.

\subsection{Interpretable Laplacian Pyramid}
Laplacian pyramid \cite{burt1987laplacian} can decompose an image into a set of high- and low-frequency bands, from which the original image can be exactly reconstructed. For an image $q_0$ of $h\times{w}$ pixels, we have $q_1 = q_0-\hat{q_0}$, where $q_1$ is the high-frequency residual, $\hat{q_0}$ is obtained by downsampling and upsampling steps of $q_0$.
By recursively downsampling the original signal $q_0$, $q_0\in{R^{h\times{w}\times{3}}}$ and performing the above operations, the high frequency component [$q_1$, $q_2$,..., $q_{n-1}$], $q_{n-1}\in     {R^{{\frac{h}{2^{L-1}}}\times{\frac{w}{2^{L-1}}\times3}}}$ and low frequency component $q_n\in{{R^{{\frac{h}{2^{L}}}\times{\frac{w}{2^{L}}\times3}}}}$ can be obtained through Laplace linear decomposition, where $L$ is the number of pyramid layers with high frequency components. ($L=n-1$).

Additionally, in the process of image reconstruction using a series of processed high and low frequency image signals, we also have $q_{n-1}''=q_n'+q_{n-1}'$, where $q_n'\in{R^{h\times{w}\times{3}}}$ is obtained by processing and upsampling steps of $q_n$, $q_{n-1}'$ is obtained by processing $q_{n-1}$, $q_{n-1}''$ is an image during reconstruction. We perform recursive operation on $q_{n-1}''$ and finally get the reconstructed image $q_1''$.

\begin{figure}[h]
	\centering
	\includegraphics[width=0.94\linewidth]{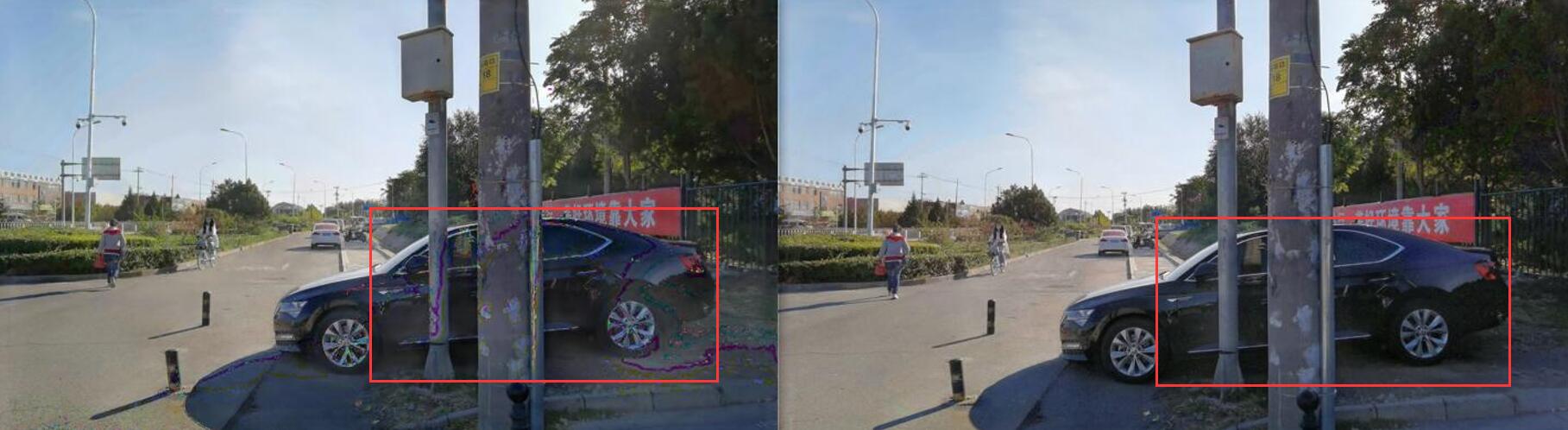}
	\includegraphics[width=0.94\linewidth]{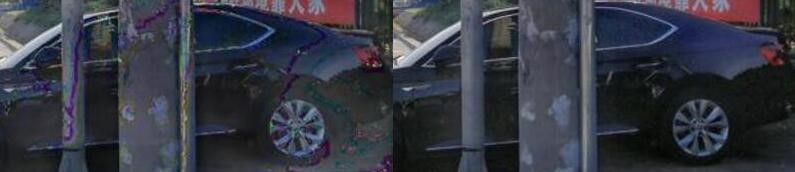}
	\caption{The UHD dehazing image on the left has obvious noise. And the image on the right with Tucker regular is clearer.}
	\label{noise}
\end{figure}

\subsection{Tucker Reconstruction on U-Net}
At present, the widely used encoding structures \cite{vincent2008extracting,vincent2010stacked,li2015deep} can effectively reduce gaussian noise. However, it is not enough to only rely on the traditional encoding and decoding structure for the high-resolution dehazing process with hazy images because most of the mapping is focused on low-frequency components like color and illumination. 

To address the problems, we propose a regularization term based on Tucker reconstruction \cite{kolda2009tensor} for U-Net. It can keep the global structure of low-frequency images well and further constrain the abnormal information in space.
As for the input tensor $\chi\epsilon{R^{I\times{J}\times{K}}}$,
\begin{small}
	\begin{eqnarray}
		\chi\approx{\mathcal{G}{\times}_1{A}\times_2B\times_3C}=\sum_{p=1}^P\sum_{q=1}^Q\sum_{r=1}^R\mathcal{G}_{pqr}a_p\circ{b_q}\circ{c_r}.
	\end{eqnarray}
\end{small}
\noindent
where $P,Q,R$ is less than $I,J,K$. $A \epsilon {R}^{I\times{P}}$, $B \epsilon {R}^{J\times{Q}}$, $C \epsilon {R}^{K\times{R}}$, are matrices of factors on different dimensions. $\mathcal{G}_{pqr}$ is the core Tensor of $\chi$. It can be considered as the compression of $\chi$.
We use tensorly's dependency library\footnote{https://github.com/ibalazevic/TuckER}  to reconstruct the input tensor, and the denoised image can be obtained effectively. The Svd is set to numpy-svd, our reconstruction error is less than $10e-5$, the maximum number of iteration is 100 and the random state is set to 1. 
As shown in Figure \ref{noise}, it can be found that there are noises and artifacts in images without Tucker reconstruction.
Please see supplementary materials for more cases of Tucker reconstruction that can reduce noise.

\begin{figure}[h]\scriptsize   
	\begin{center}
		\begin{tabular}{@{}cc@{}}
			\includegraphics[width = 0.46\linewidth]{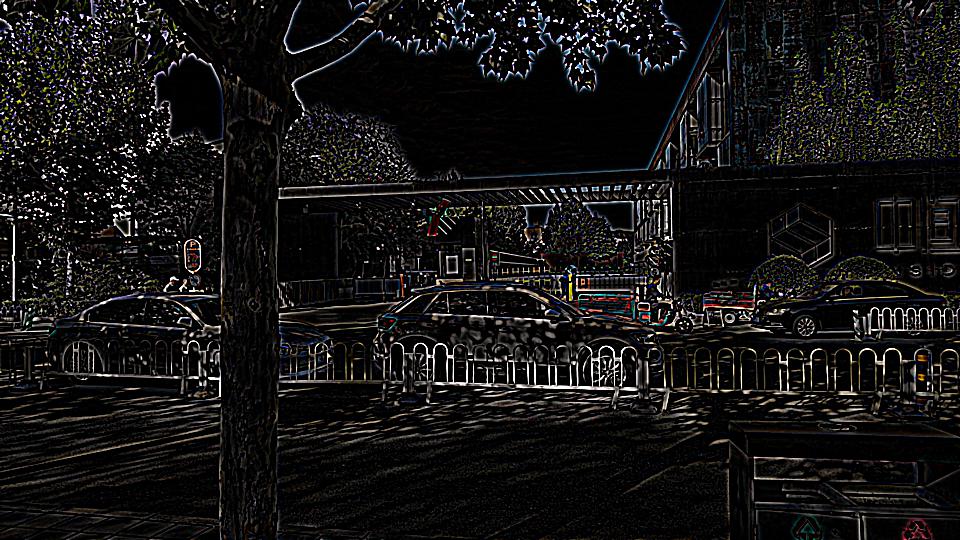}              &
			\includegraphics[width = 0.46\linewidth]{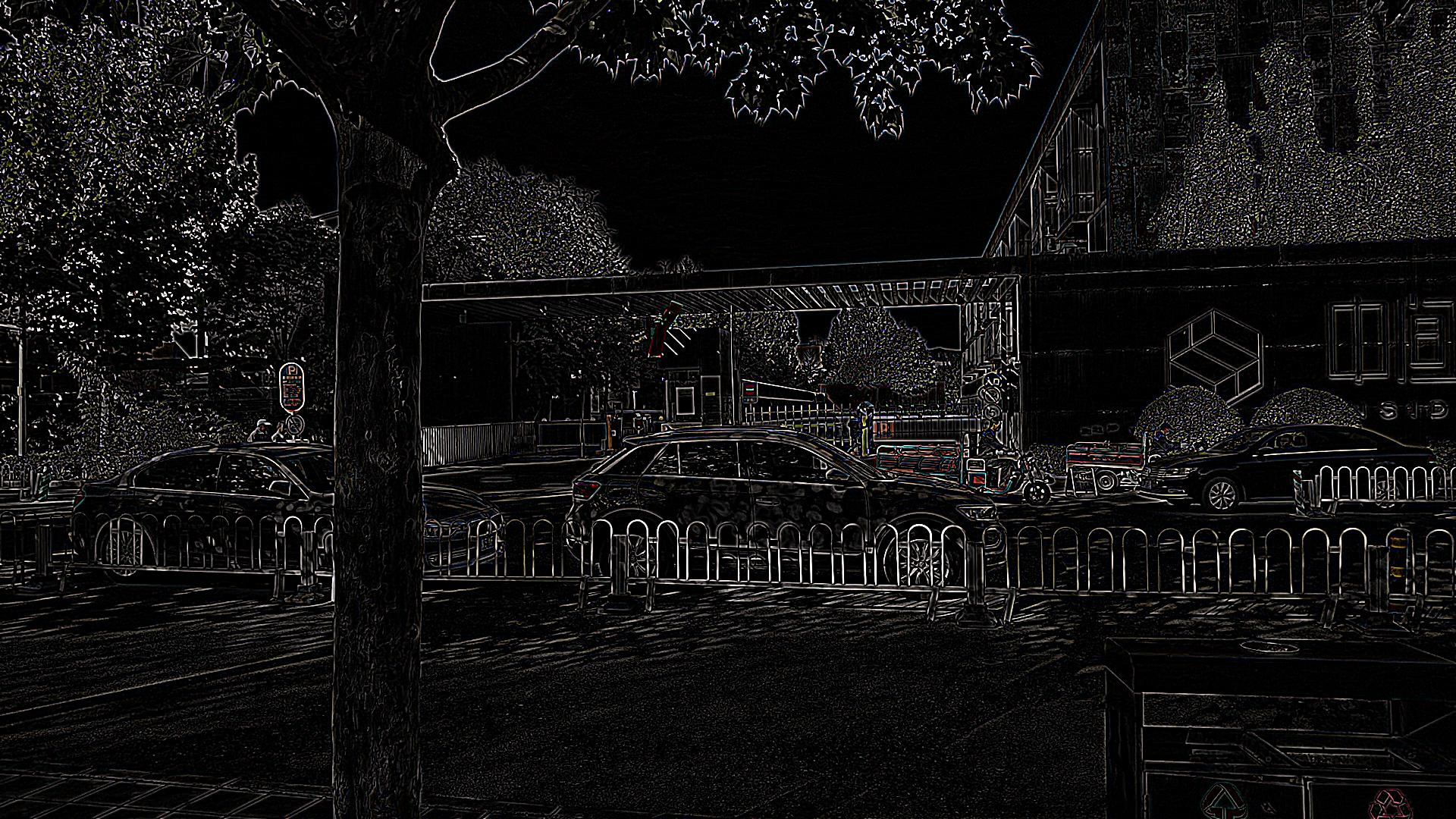}                     \\
			(a) The second level of pyramid.   &
			(b) The third level of pyramid.      \\
			
		\end{tabular}
	\end{center}
	\vspace{-4mm}	
	\caption{Different high-frequency components of the image have structural similarities.}
	\vspace{-2mm}
	\label{high-input}
\end{figure}

\subsection{The Attention Sharing Tensor}
We obtain an attention sharing tensor $K$ by learning the Taylor low-order components $q_n$,  $q_n'$ and the high-frequency components $q_{n-1}$ obtained by Laplace decomposition through the low-rank U-Net (see Figure \ref{network}). We approximate the differential operation in Taylor expansion by multiplying each high-frequency component with $K$. As shown in Figure \ref{high-input}, the tensor can be shared because  the high-order components have a high degree of structural similarity. Importantly, the attention sharing tensor focus on texture details in high-frequency components. This allows texture details to be better expressed. As shown in Figure \ref{Lap}, each high-order Taylor term is visualized. This means that the high order Taylor terms obtained by the attention sharing weight tensor work well. 

	\vspace{-2.5mm}
\section{Experiment}
In the section, we evaluate the proposed method by conducting experiments on both synthetic datasets and real world images. All the results are compared against nine state-of-the-art dehazing methods: DCP~\cite{he2010single}, AOD-NET~\cite{li2017all}, PSD~\cite{chen2021psd}, CAP~\cite{zhu2015fast}, non-local~\cite{berman2016non}, MGBL~\cite{zheng2021ultra}, DehazeNet~\cite{cai2016dehazenet}, PFF~\cite{mei2018progressive}, GCA~\cite{chen2019gated}. In addition, we conduct ablation studies to demonstrate effectiveness of each module of our approach.

\begin{table*}[!h]
	\scriptsize
	\begin{center}
		\resizebox{\textwidth}{!}{
			\begin{tabular}{ccccccccccccccc}
				\hline
				&      & Input  & CAP\cite{zhu2015fast} & non-local\cite{berman2016non} & DCP \cite{he2010single} & PSD \cite{chen2021psd} & DehazeNet\cite{cai2016dehazenet}& AOD \cite{li2017all} & MGBL \cite{zheng2021ultra}& PFF\cite{mei2018progressive}  & GCA \cite{chen2019gated}&   Ours           \\ \hline
				\multirow {3} {*} {4KID} & PSNR & 11.86 & 17.67 & 15.61 & 15.32 & 16.22 & 14.02 & 17.86 & 22.58  & 21.52 & 19.26 &  \textbf{27.79} \\
				& SSIM & 0.7421 & 0.8361 & 0.67   & 0.7511 & 0.81  & 0.7644   & 0.8519 & 0.87 & 0.7352     & 0.8341 &  \textbf{0.92} \\ 
				& time & - & 23s & 286s & 416s & 20ms  & 97ms   & 26ms & 15ms & 38ms & 95ms &  \textbf{14ms} \\ 
				\multirow {3} {*} {O-HAZE}   & PSNR & 13.11 & 14.55 & 18.44 & 16.57 & 16.45 & 17.57 & 15.10 & 18.43 &22.93  & 18.05 &  \textbf{23.2} \\
				& SSIM & 0.5643 & 0.5674 & 0.7219 & 0.7350 & 0.61 & 0.7676    & 0.7448 &  0.8164 & 0.75 & 0.7436 &  \textbf{0.8338} \\ 
				& time & -& 39s & 506s & 612s  & 25ms & 105ms  & 34ms   & 27ms & 45ms & 102ms  & \textbf{26ms} \\ 
				\multirow {3} {*} {I-HAZE}     & PSNR & 13.15 & 13.42 & 16.02 & 13.57 &  14.49 & 10.10 & 17.63 & 18.75 & 18.41 &18.57  &  \textbf{18.77} \\
				& SSIM & 0.75 & 0.65 & 0.76   & 0.68 & 0.626 & 0.5976    &  0.7850 & 0.8164 & 0.7821  & 0.7436 &  \textbf{0.7838} \\ 
				& time & - & 29s & 559s   & 526s & 27ms  & 97ms  & 30ms & 26ms & 42ms & 101ms &  \textbf{25ms} \\ 
				\hline
			\end{tabular} 
		}
	\end{center}
		\caption{Our method has the best evaluation on the proposed dataset (4KID), O-HAZE, and I-HAZE.}
	\label{dataset}
\end{table*}
\vspace{-2mm}
\begin{figure*}[!hpt]\tiny
	\tabcolsep 1pt
	\begin{tabular}{@{}cccccc@{}}
		\includegraphics[width = 0.16\textwidth]{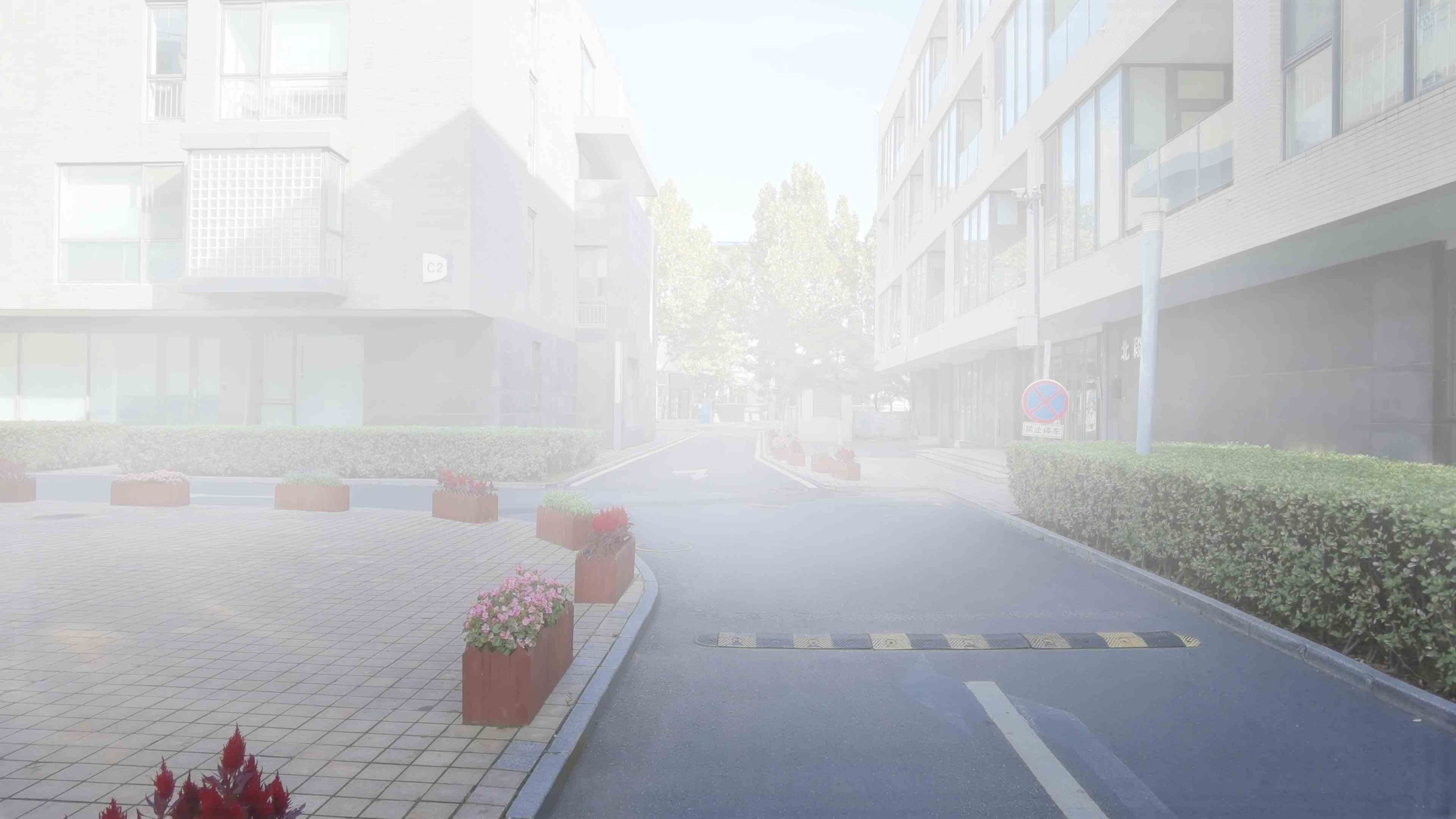}              &
		\includegraphics[width = 0.16\textwidth]{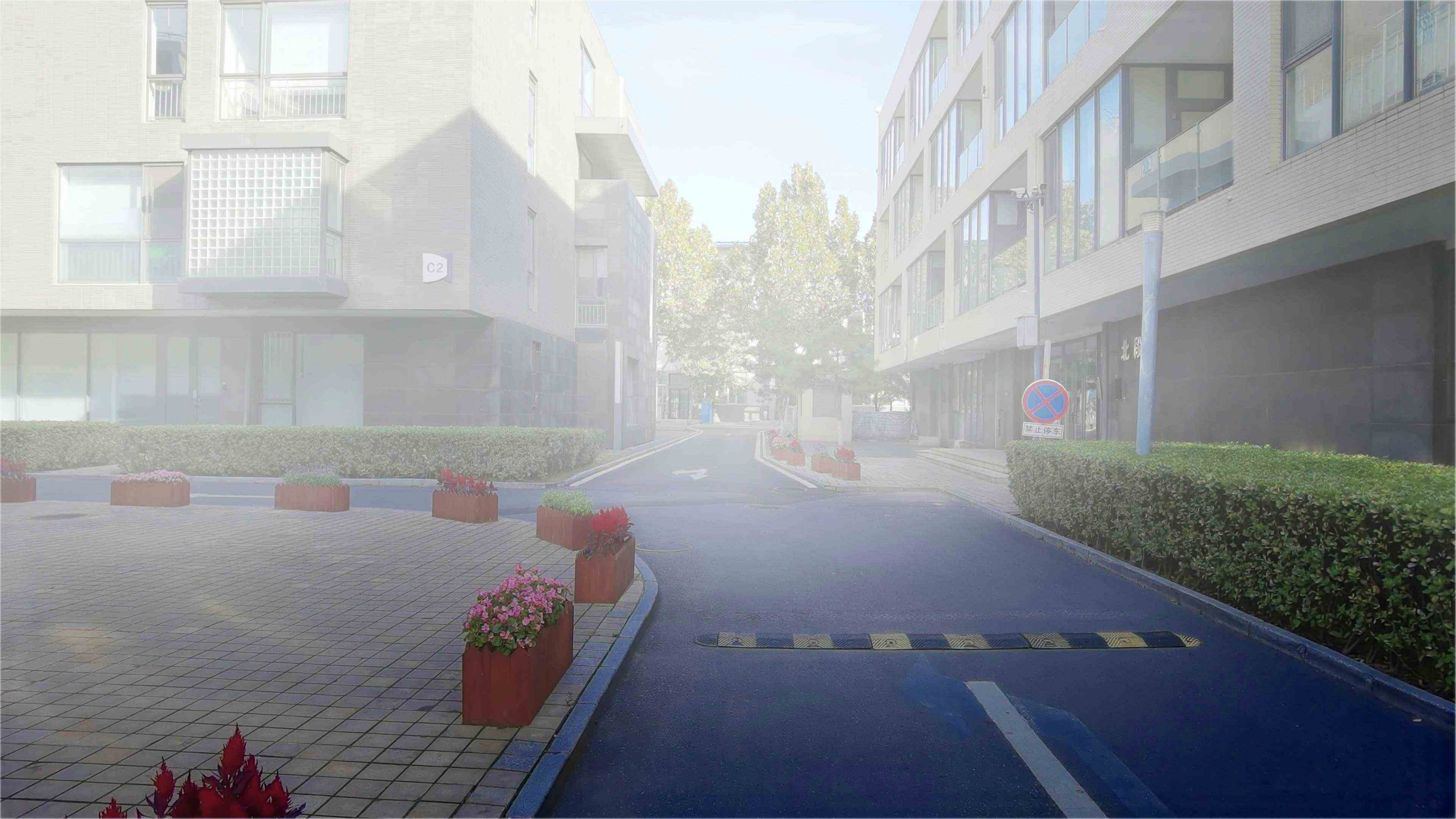}                 &
		\includegraphics[width = 0.16\textwidth]{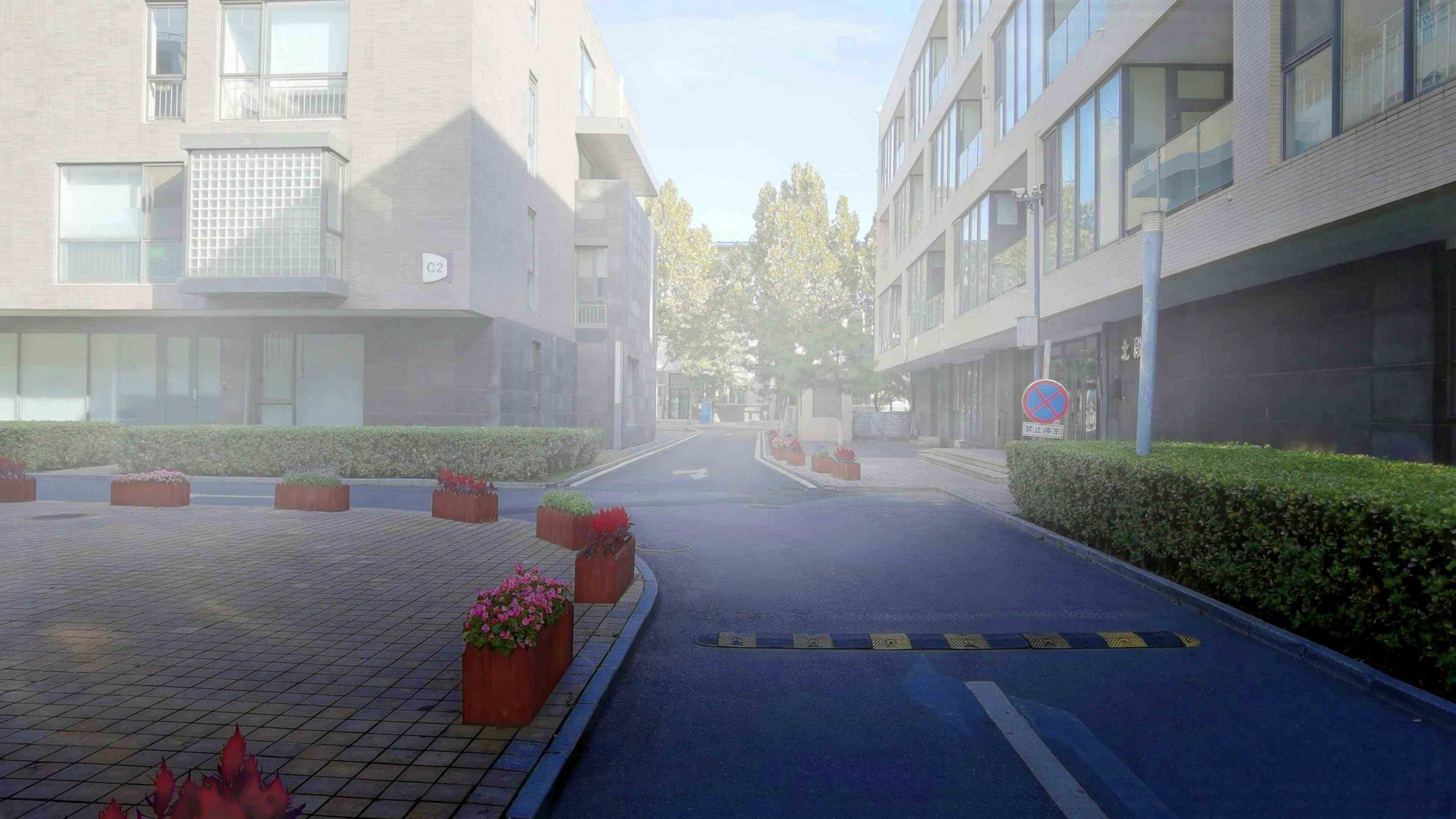}               &
		\includegraphics[width = 0.16\textwidth]{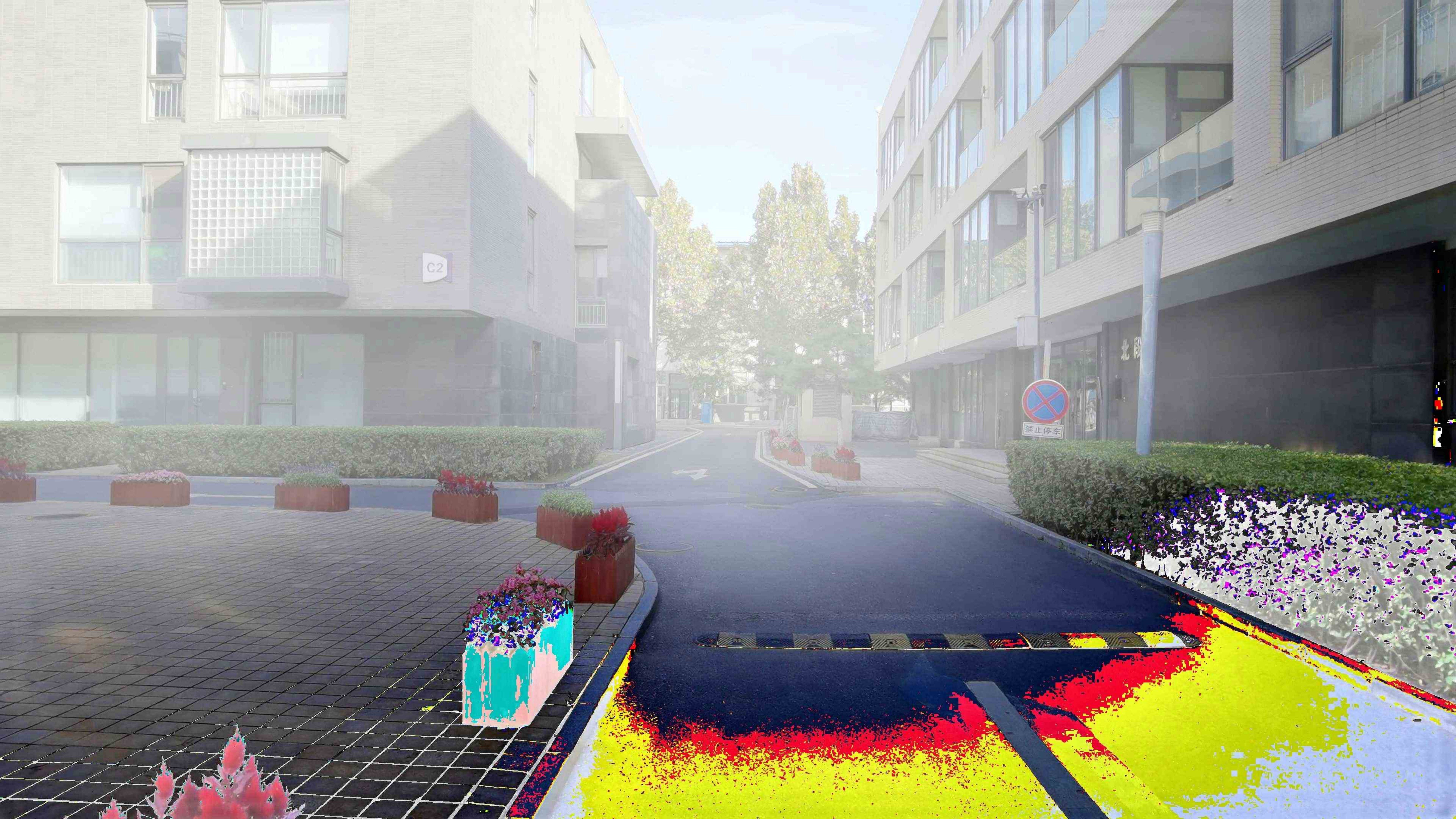}                &
		\includegraphics[width = 0.16\textwidth]{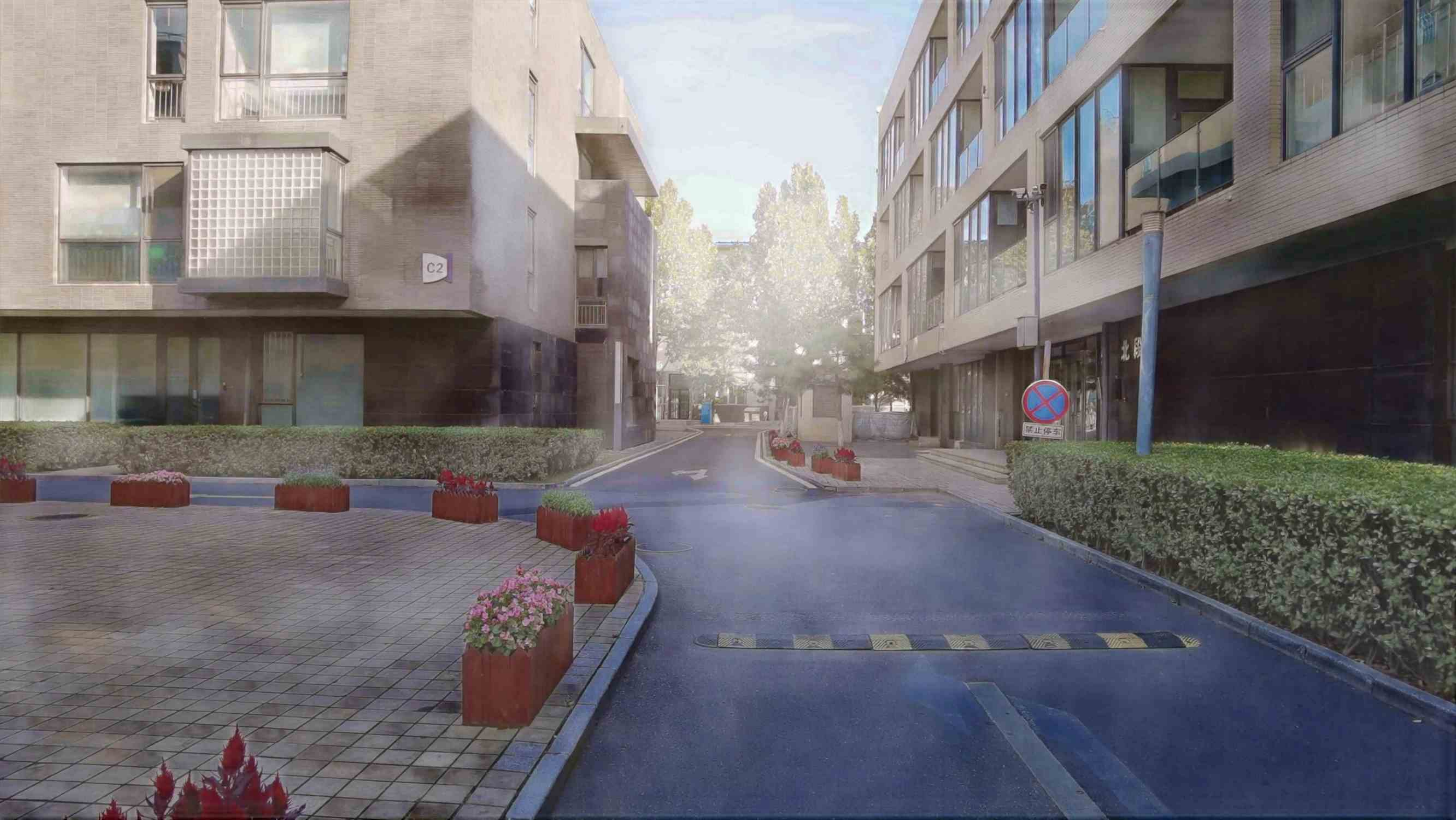}                 &
		\includegraphics[width = 0.16\textwidth]{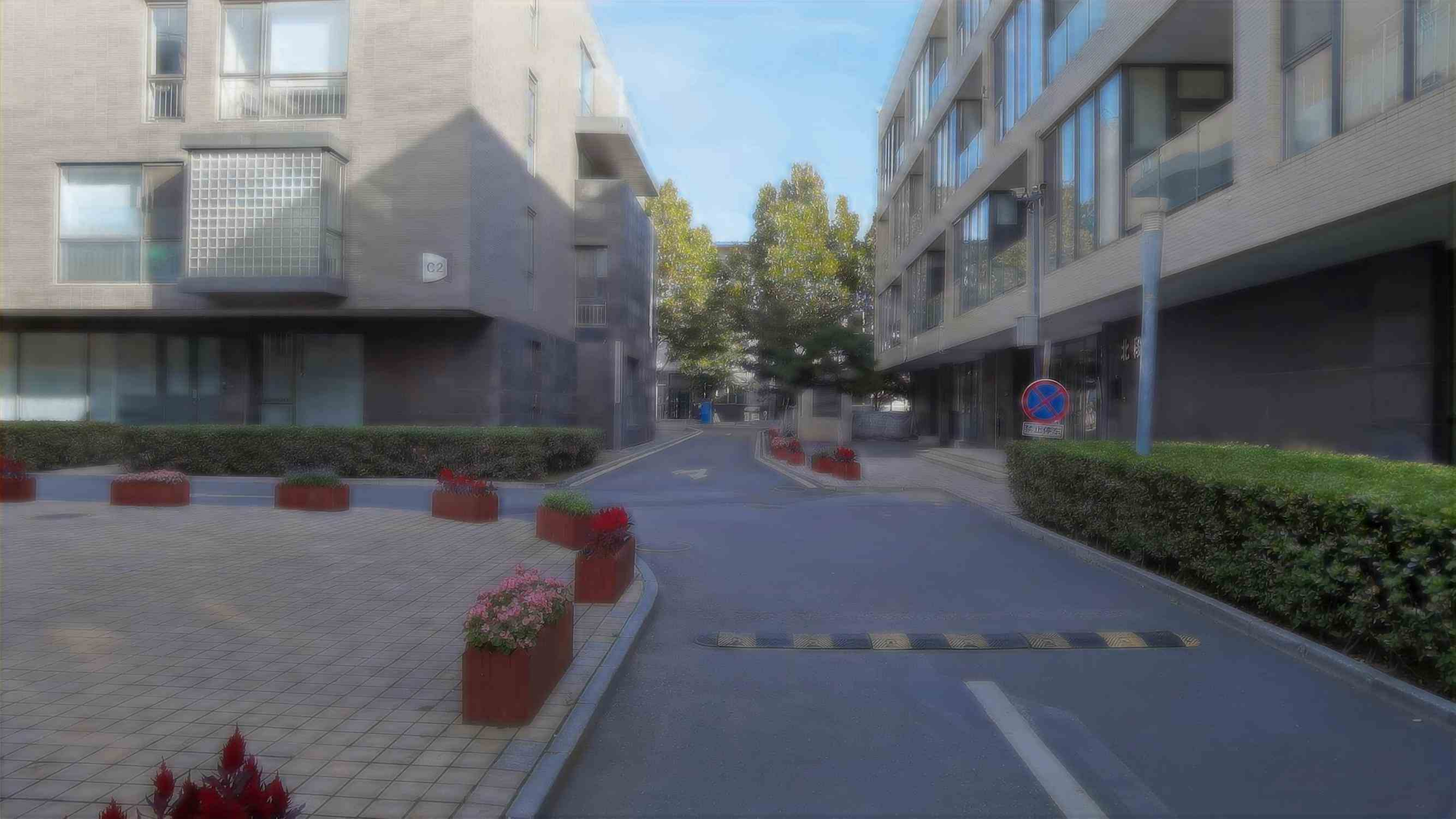}                       \\
		
		\includegraphics[width = 0.16\textwidth]{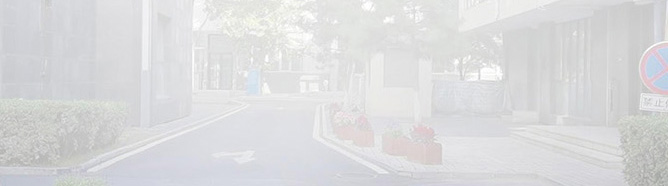}              &
		\includegraphics[width = 0.16\textwidth]{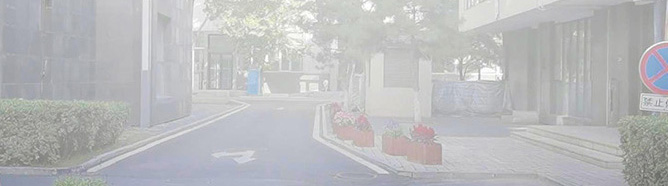}                 &
		\includegraphics[width = 0.16\textwidth]{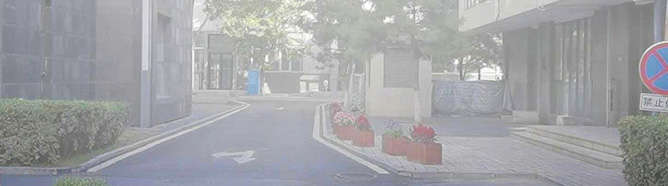}               &
		\includegraphics[width = 0.16\textwidth]{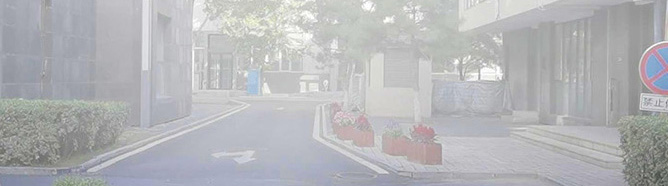}                &
		\includegraphics[width = 0.16\textwidth]{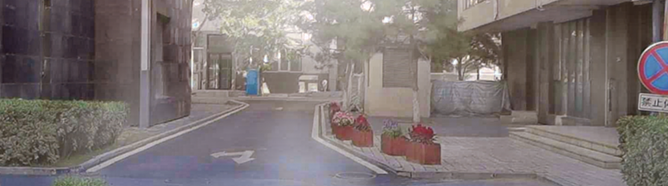}                 &
		\includegraphics[width = 0.16\textwidth]{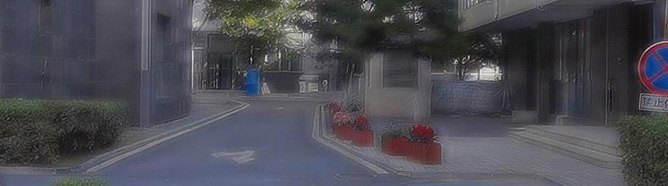}                       \\
		
		Input (PSNR/SSIM)&
		AOD-NET~\cite{li2017all} (11.75/0.5235)& 
		CAP~\cite{zhu2015fast} (13.19/0.5208)& 
		DehazeNet~\cite{cai2016dehazenet} (10.11/0.4685)& 
		GCA~\cite{chen2019gated} (17.21/0.6264)& 
		MGBL~\cite{zheng2021ultra} (21.79/0.7004)\\
		\includegraphics[width = 0.16\textwidth]{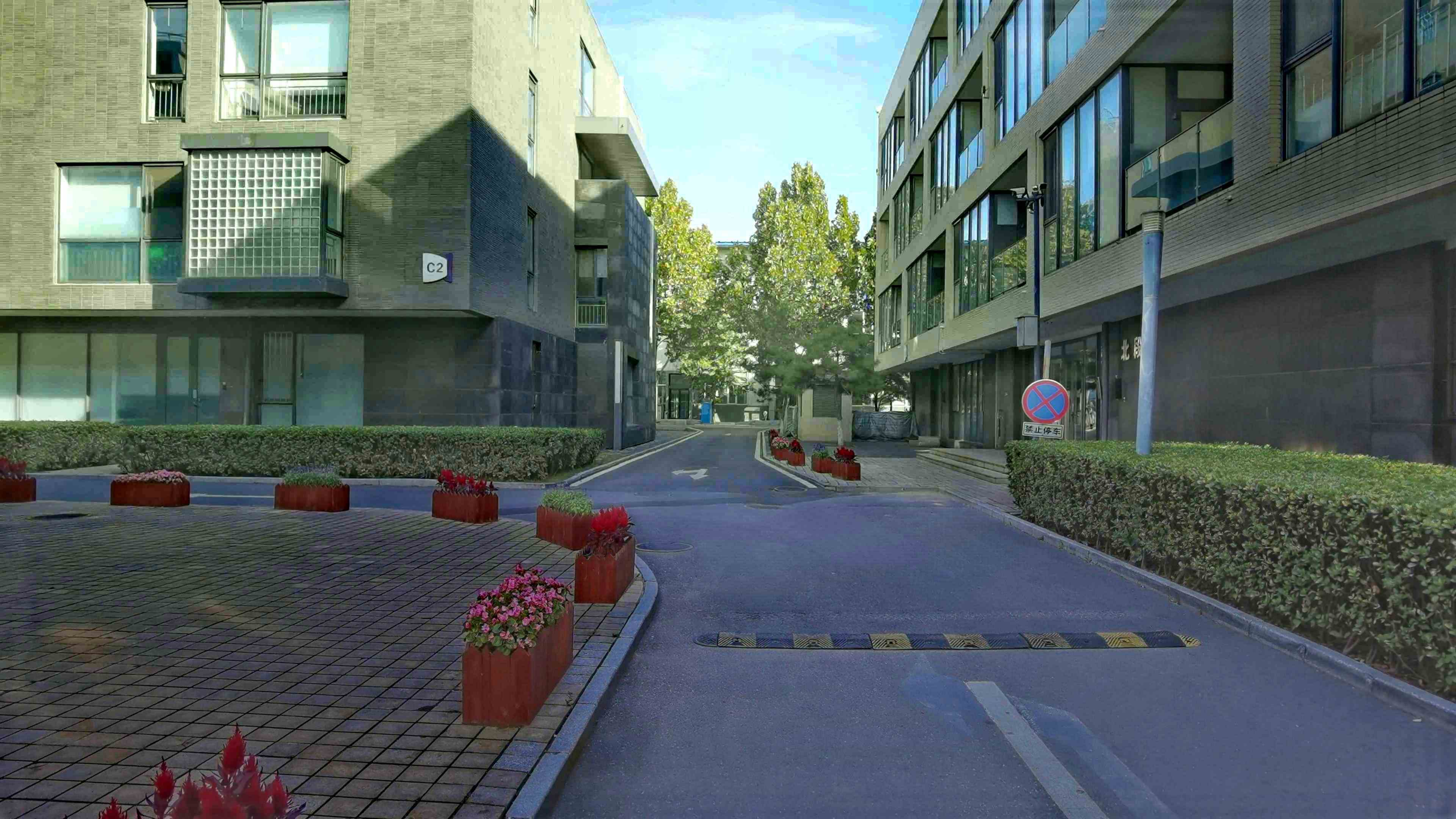}             & 
		\includegraphics[width = 0.16\textwidth]{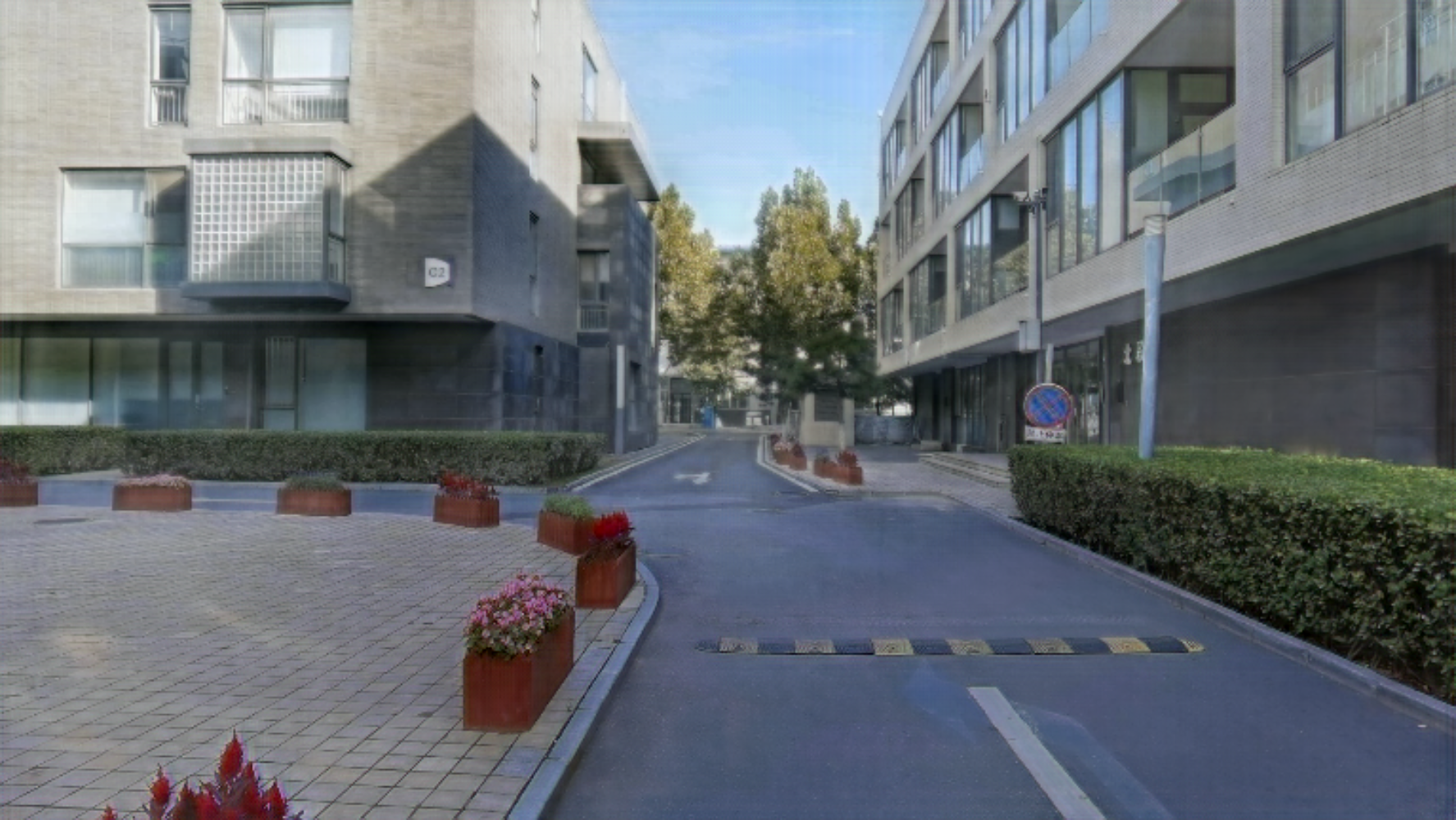}       &
		\includegraphics[width = 0.16\textwidth]{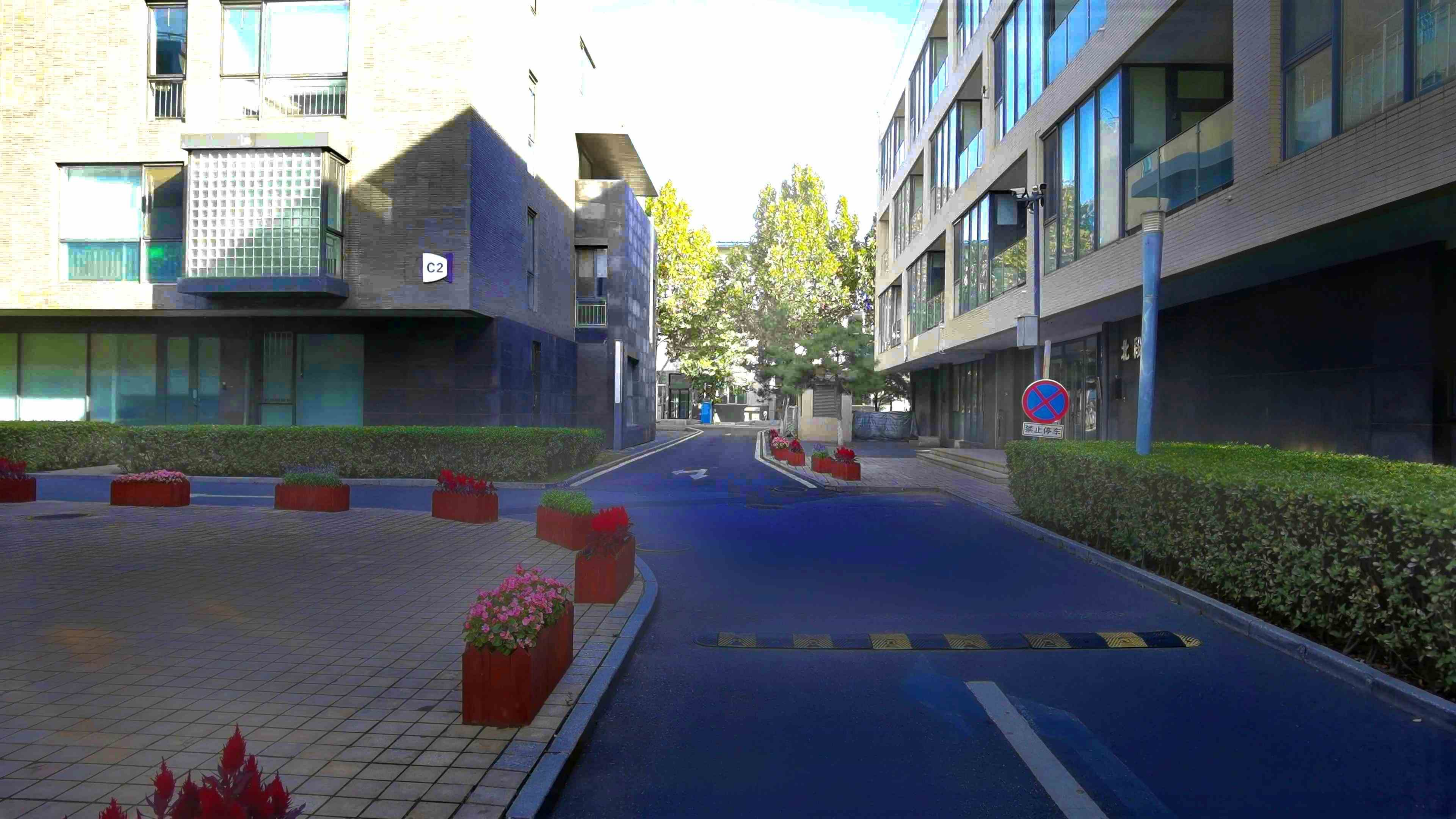}                & 
		\includegraphics[width = 0.16\textwidth]{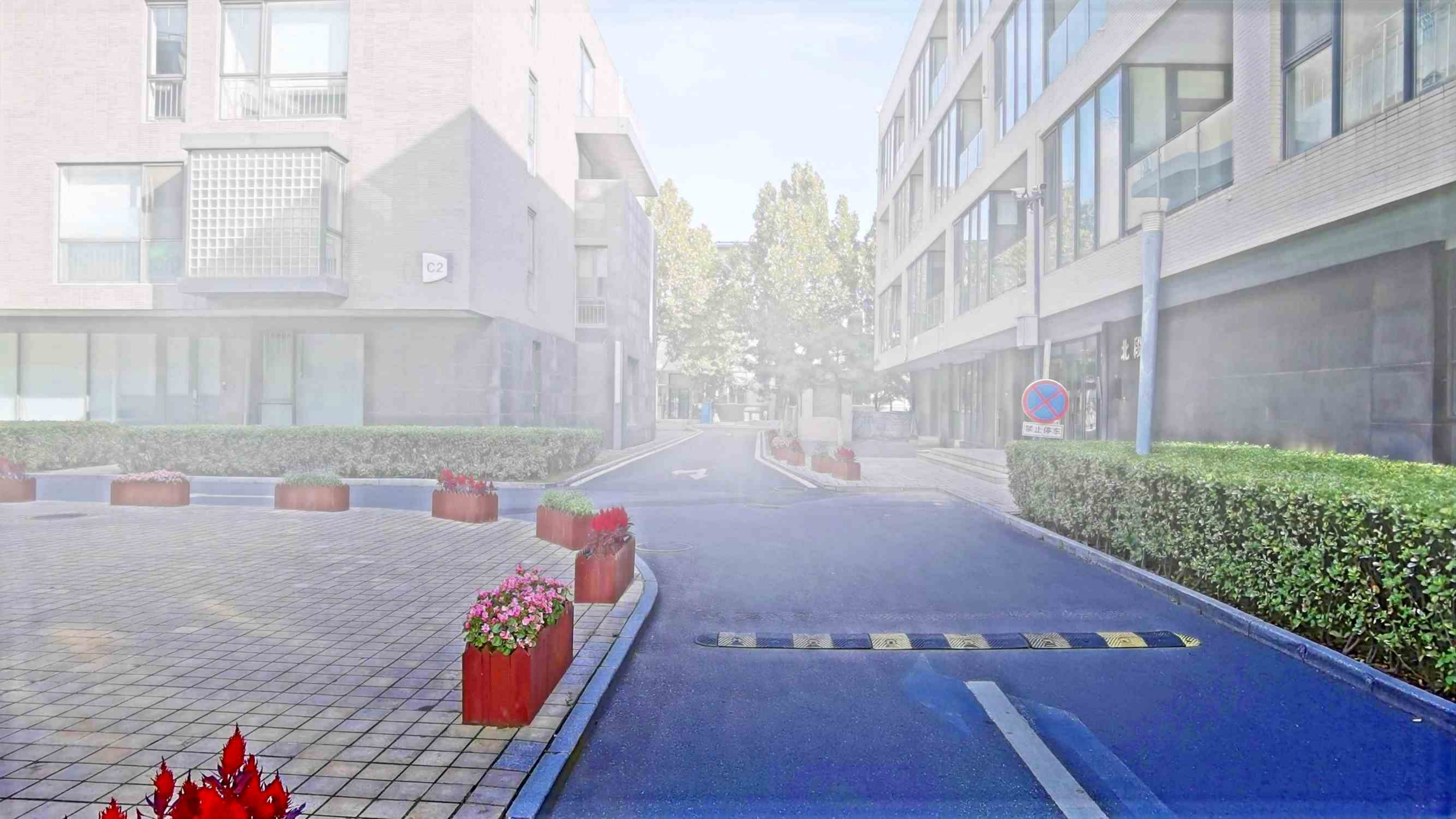}              &
		\includegraphics[width = 0.16\textwidth]{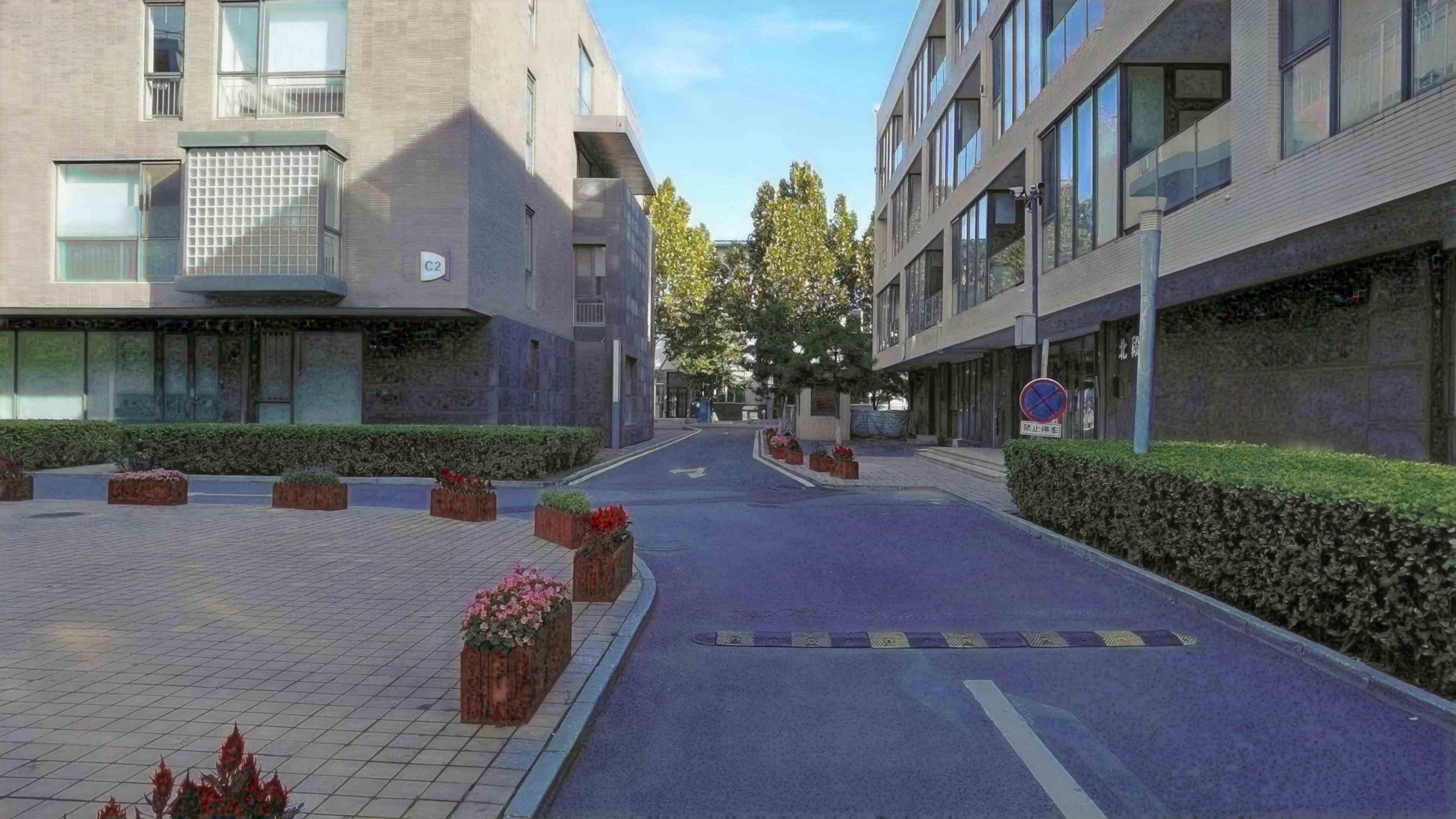}                 & 
		\includegraphics[width = 0.16\textwidth]{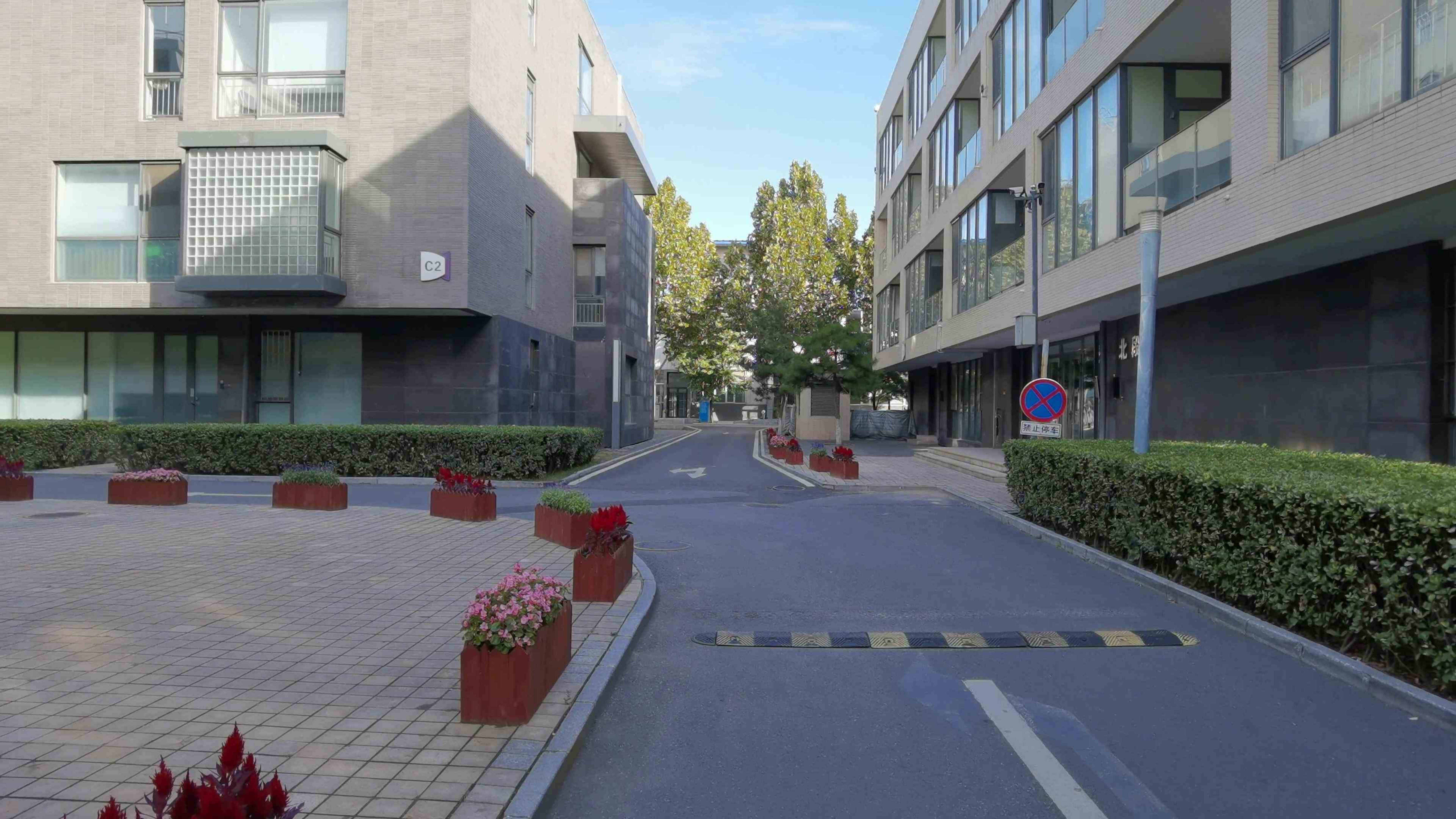}                     \\

		\includegraphics[width = 0.16\textwidth]{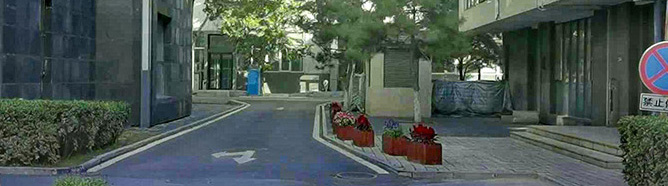}             & 
		\includegraphics[width = 0.16\textwidth]{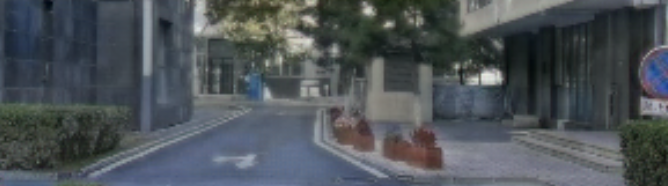}       &
		\includegraphics[width = 0.16\textwidth]{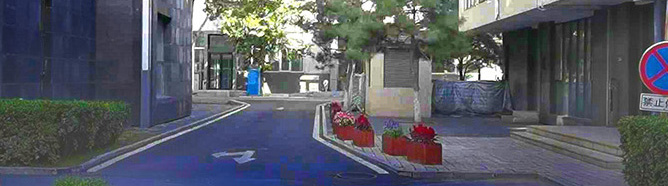}                & 
		\includegraphics[width = 0.16\textwidth]{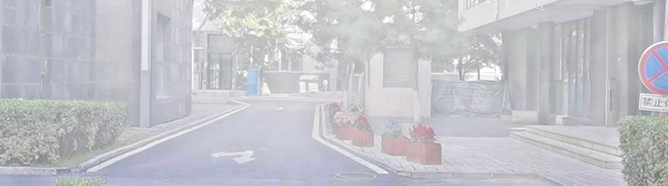}              &
		\includegraphics[width = 0.16\textwidth]{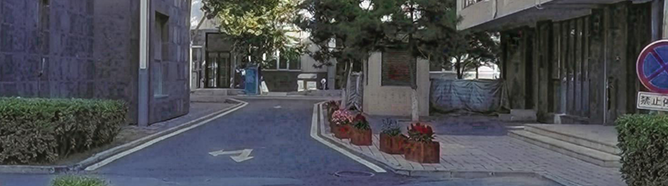}                 & 
		\includegraphics[width = 0.16\textwidth]{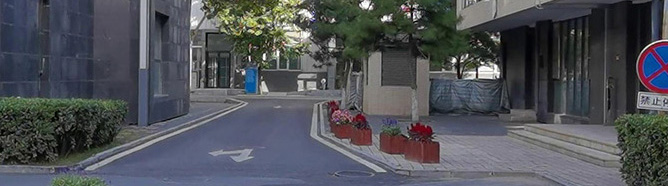}                     \\          
		NLD~\cite{berman2016non} (15.65/0.5132)&
		PFF~\cite{mei2018progressive} (19.5/0.51)& 
		DCP~\cite{he2010single} (15.15/0.6192)& 
		PSD~\cite{chen2021psd} (12.73/0.5723)& 
		\textbf{Ours (26.55/0.9045)}&
		GT ($+ \infty$/1)
		\\

		\includegraphics[width = 0.16\textwidth]{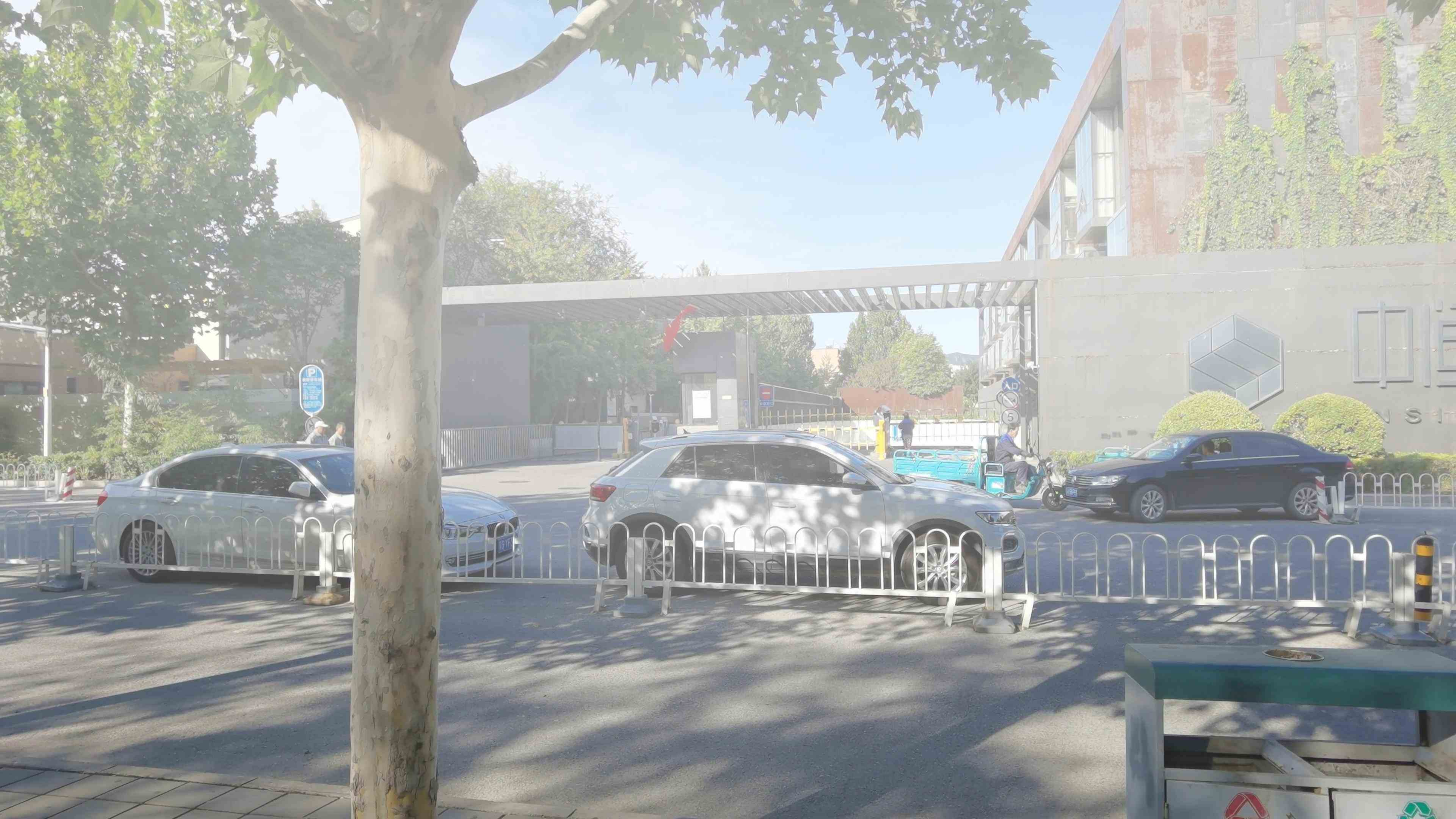}              &
		\includegraphics[width = 0.16\textwidth]{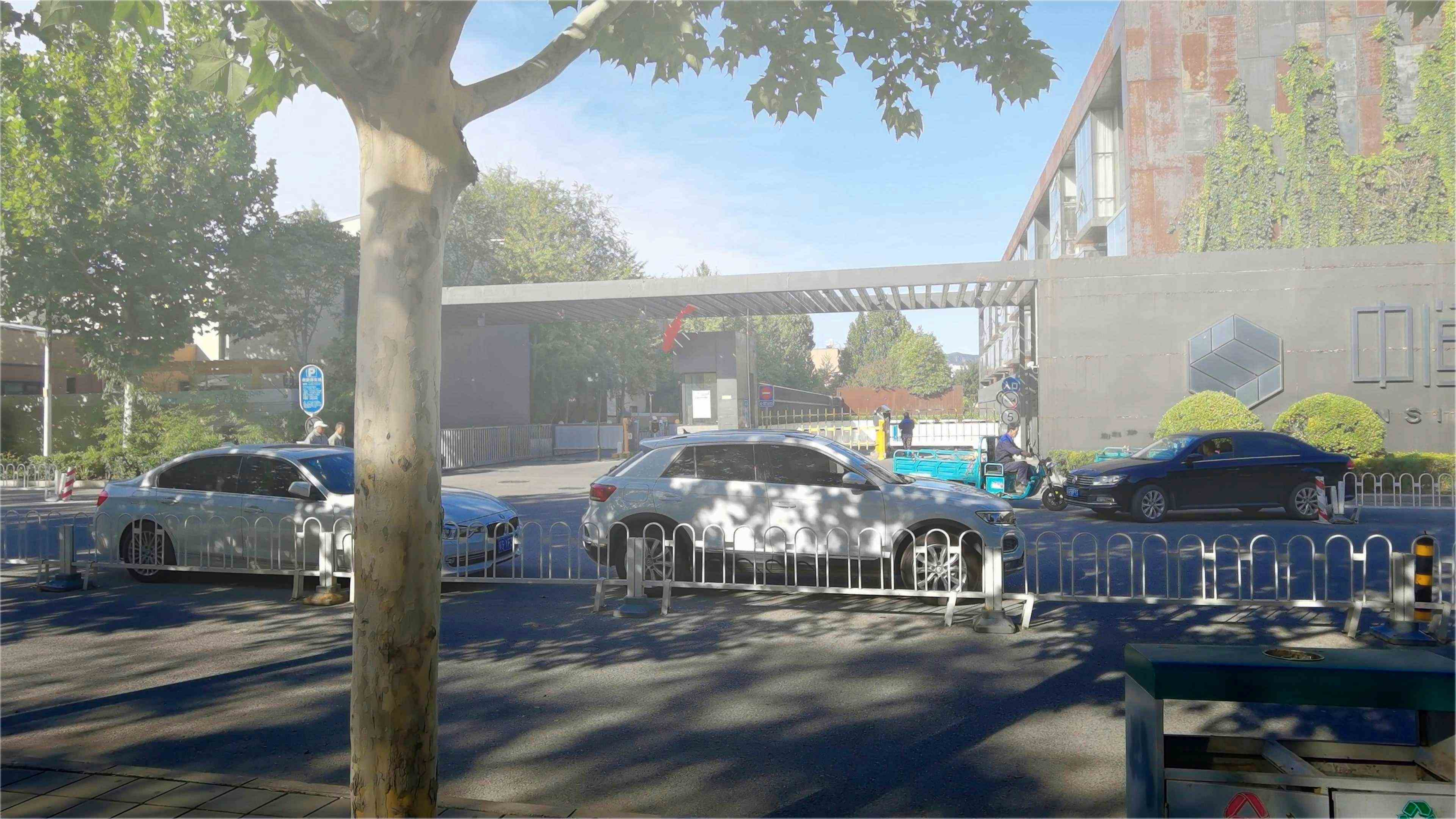}                 &
		\includegraphics[width = 0.16\textwidth]{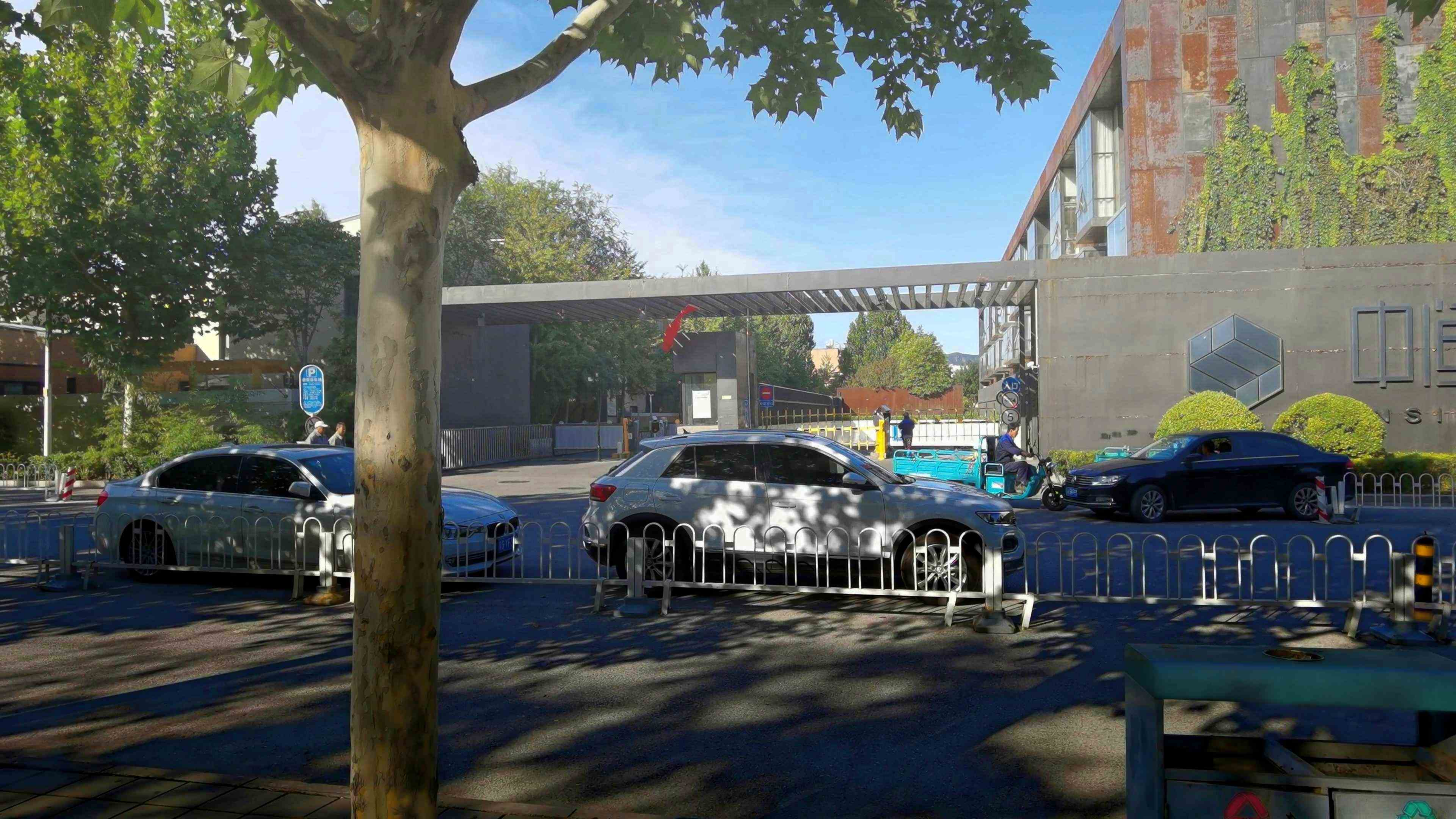}               &
		\includegraphics[width = 0.16\textwidth]{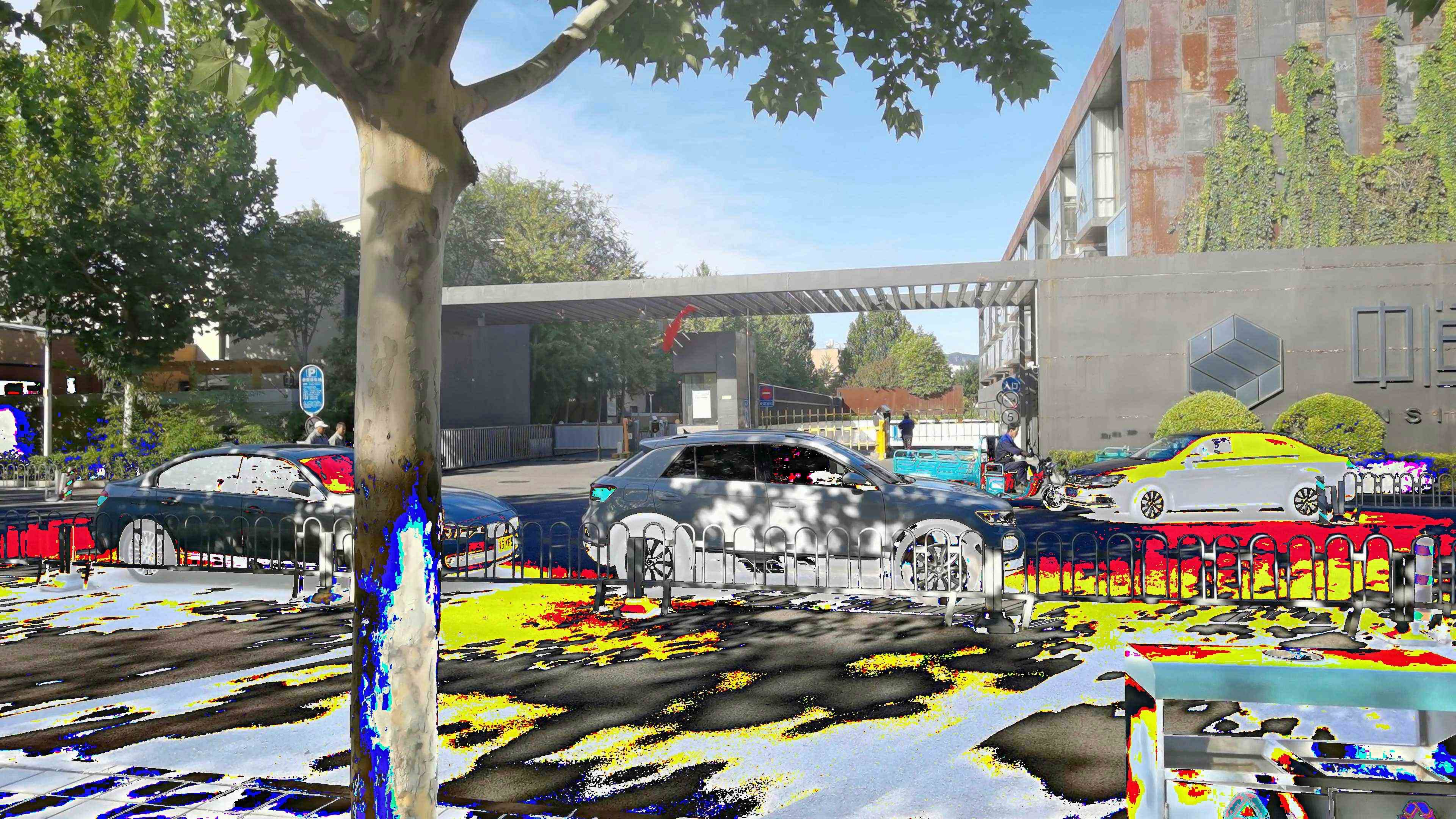}                &
		\includegraphics[width = 0.16\textwidth]{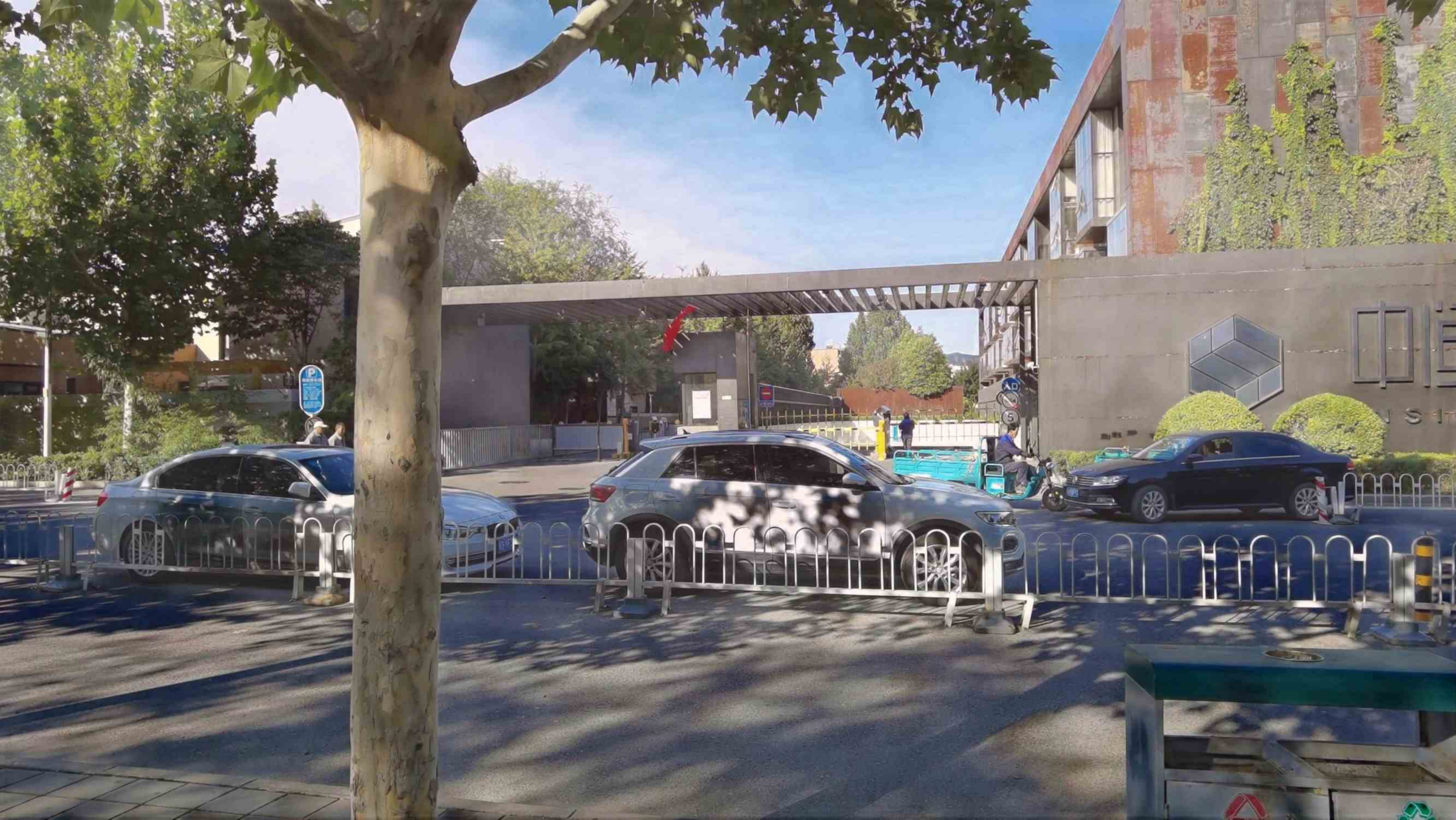}                 &
		\includegraphics[width = 0.16\textwidth]{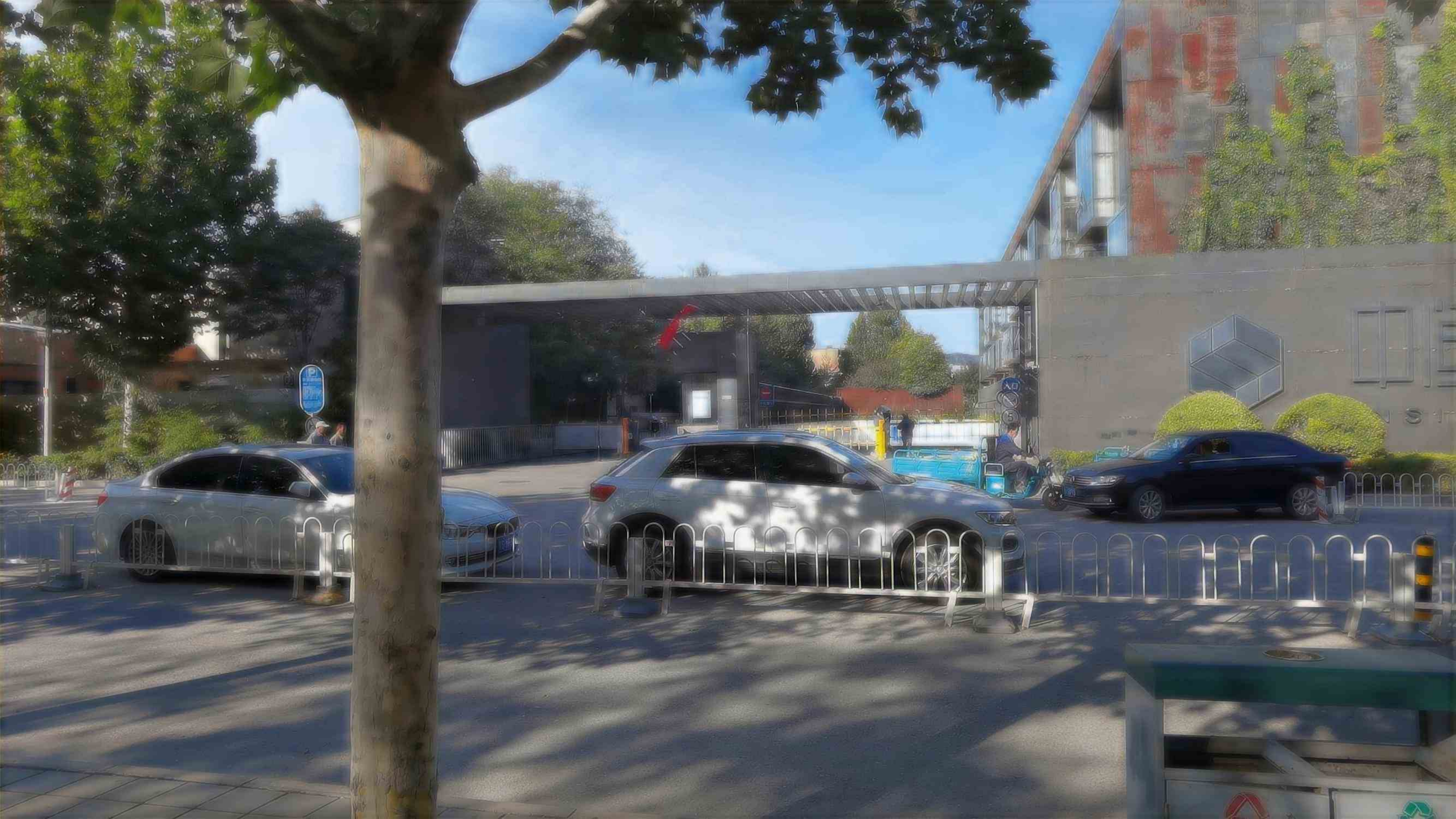}                       \\
		
		\includegraphics[width = 0.16\textwidth]{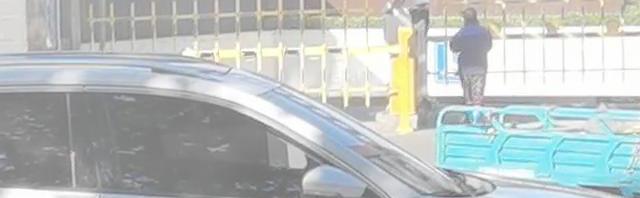}              &
		\includegraphics[width = 0.16\textwidth]{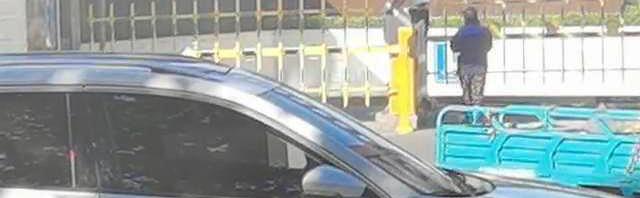}                 &
		\includegraphics[width = 0.16\textwidth]{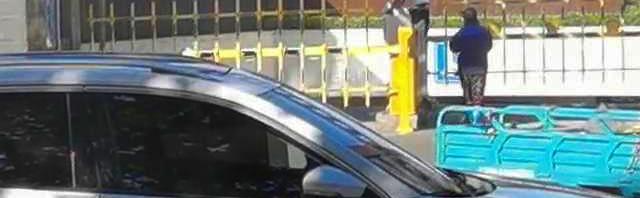}               &
		\includegraphics[width = 0.16\textwidth]{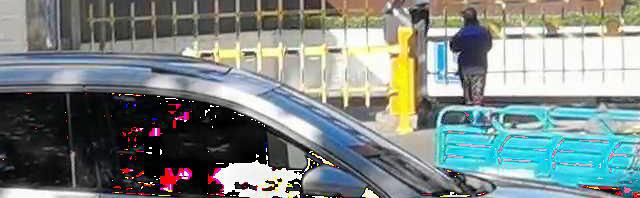}                &
		\includegraphics[width = 0.16\textwidth]{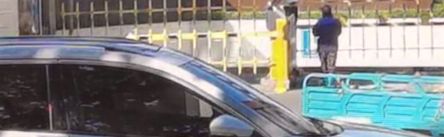}                 &
		\includegraphics[width = 0.16\textwidth]{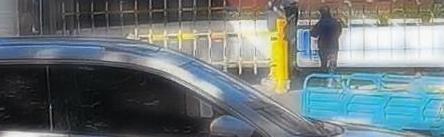}                       \\
		
		Input (PSNR/SSIM)&
		AOD-NET~\cite{li2017all} (14.96/0.5764)& 
		CAP~\cite{zhu2015fast} (15.60/0.5342)& 
		DehazeNet~\cite{cai2016dehazenet} (10.58/0.45)& 
		GCA~\cite{chen2019gated} (18.92/0.6264)& 
		MGBL~\cite{zheng2021ultra} (20.88/0.6848)\\
		
		\includegraphics[width = 0.16\textwidth]{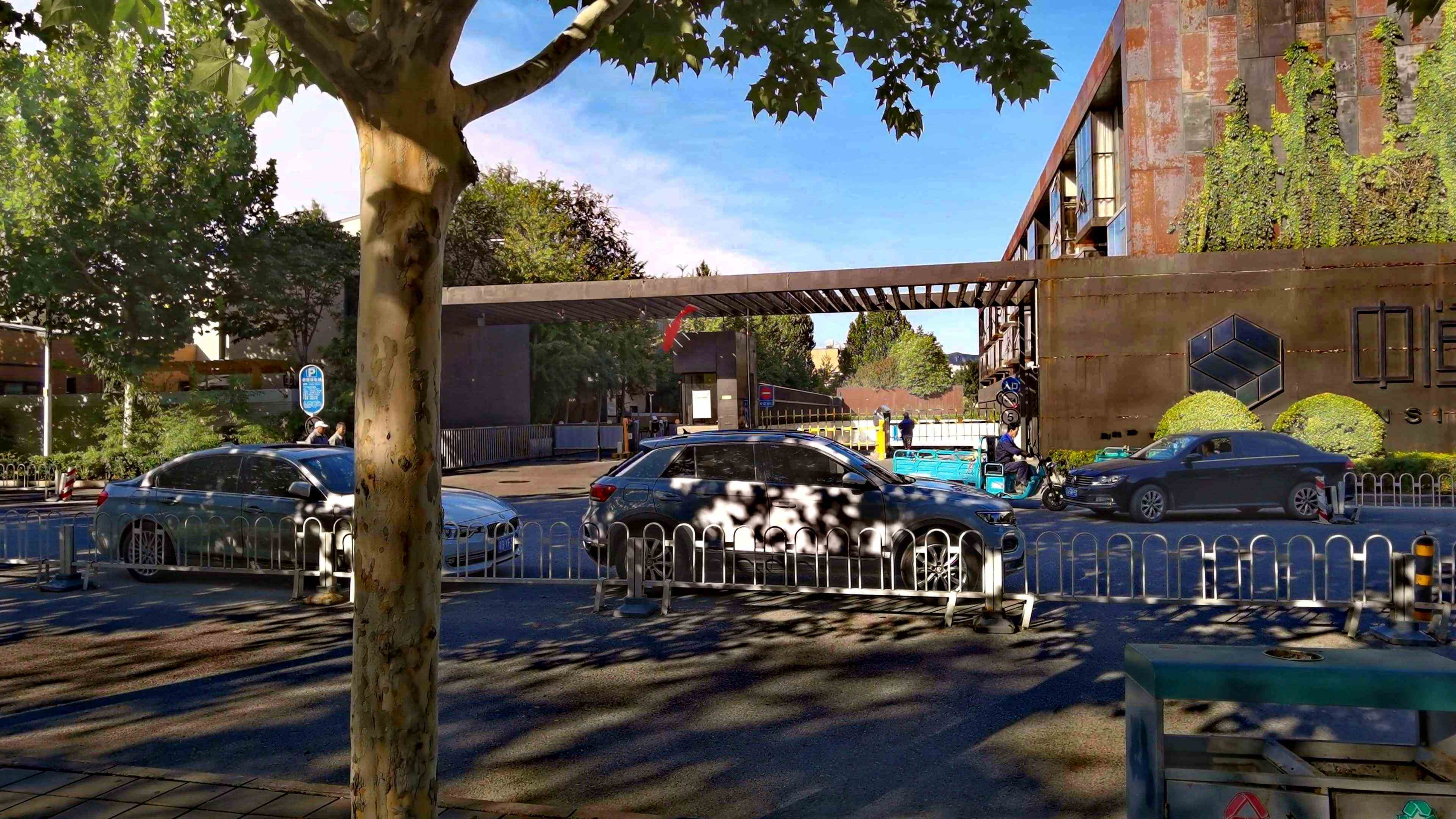}             & 
		\includegraphics[width = 0.16\textwidth]{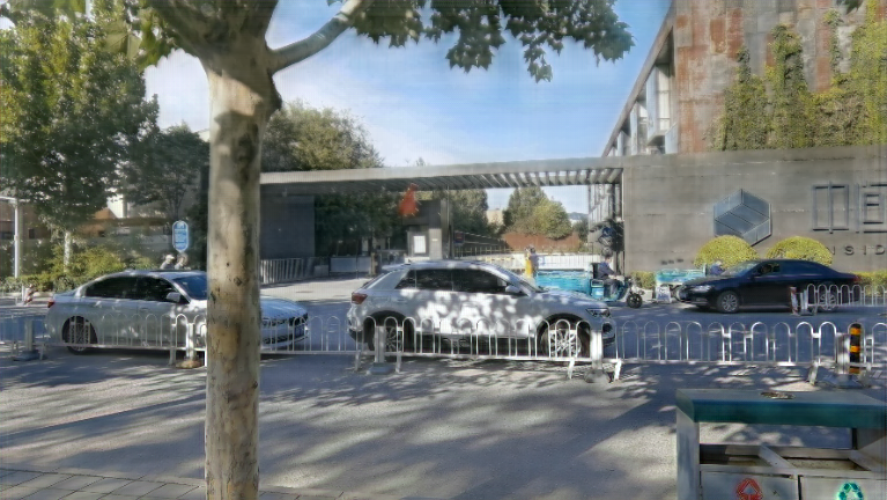}       &
		\includegraphics[width = 0.16\textwidth]{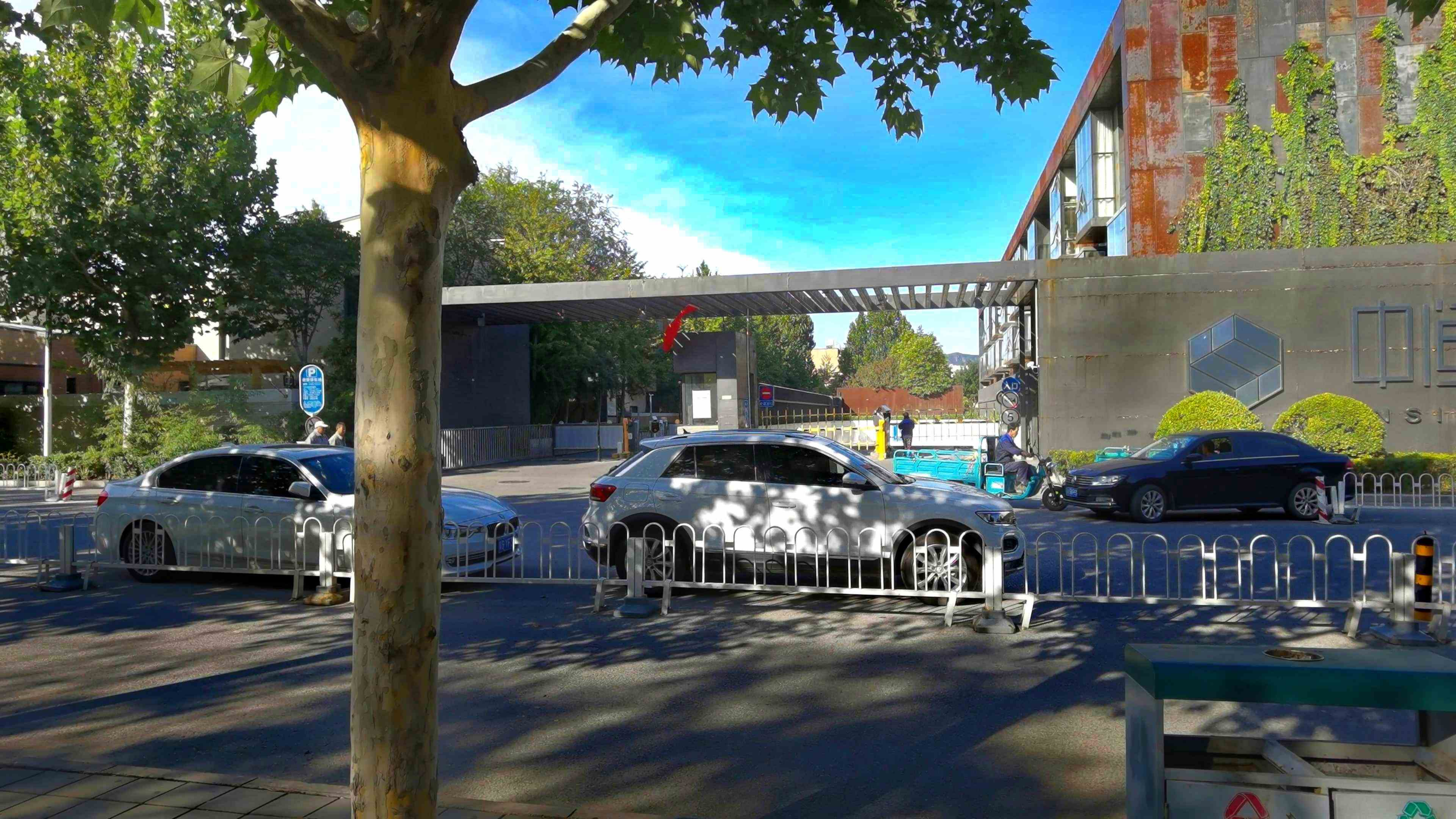}                & 
		\includegraphics[width = 0.16\textwidth]{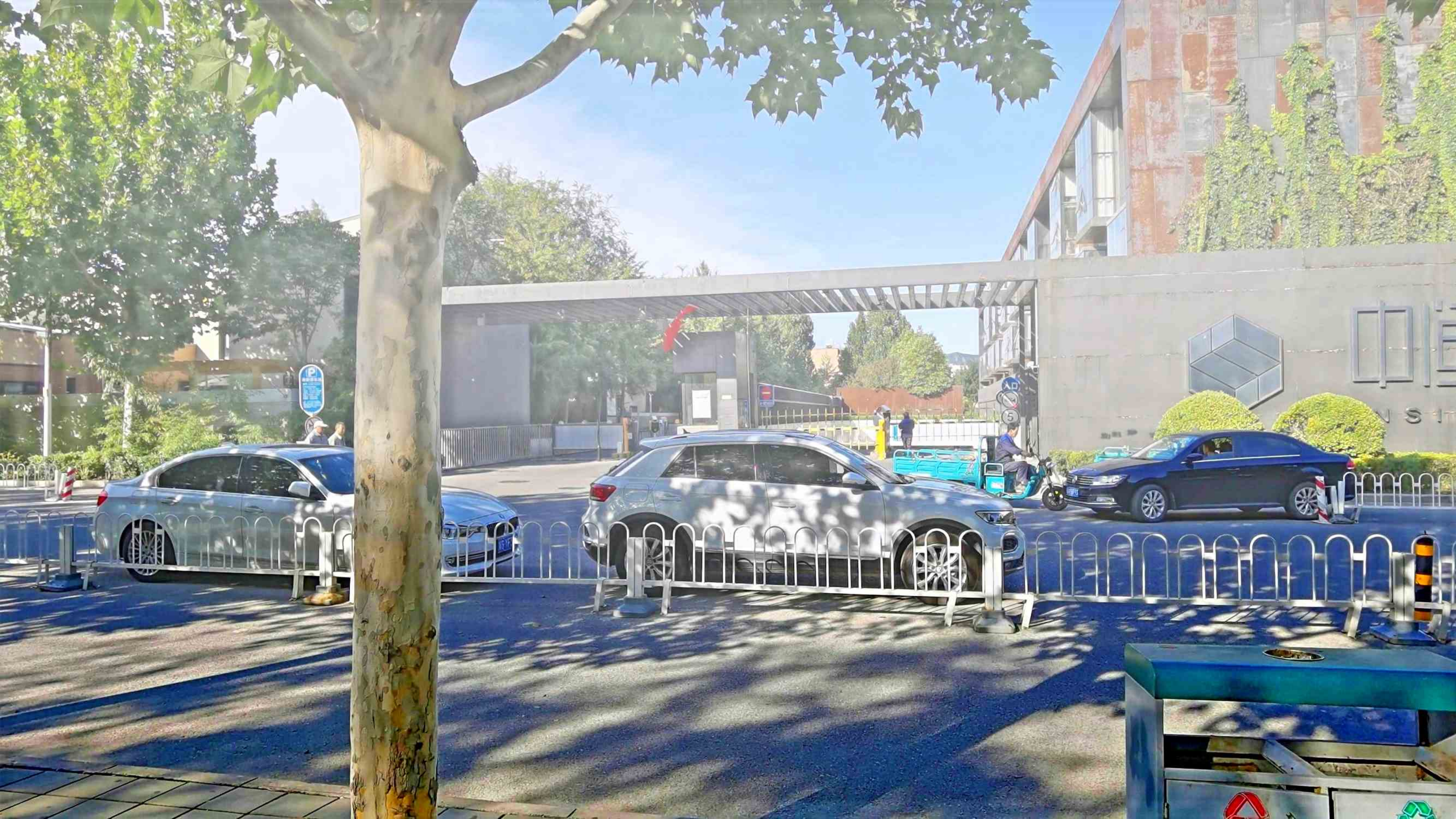}              &
		\includegraphics[width = 0.16\textwidth]{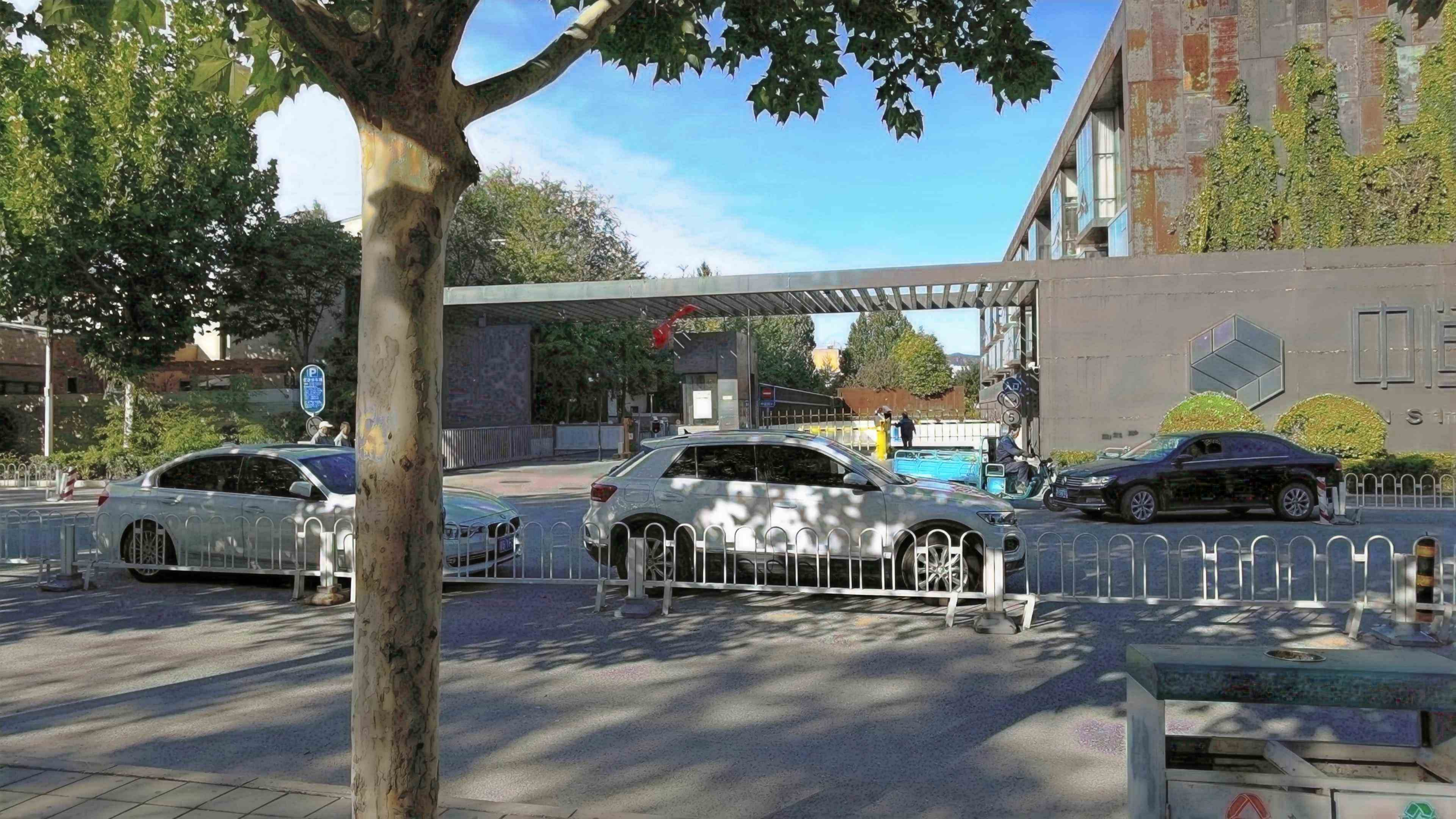}                 & 
		\includegraphics[width = 0.16\textwidth]{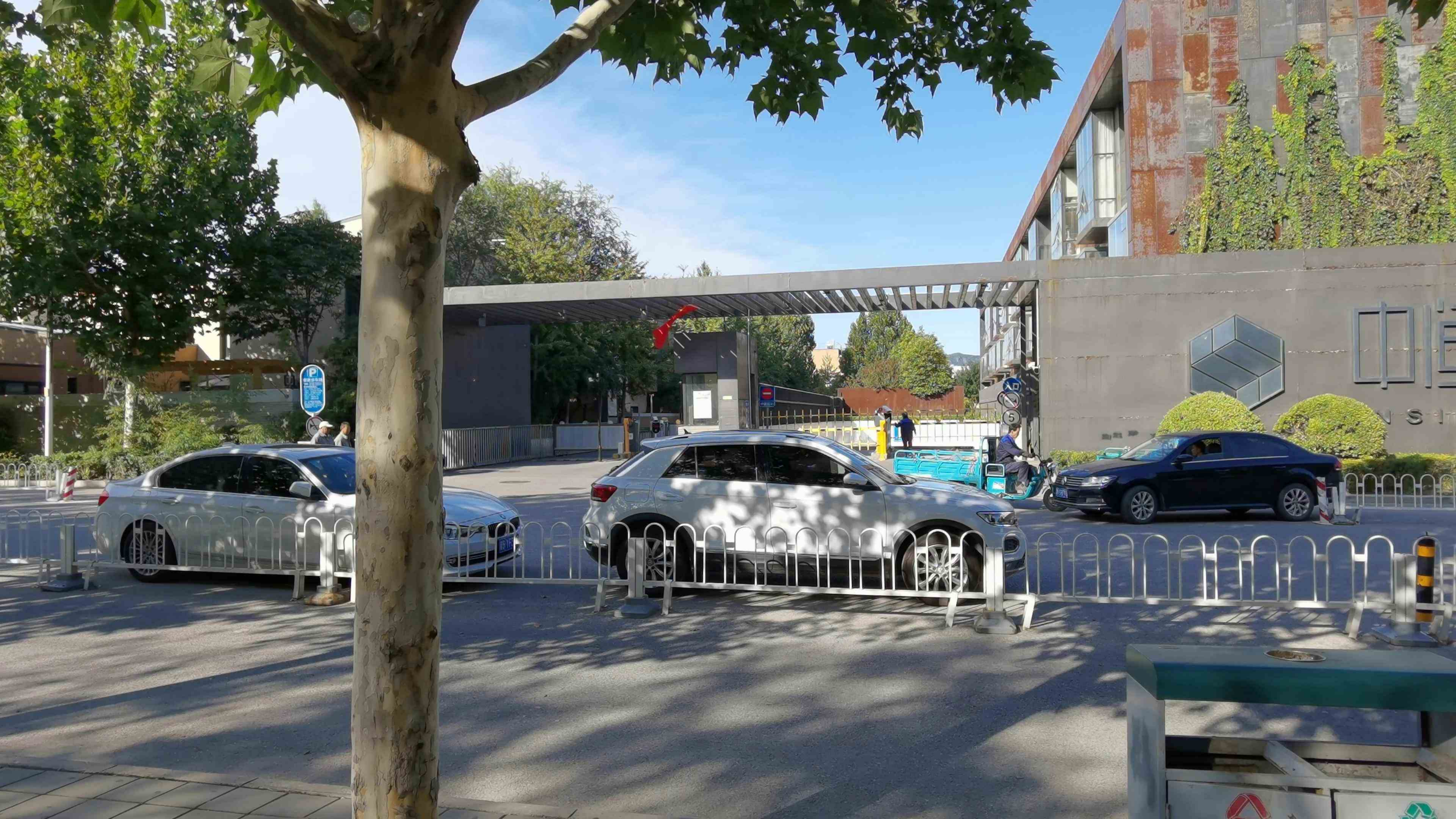}                     \\   
		
		\includegraphics[width = 0.16\textwidth]{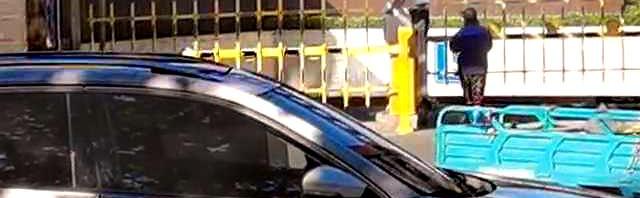}             & 
		\includegraphics[width = 0.16\textwidth]{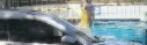}       &
		\includegraphics[width = 0.16\textwidth]{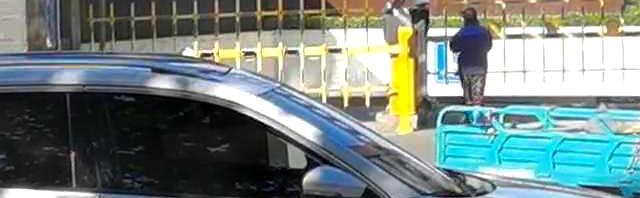}                & 
		\includegraphics[width = 0.16\textwidth]{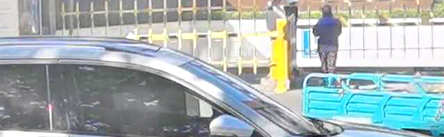}              &
		\includegraphics[width = 0.16\textwidth]{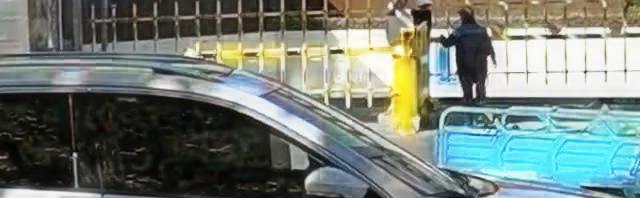}                 & 
		\includegraphics[width = 0.16\textwidth]{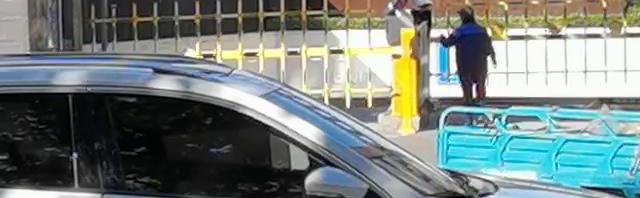}                     \\ 		
		
		NLD~\cite{berman2016non} (14.50/0.5418)&
		PFF~\cite{mei2018progressive} (14.74/0.28)& 
		DCP~\cite{he2010single} (16.85/0.5960)& 
		PSD~\cite{chen2021psd} (12.22/0.5134)& 
		\textbf{Ours (26.87/0.8941)}&
		GT ($+ \infty$/1)
		\\
		
	\end{tabular}
	
	\caption{Our method obtains better visual quality and recovers more image details compared with other state-of-the-art methods in the 4KID dataset.}	
	\label{4KID}
	\vspace{-2mm}
\end{figure*}

\begin{figure*}[ht]\tiny
	\begin{center}
		\tabcolsep 1pt
		\begin{tabular}{@{}cccccc@{}}
			\includegraphics[width = 0.16\textwidth]{img0_input.pdf}              &
			\includegraphics[width = 0.16\textwidth]{img0_AOD-Net.pdf}                 &
			\includegraphics[width = 0.16\textwidth]{img0_CAP.pdf}               &
			\includegraphics[width = 0.16\textwidth]{img0_DehazeNet.pdf}                &
			\includegraphics[width = 0.16\textwidth]{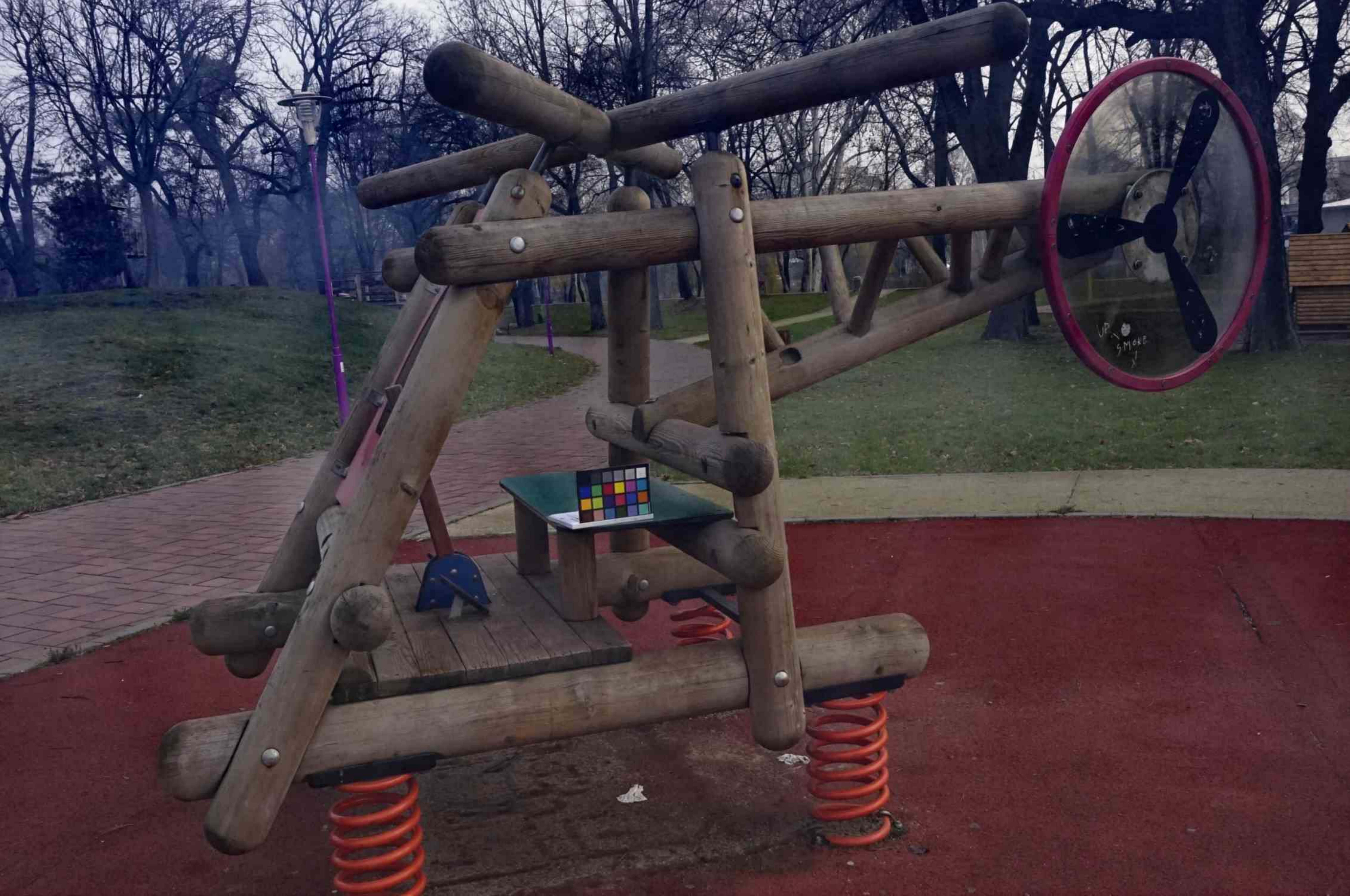}                 &
			\includegraphics[width = 0.16\textwidth]{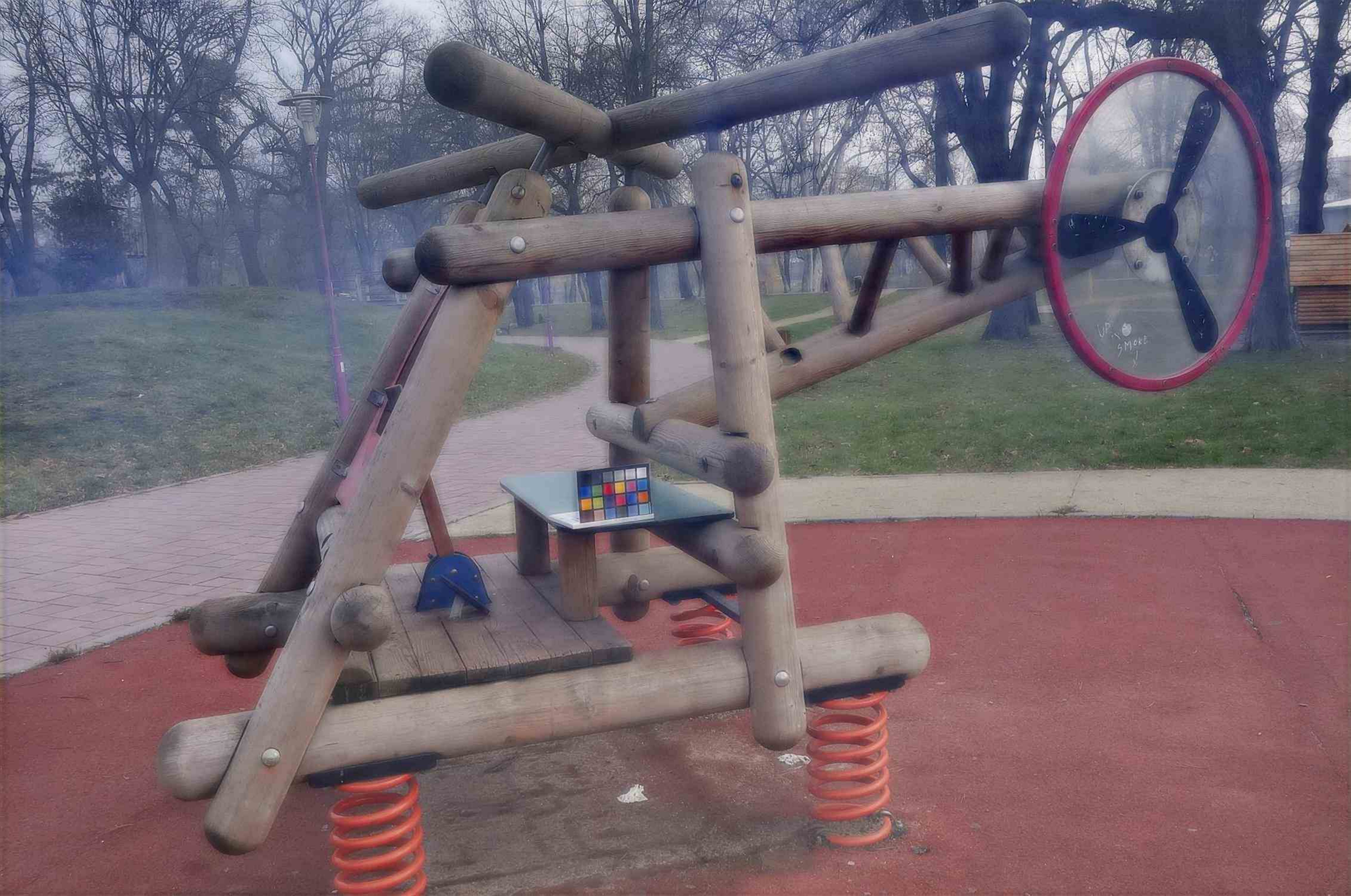}           \\
			
			\includegraphics[width = 0.16\textwidth]{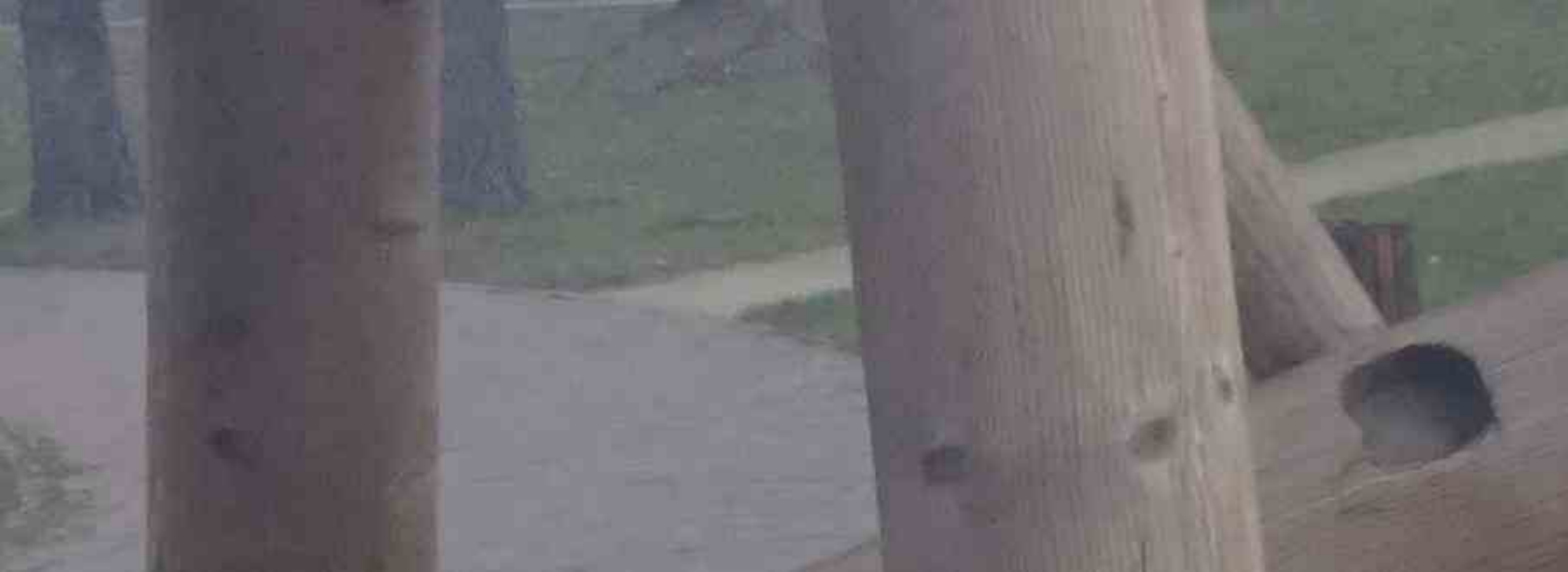}              &
			\includegraphics[width = 0.16\textwidth]{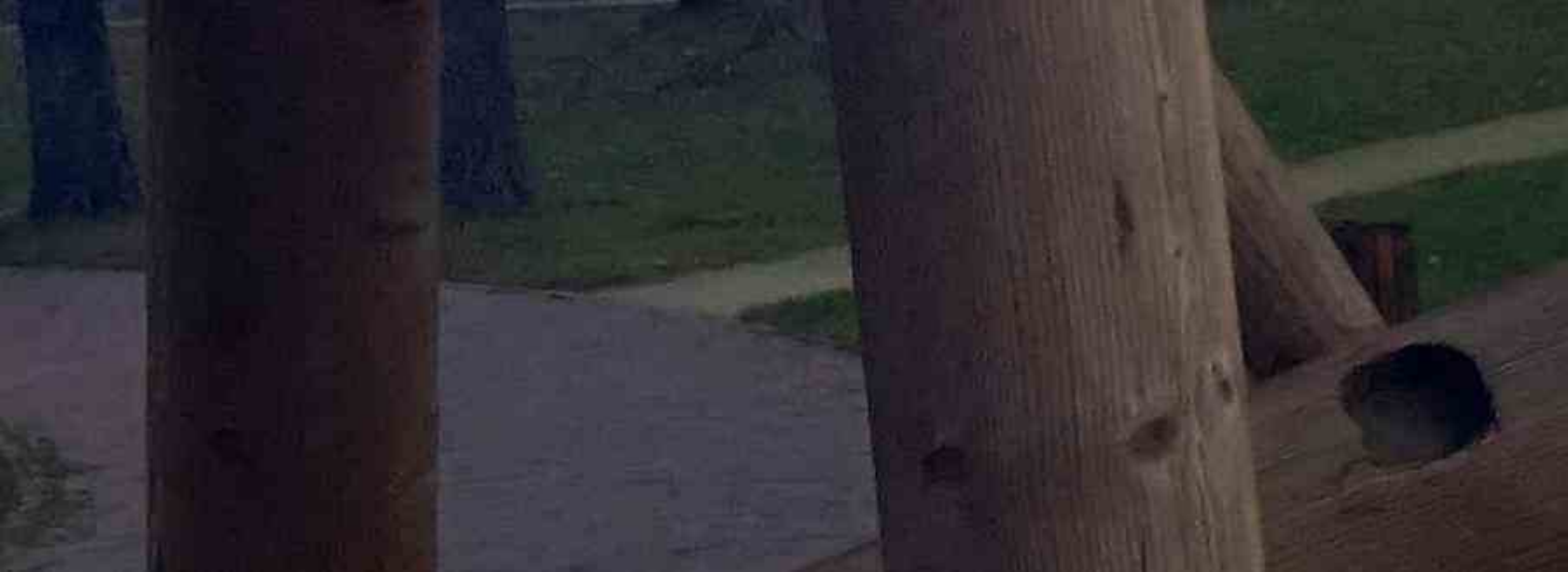}                 &
			\includegraphics[width = 0.16\textwidth]{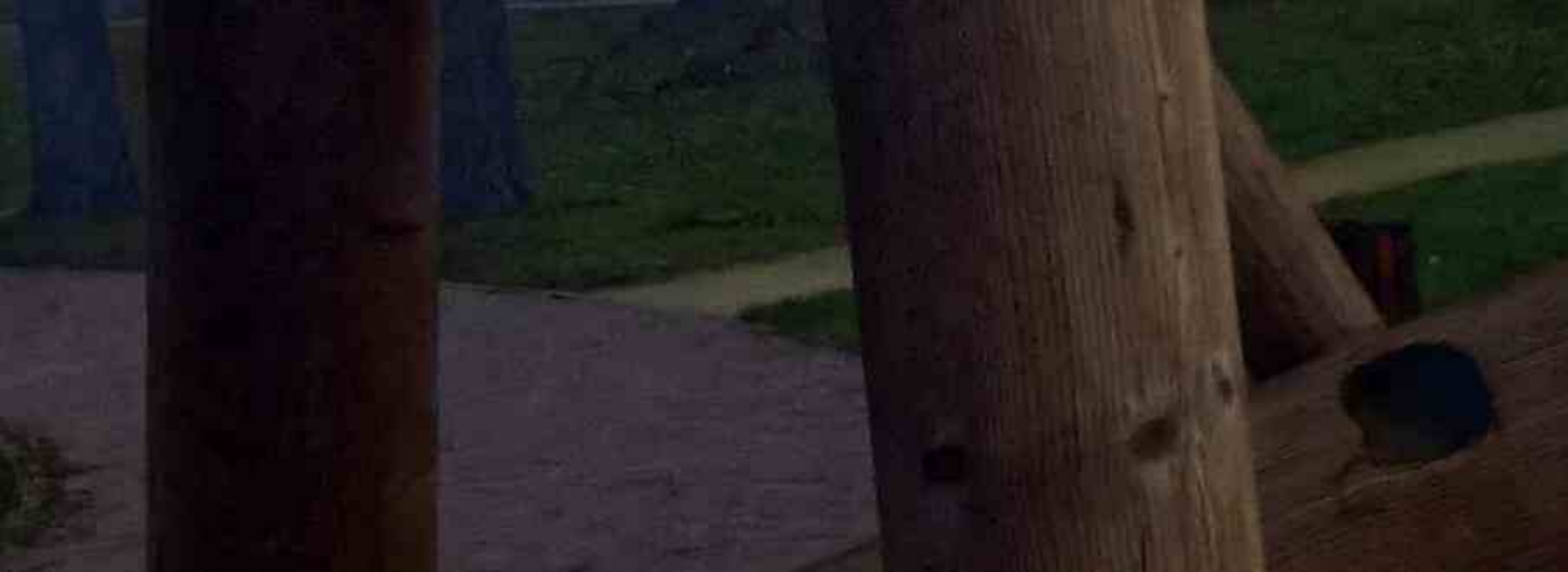}               &
			\includegraphics[width = 0.16\textwidth]{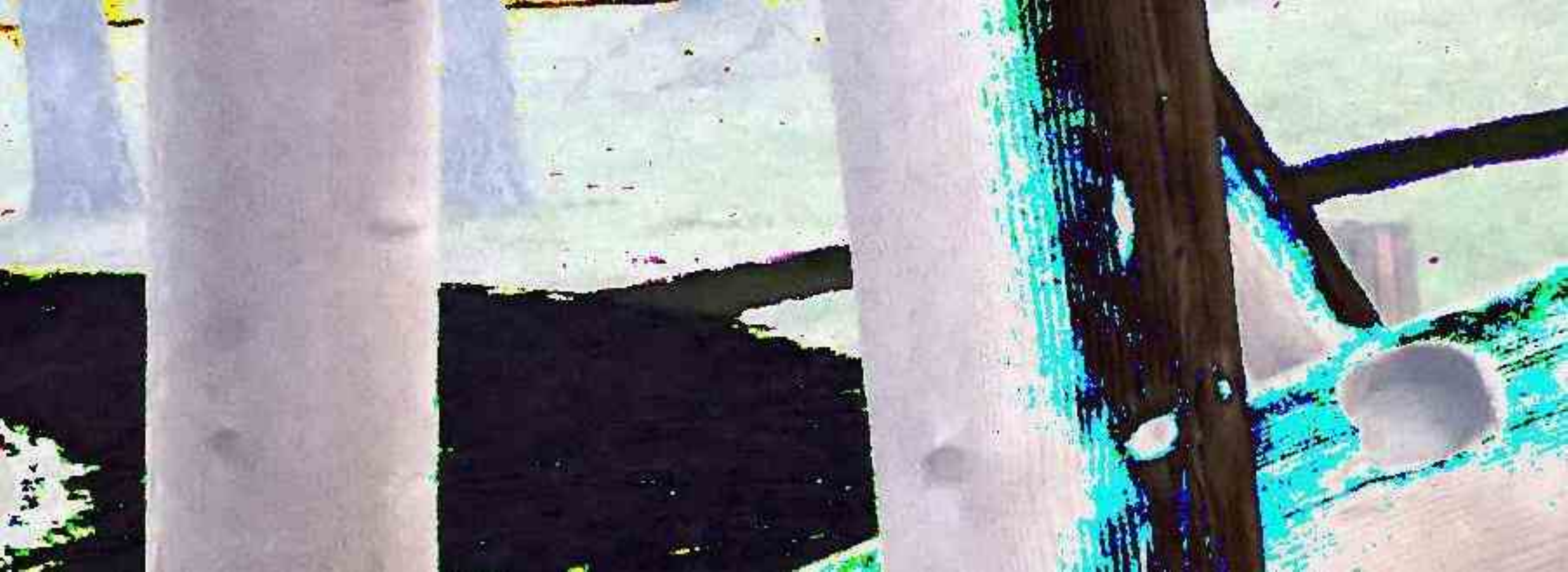}                &
			\includegraphics[width = 0.16\textwidth]{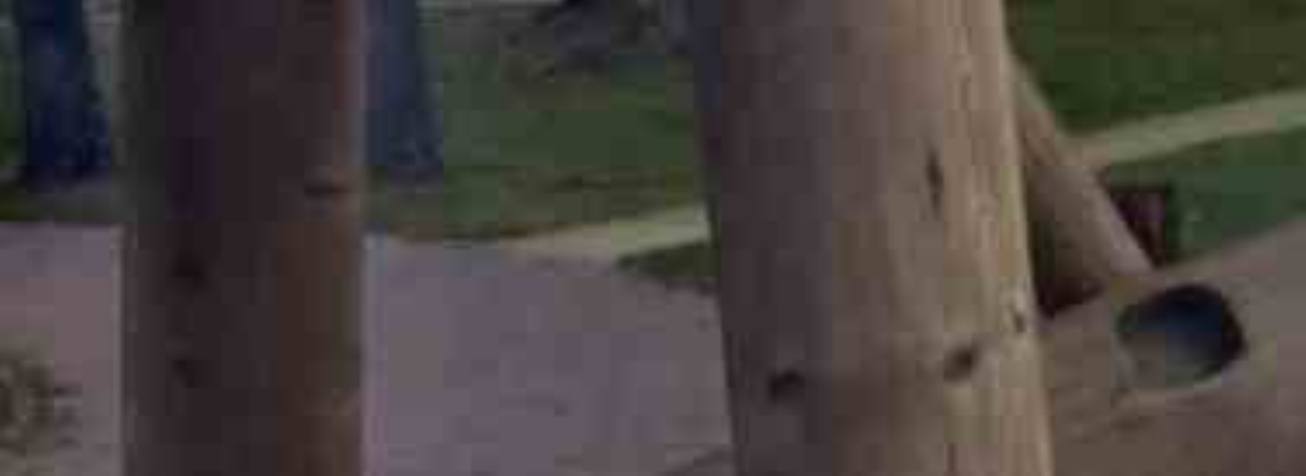}                 &
			\includegraphics[width = 0.16\textwidth]{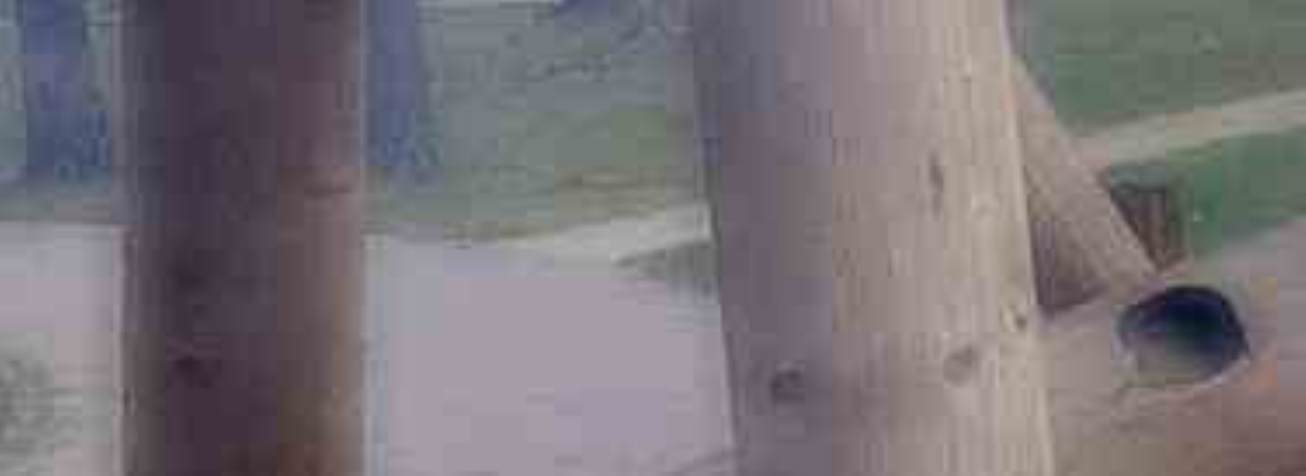}           \\
			
			Input (PSNR/SSIM)&
			AOD-NET~\cite{li2017all} (18.12/0.66)& 
			CAP~\cite{zhu2015fast} (14.87/0.6012)& 
			DehazeNet~\cite{cai2016dehazenet} (6.927/0.3247)& 
			GCA~\cite{chen2019gated} (22.05/0.83)& 
			MGBL~\cite{zheng2021ultra} (20.53/0.7768)\\

			\includegraphics[width = 0.16\textwidth]{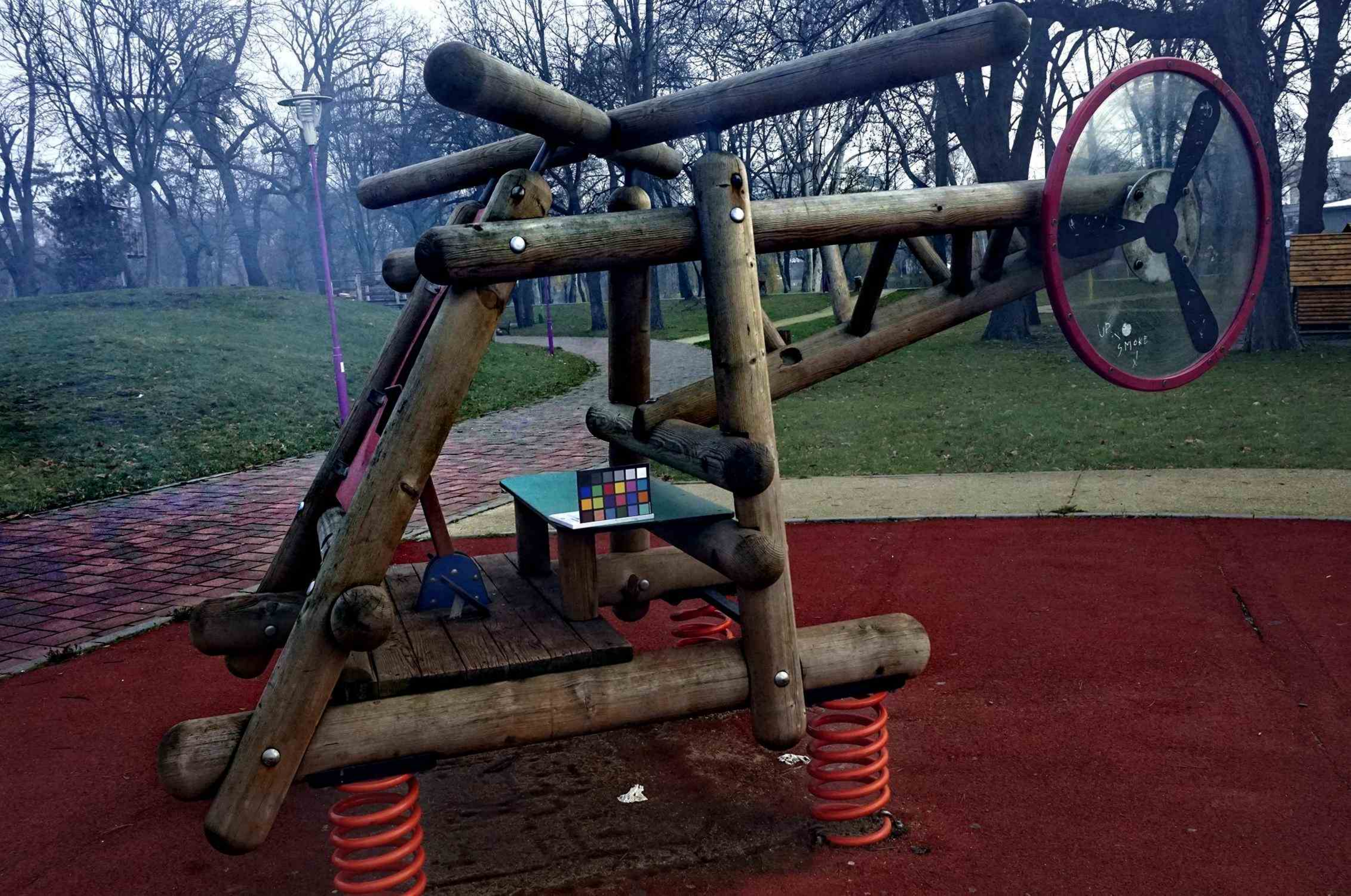}             & 
			\includegraphics[width = 0.16\textwidth]{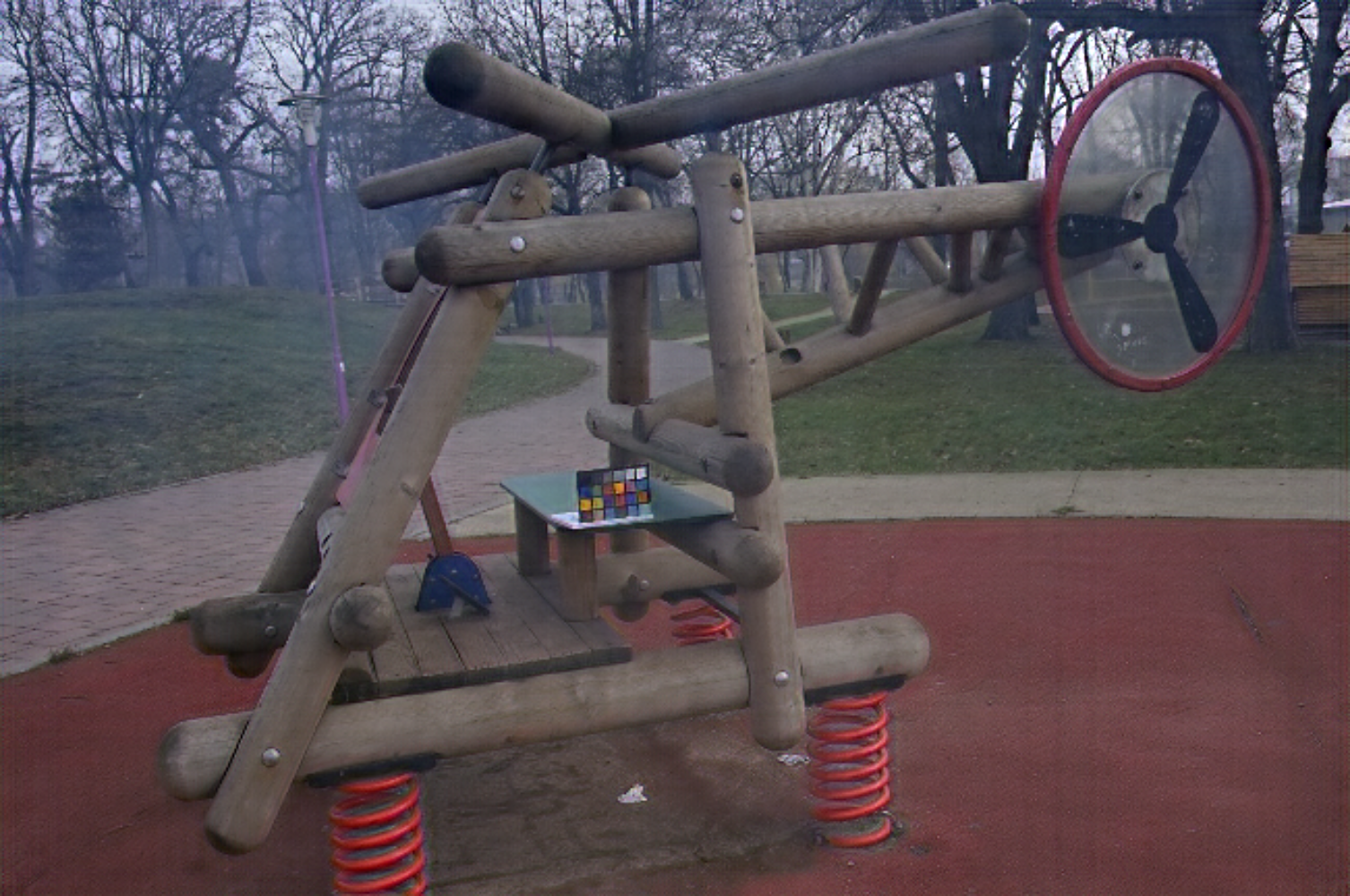}       &
			\includegraphics[width = 0.16\textwidth]{img0_DCP.pdf}                & 
			\includegraphics[width = 0.16\textwidth]{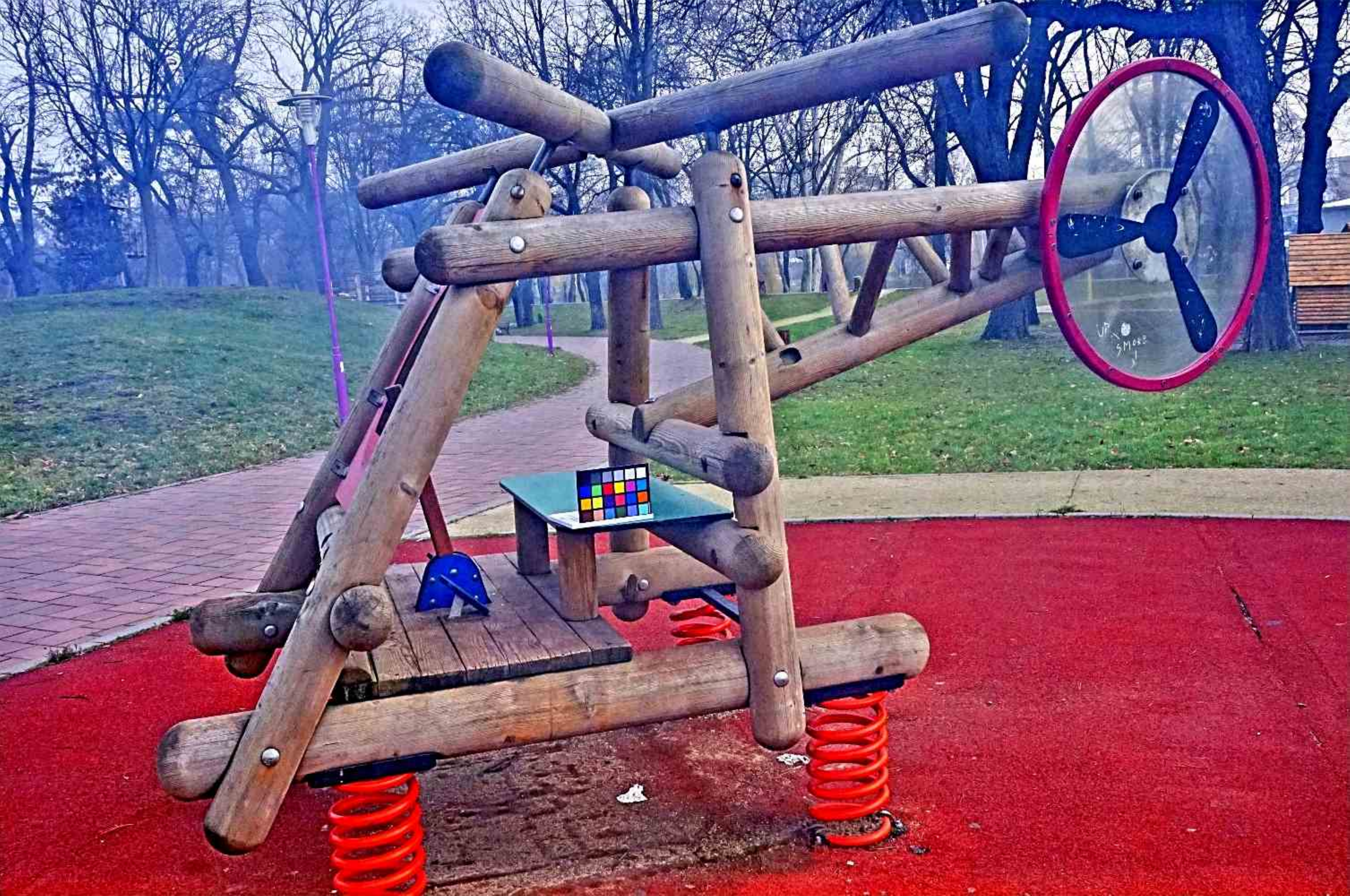}              &
			\includegraphics[width = 0.16\textwidth]{img0_ours.pdf}                 & 
			\includegraphics[width = 0.16\textwidth]{img0_GT.pdf}                     \\       
			
			\includegraphics[width = 0.16\textwidth]{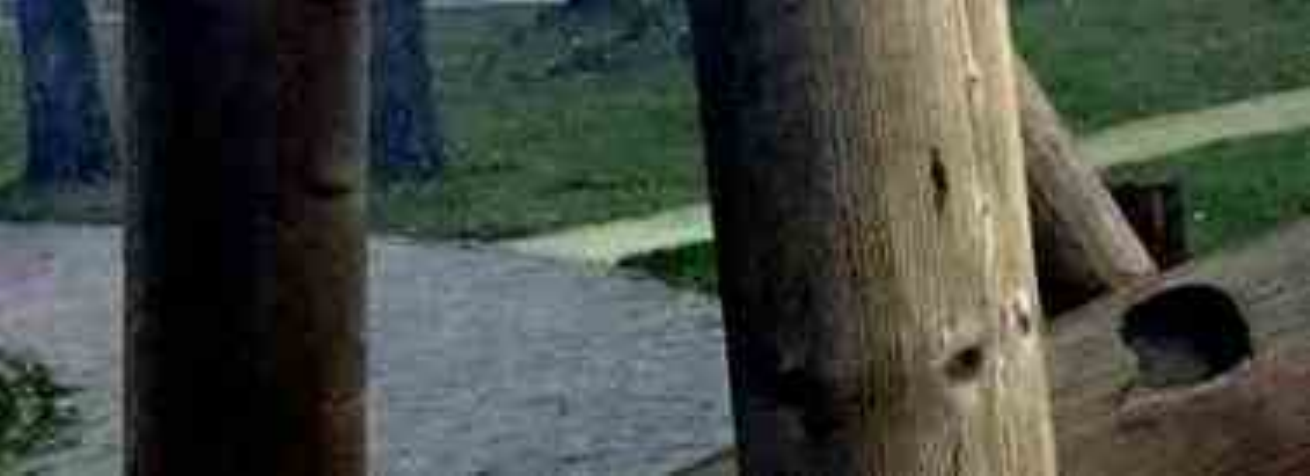}             & 
			\includegraphics[width = 0.16\textwidth]{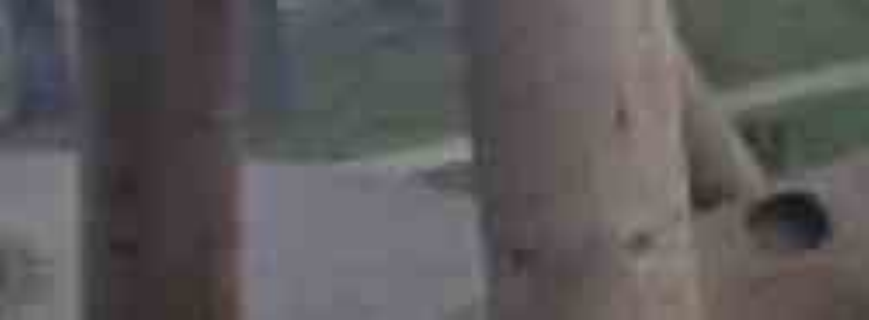}       &
			\includegraphics[width = 0.16\textwidth]{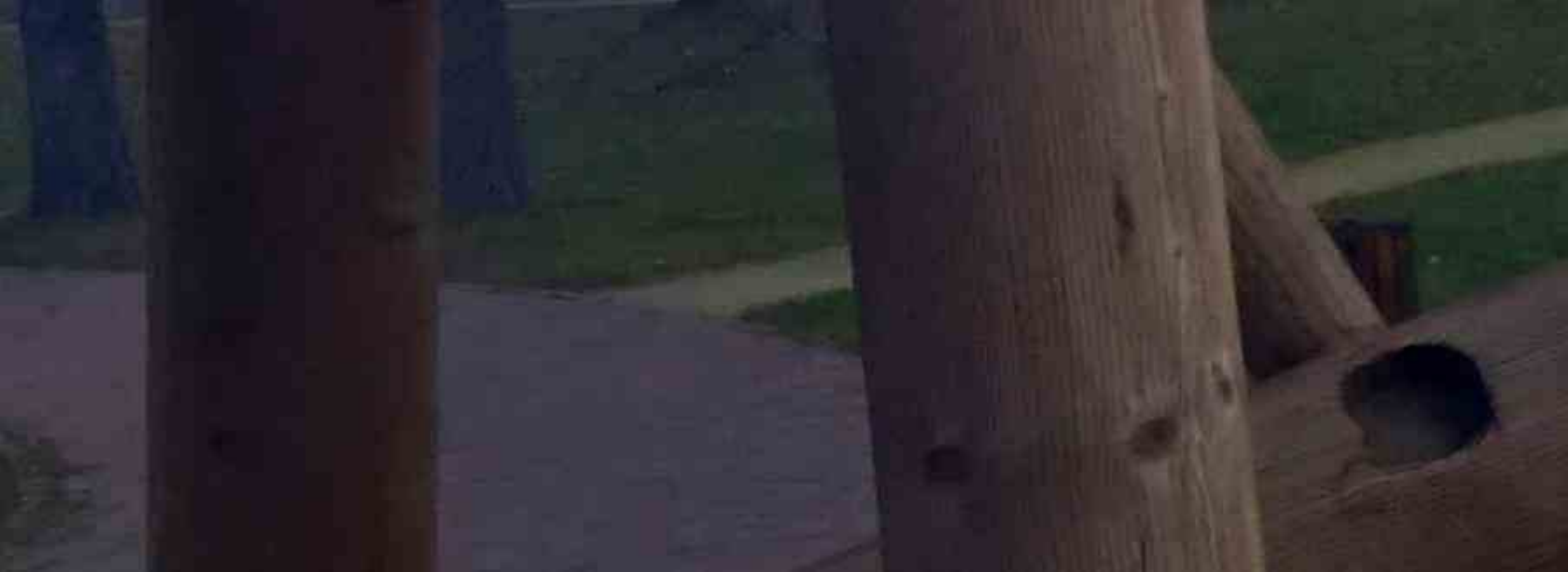}                & 
			\includegraphics[width = 0.16\textwidth]{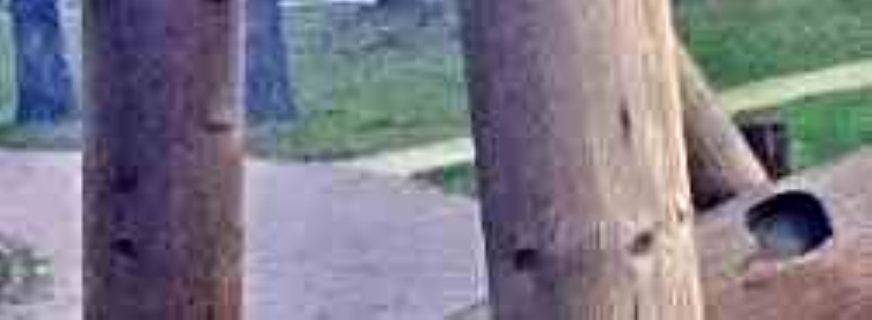}              &
			\includegraphics[width = 0.16\textwidth]{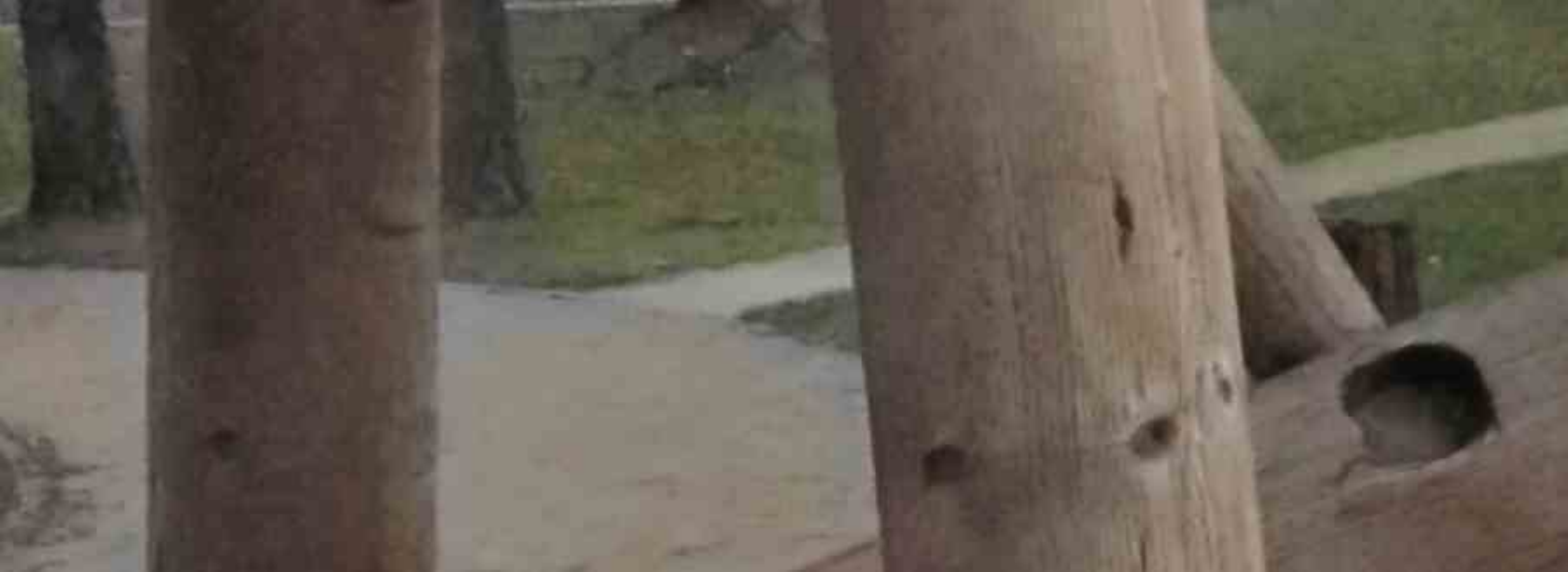}                 & 
			\includegraphics[width = 0.16\textwidth]{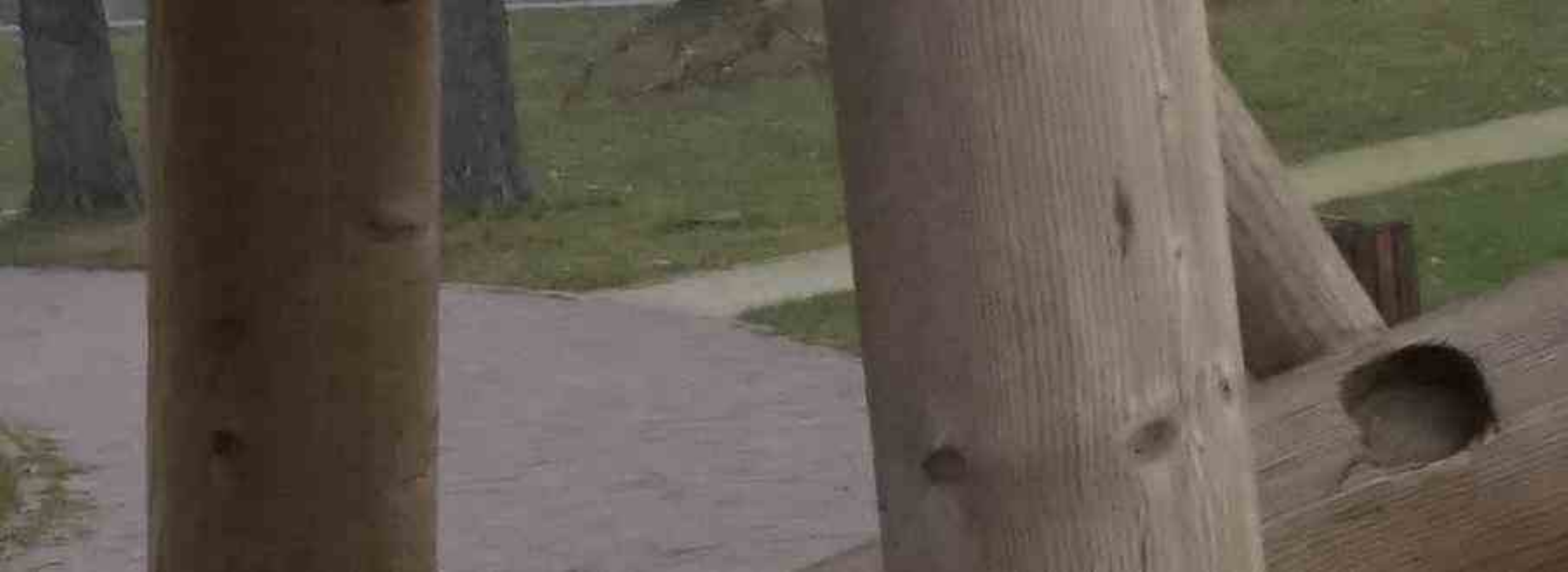}                     \\      
			NLD~\cite{berman2016non} (18.44/0.66)&
			PFF~\cite{mei2018progressive} (22.93/0.72)& 
			DCP~\cite{he2010single} (18.91/0.7392)& 
			PSD~\cite{chen2021psd} (13.32/0.4697)& 
			\textbf{Ours (25.62/0.8582)}& 
			GT ($+ \infty$/1)
			\\
		\end{tabular}
	\end{center}
	
	\caption{Our method obtains better visual quality and recovers more image details compared with other state-of-the-art methods in the O-HAZE dataset.}
	\label{O-HAZE}
	\vspace{-2mm}
\end{figure*}

\begin{figure*}[!hpt]\tiny
	\begin{center}
		\tabcolsep 1pt
		\begin{tabular}{@{}cccccc@{}}
			\includegraphics[width = 0.16\textwidth]{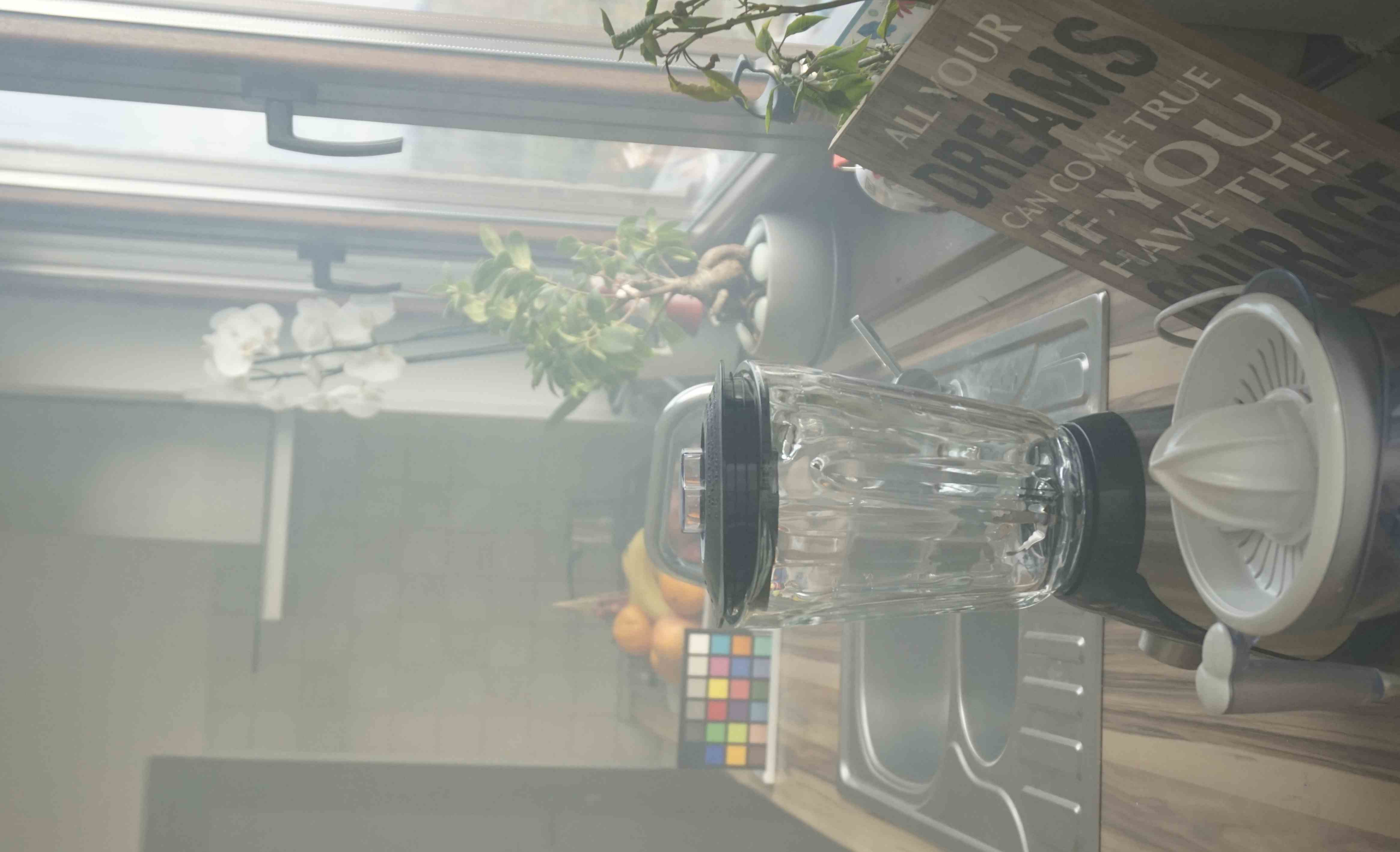}              &
			\includegraphics[width = 0.16\textwidth]{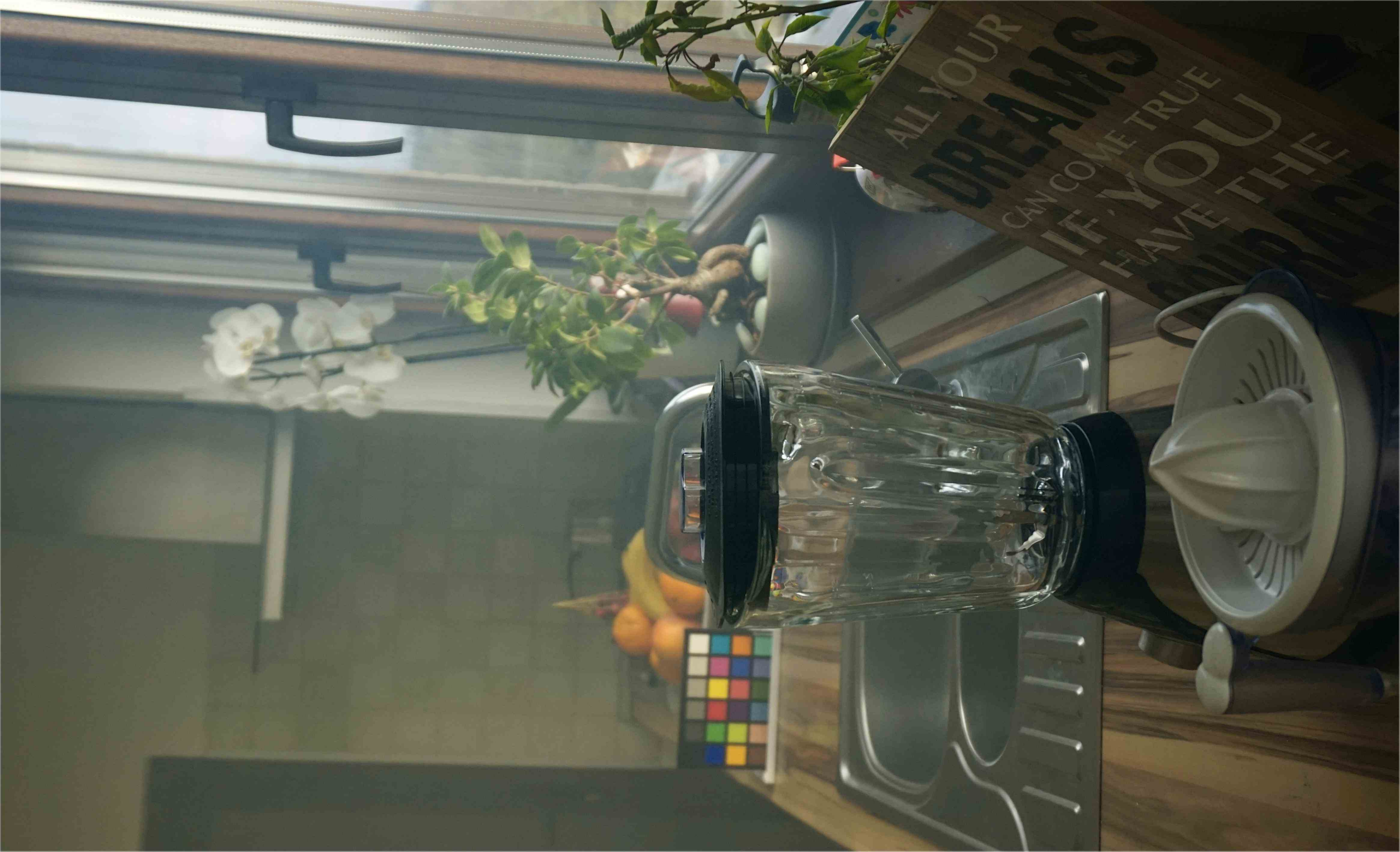}                 &
			\includegraphics[width = 0.16\textwidth]{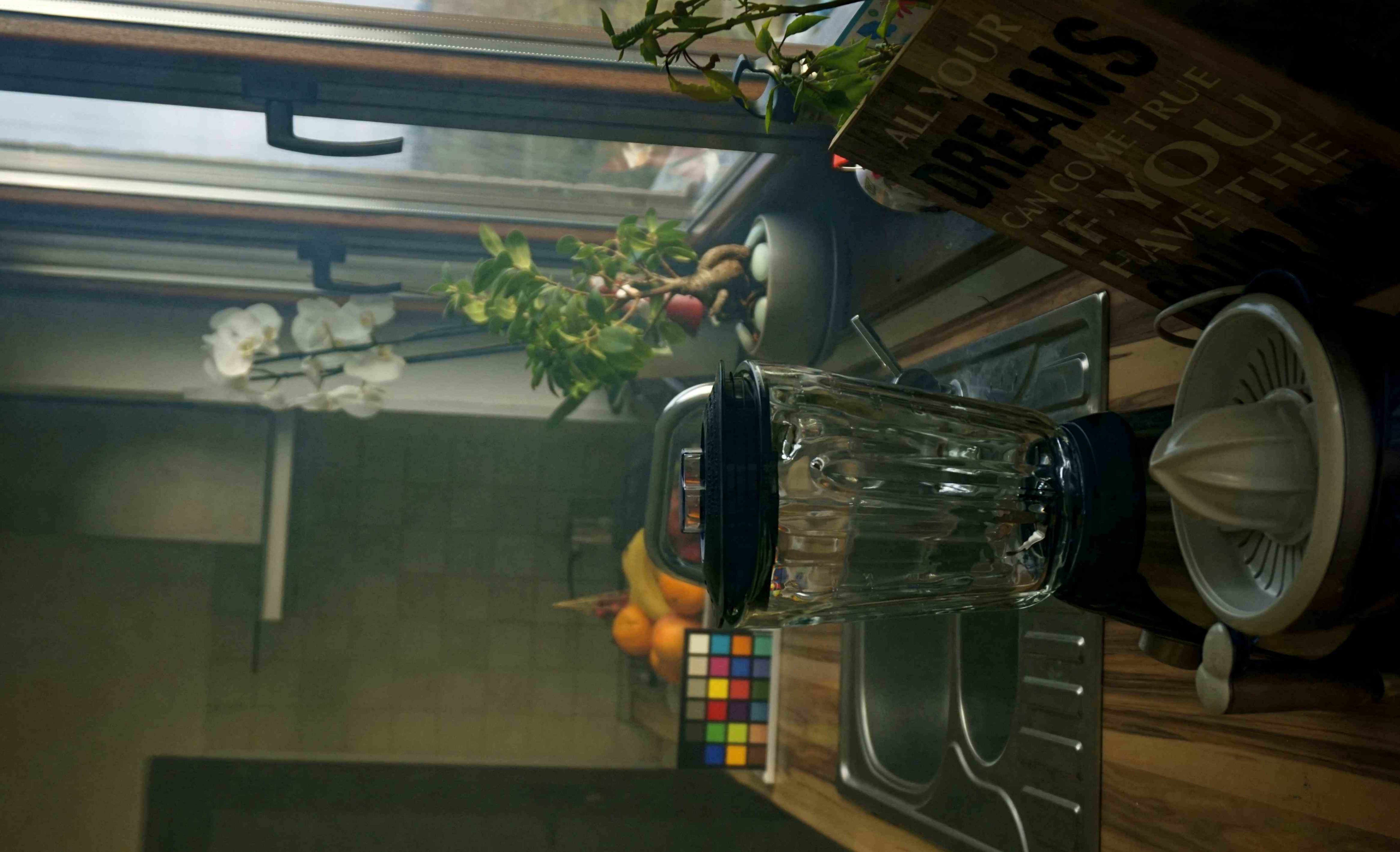}               &
			\includegraphics[width = 0.16\textwidth]{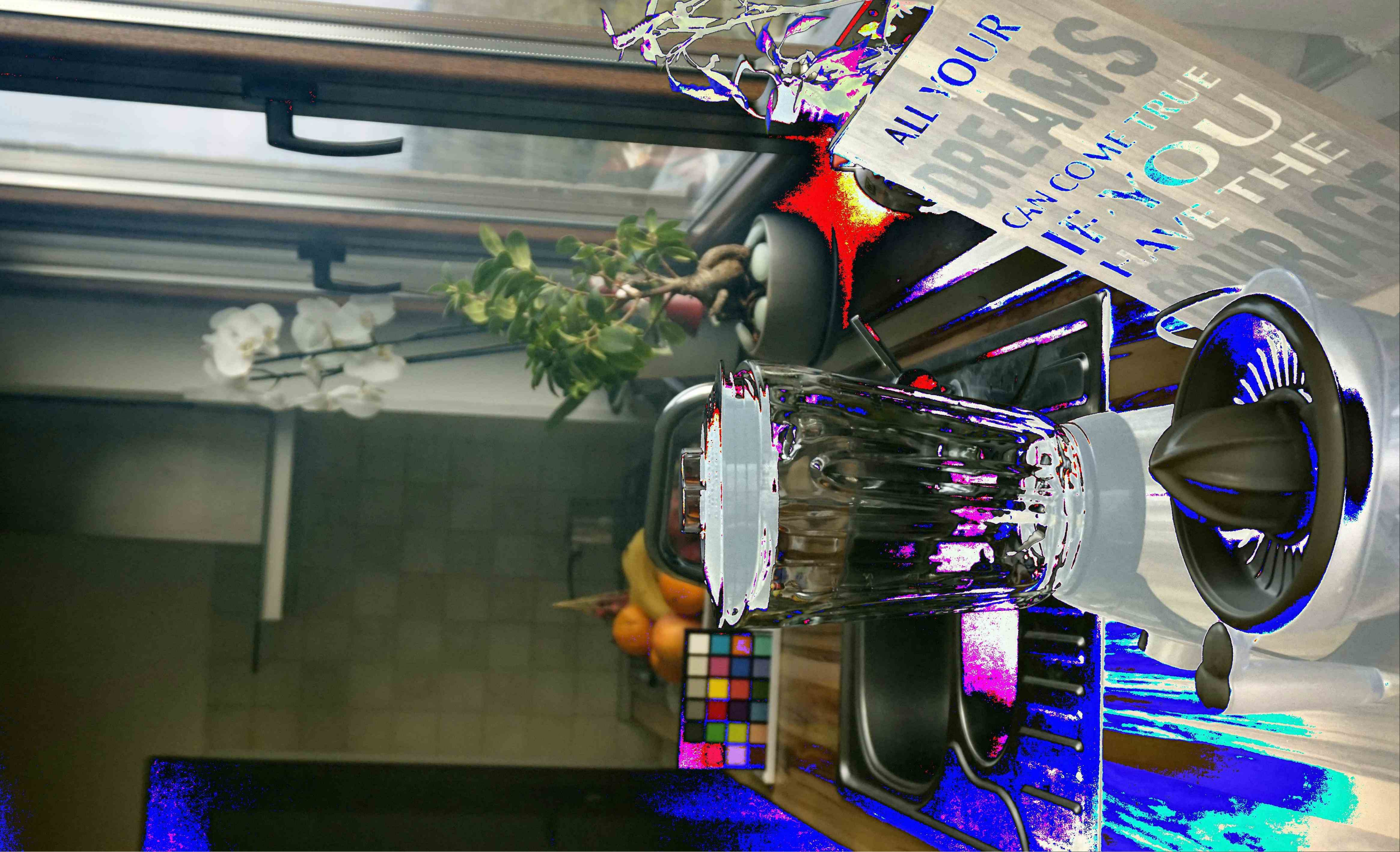}                &
			\includegraphics[width = 0.16\textwidth]{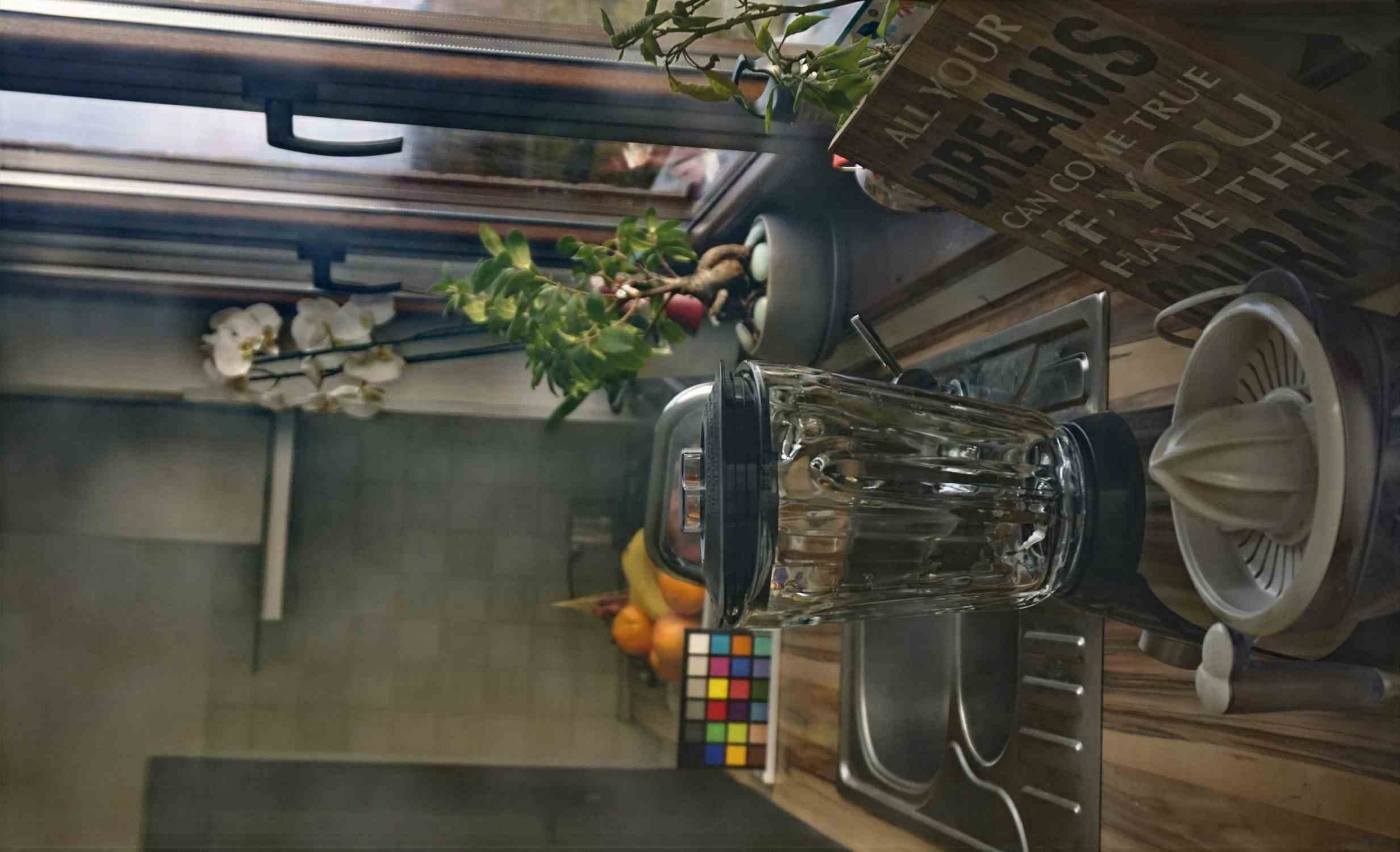}                 &
			\includegraphics[width = 0.16\textwidth]{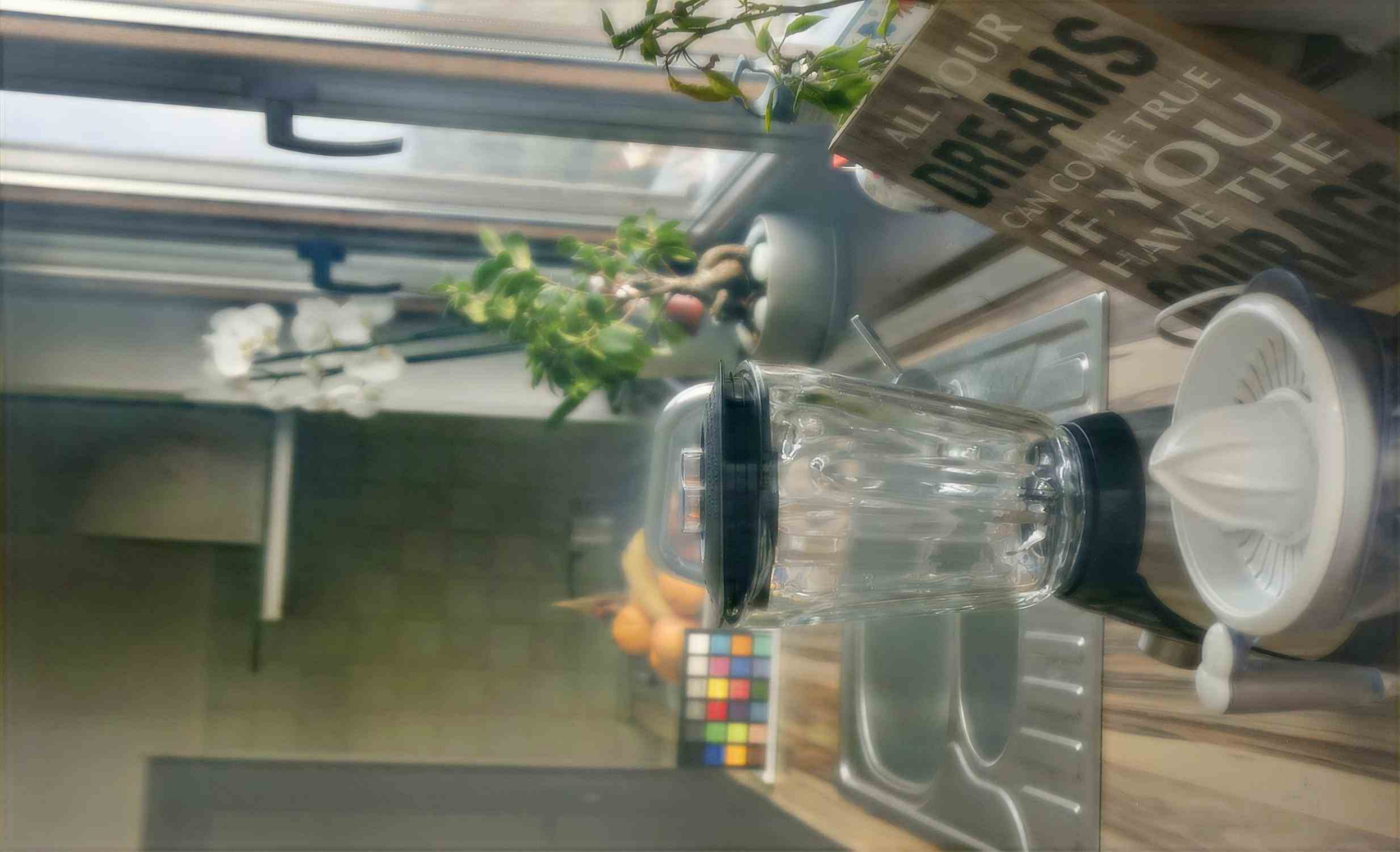}           \\
			
			\includegraphics[width = 0.16\textwidth]{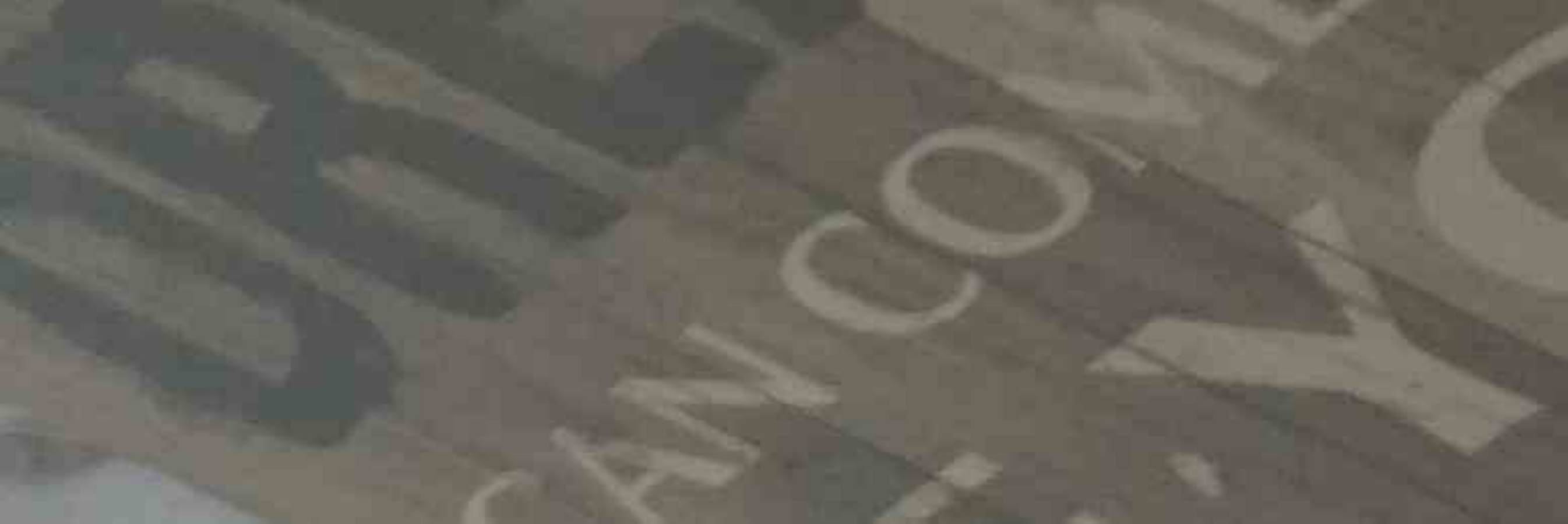}              &
			\includegraphics[width = 0.16\textwidth]{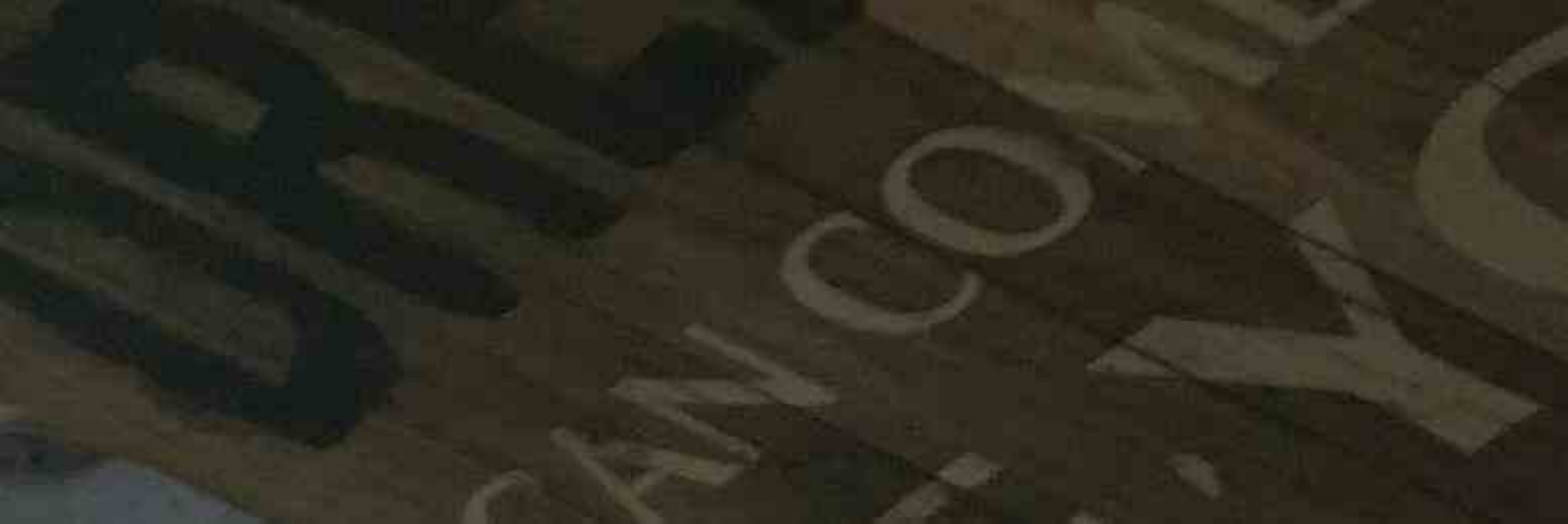}                 &
			\includegraphics[width = 0.16\textwidth]{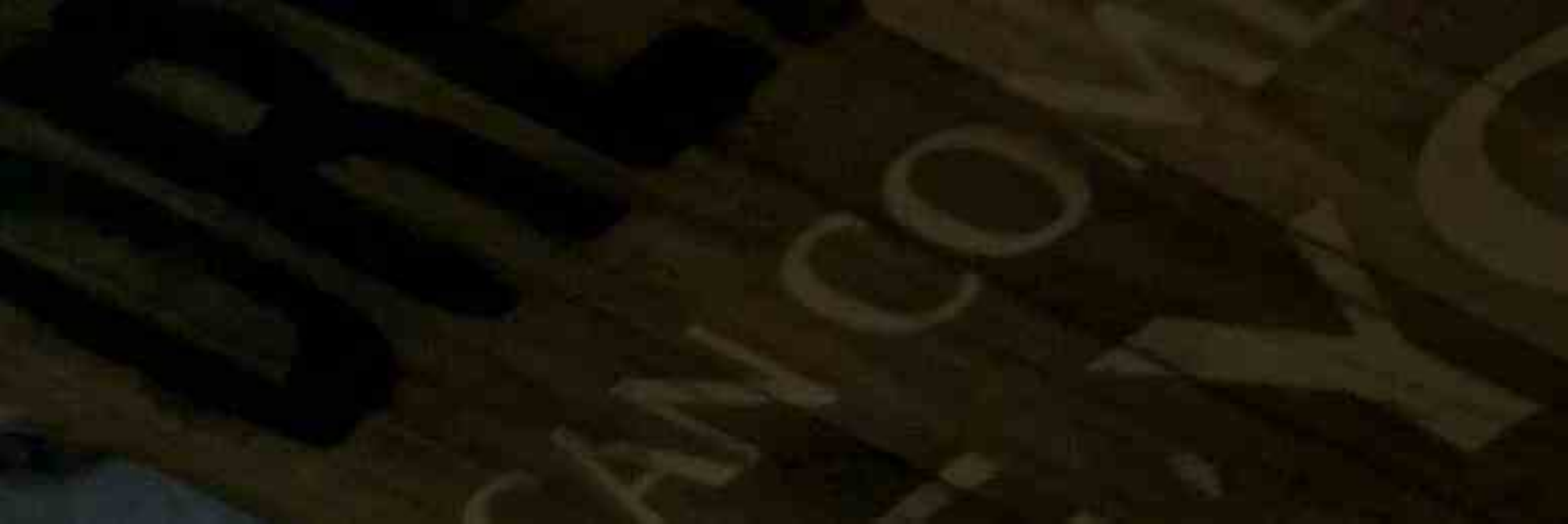}               &
			\includegraphics[width = 0.16\textwidth]{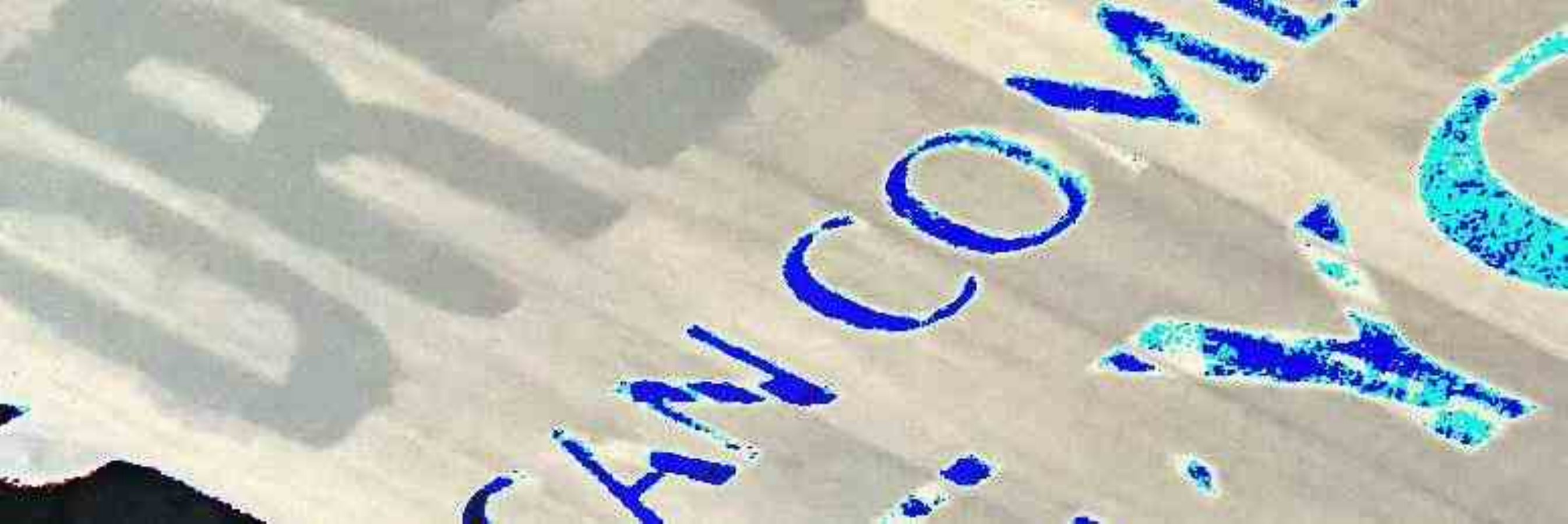}                &
			\includegraphics[width = 0.16\textwidth]{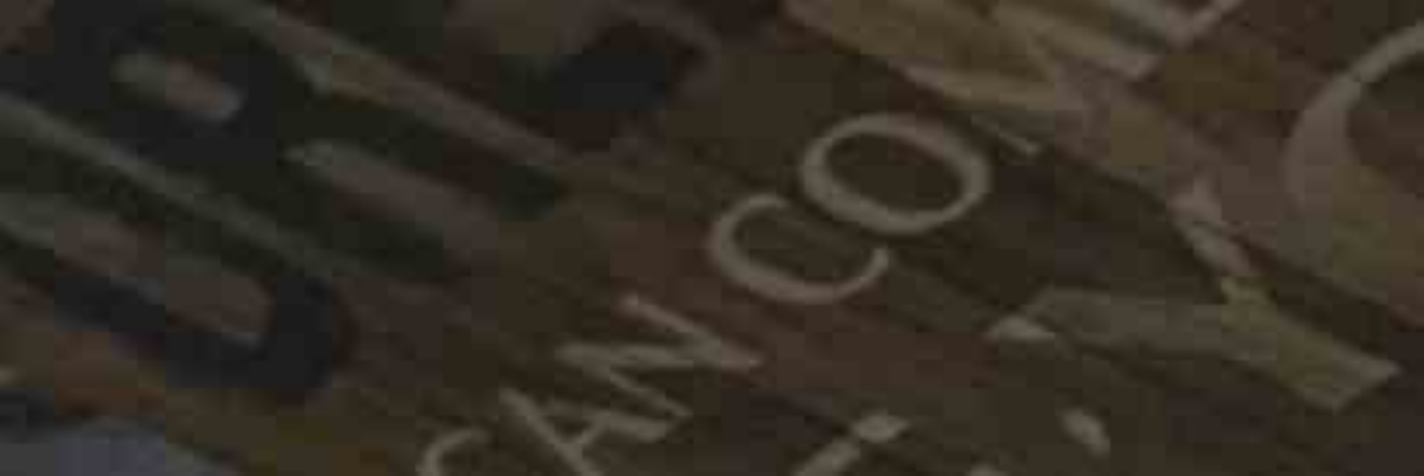}                 &
			\includegraphics[width = 0.16\textwidth]{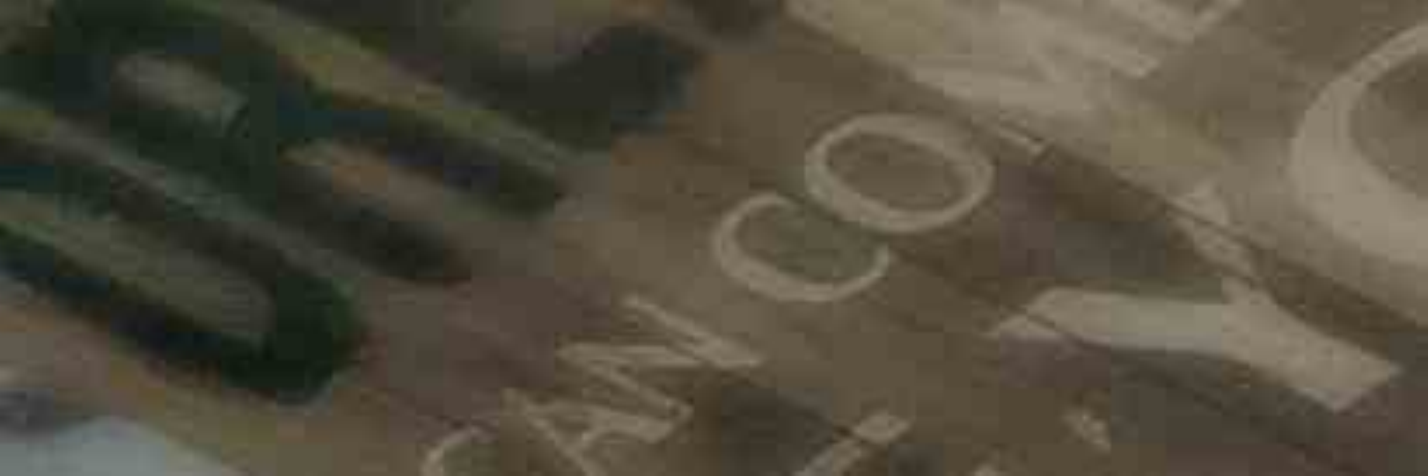}           \\
			Input (PSNR/SSIM)&
			AOD-NET~\cite{li2017all} (17.12/0.78)& 
			CAP~\cite{zhu2015fast} (13.87/0.6512)& 
			DehazeNet~\cite{cai2016dehazenet} (9.87/0.5647)& 
			GCA~\cite{chen2019gated} (15.21/0.78)& 
			MGBL~\cite{zheng2021ultra} (19.77/0.8275)\\
			
			\includegraphics[width = 0.16\textwidth]{img_non-local.pdf}             & 
			\includegraphics[width = 0.16\textwidth]{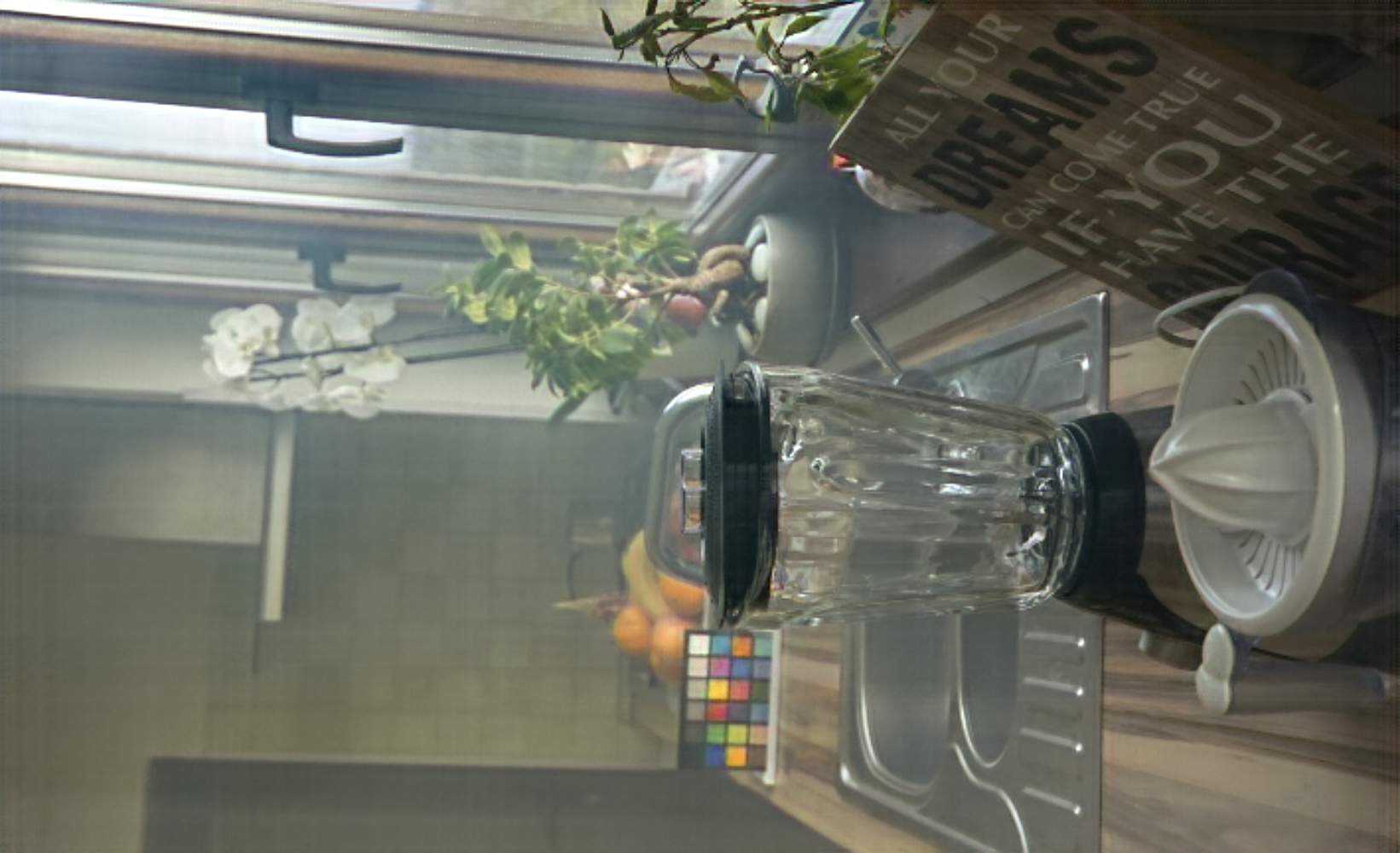}       &
			\includegraphics[width = 0.16\textwidth]{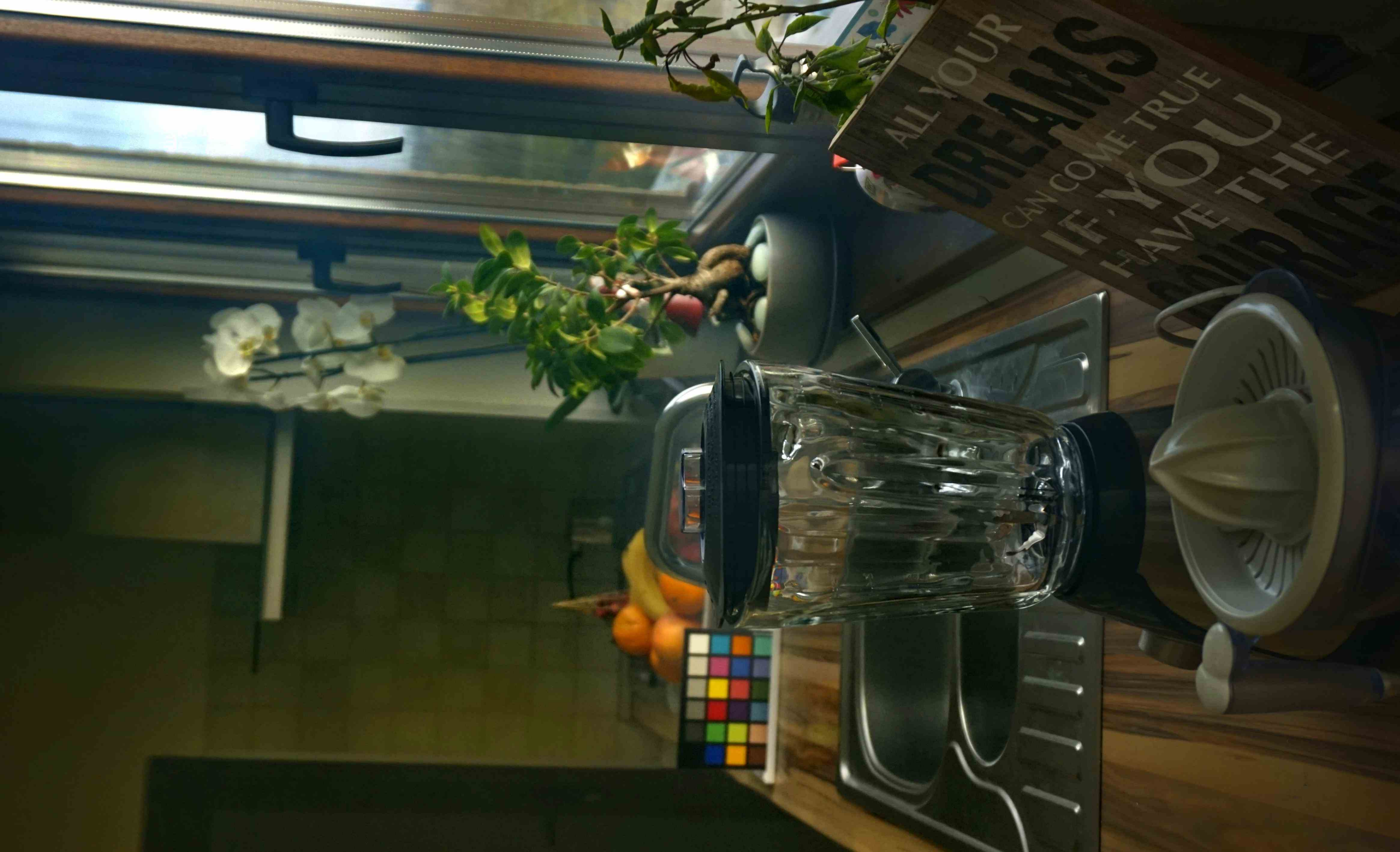}                & 
			\includegraphics[width = 0.16\textwidth]{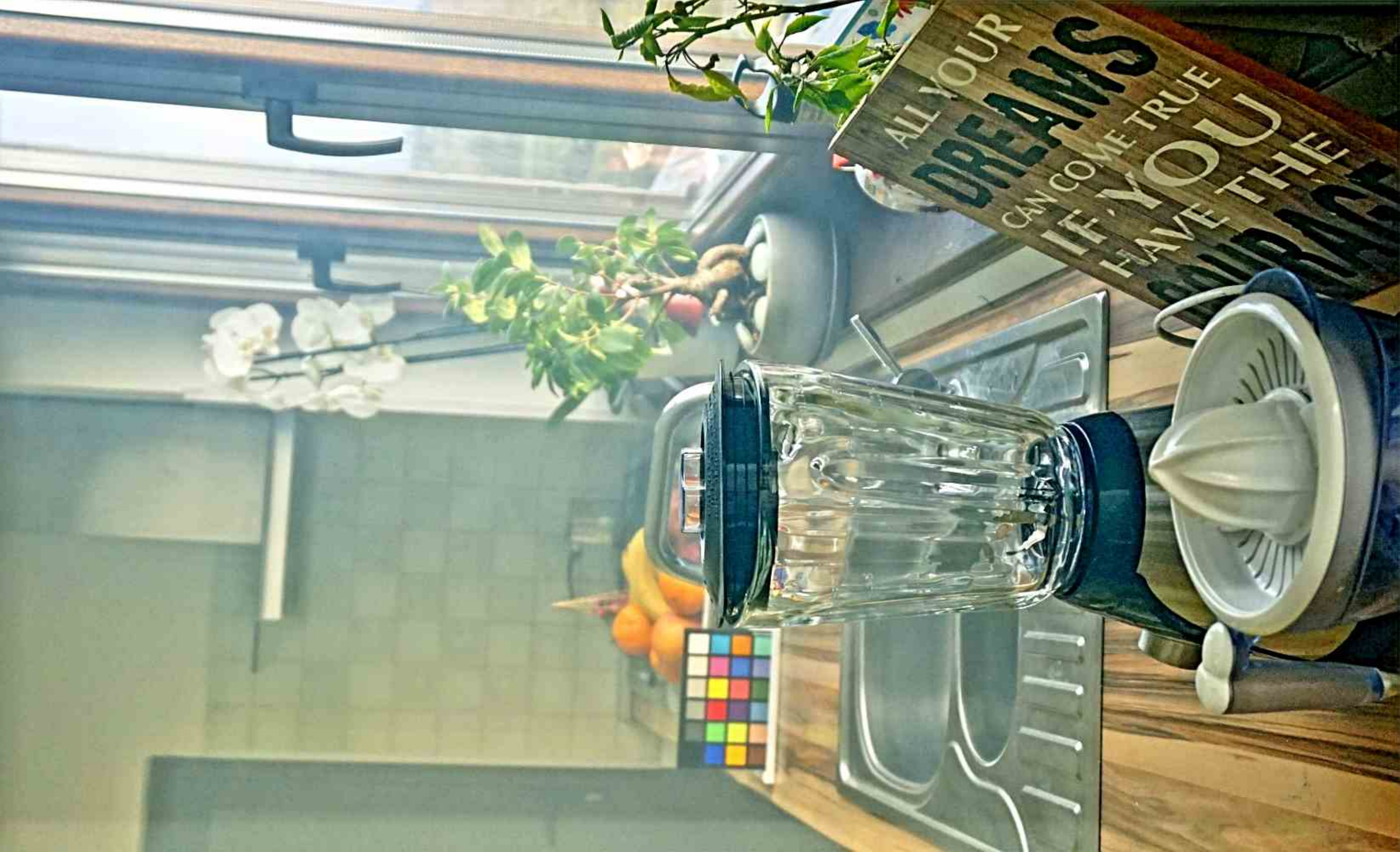}              &
			\includegraphics[width = 0.16\textwidth]{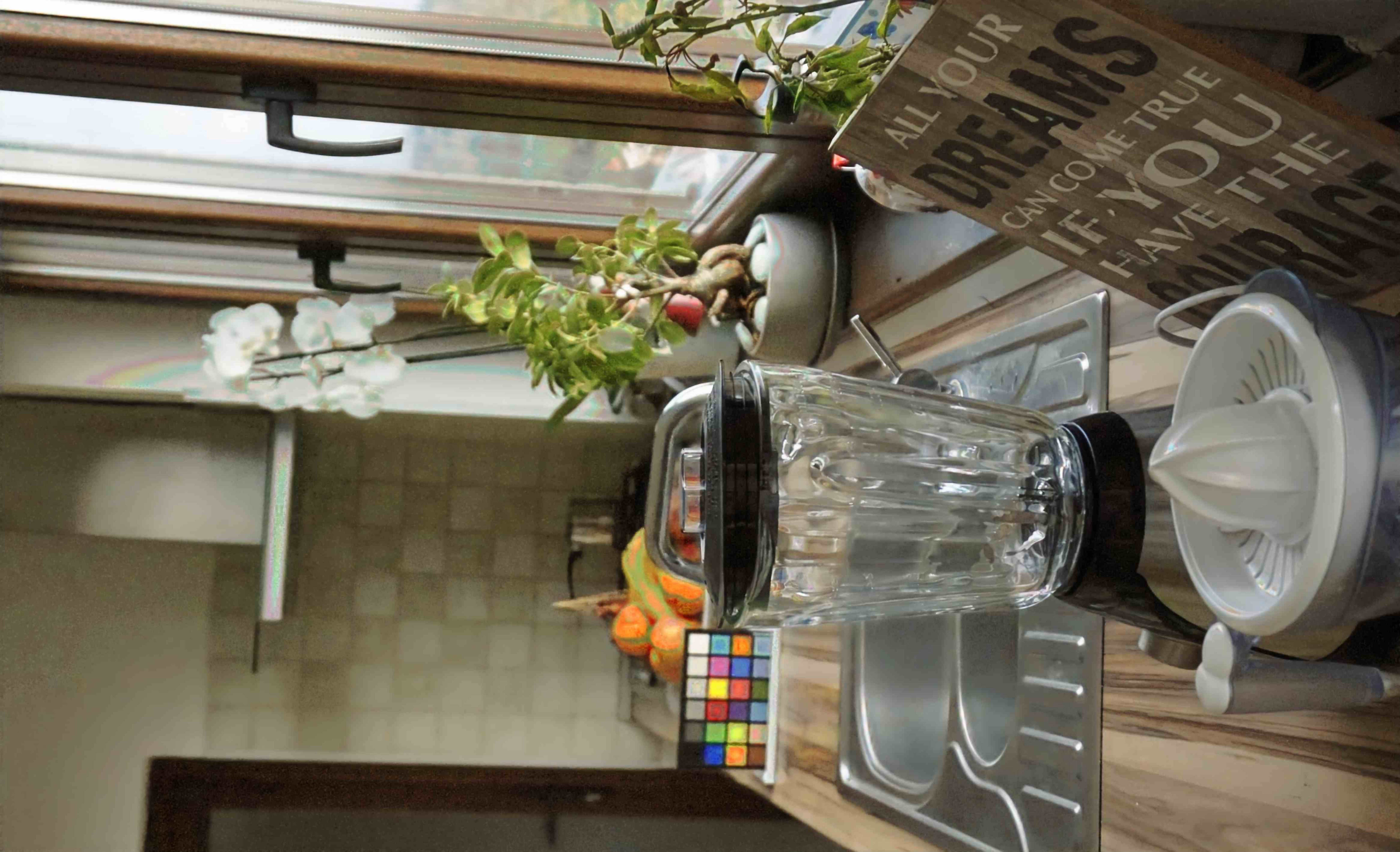}                 & 
			\includegraphics[width = 0.16\textwidth]{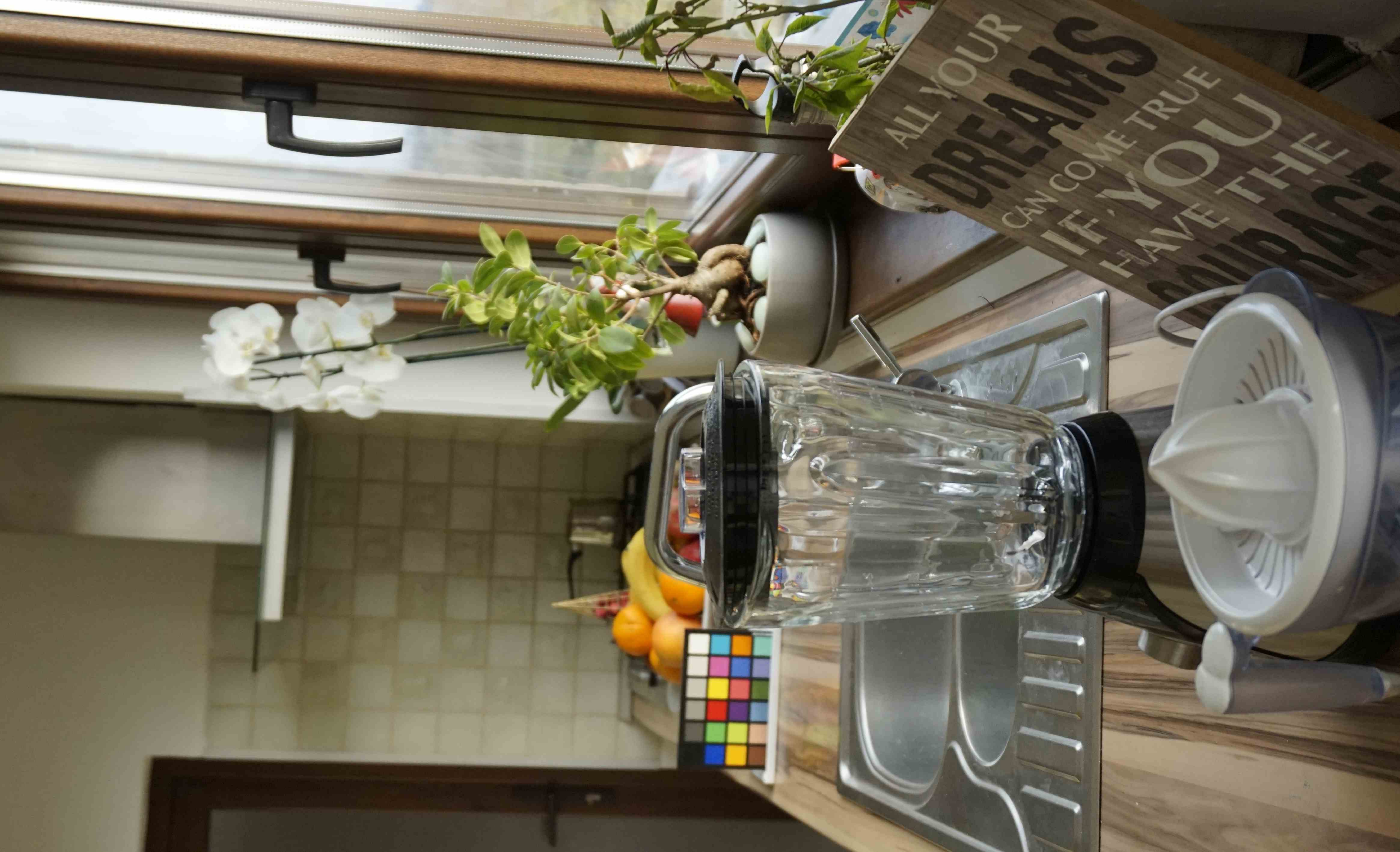}                     \\        
			
			\includegraphics[width = 0.16\textwidth]{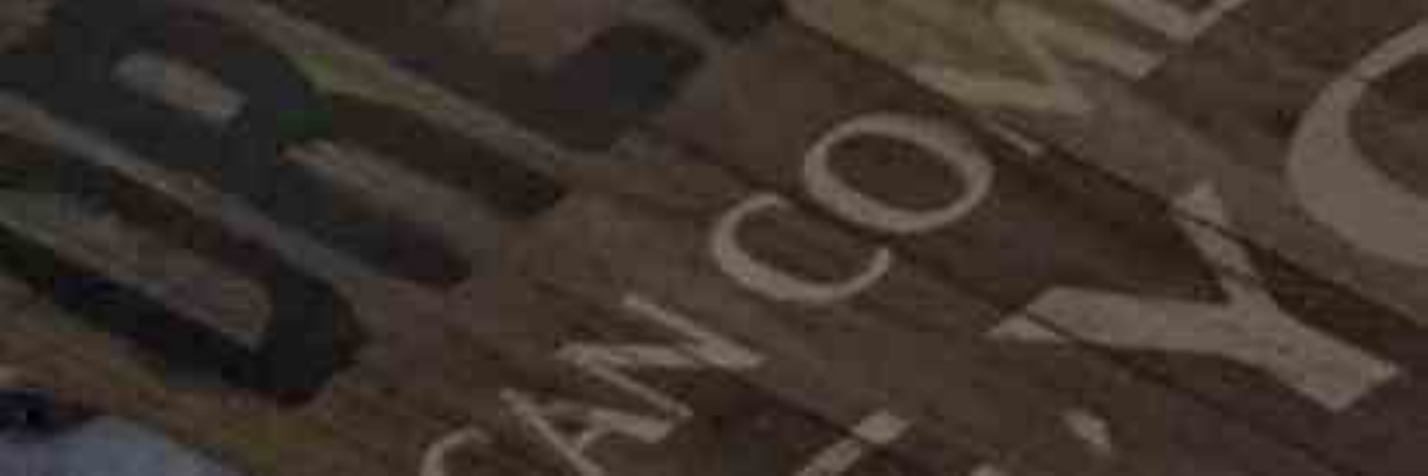}             & 
			\includegraphics[width = 0.16\textwidth]{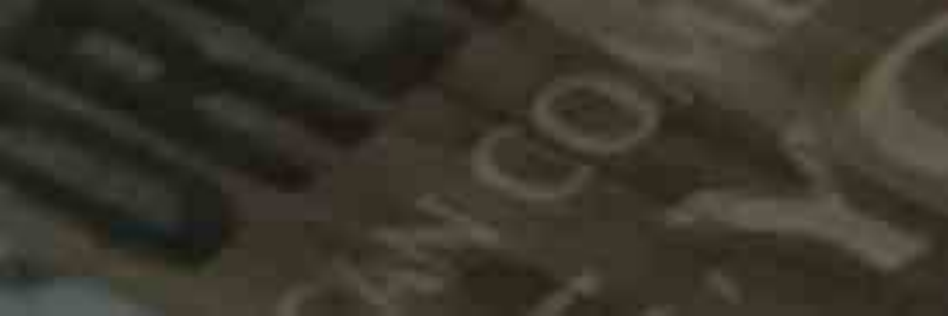}       &
			\includegraphics[width = 0.16\textwidth]{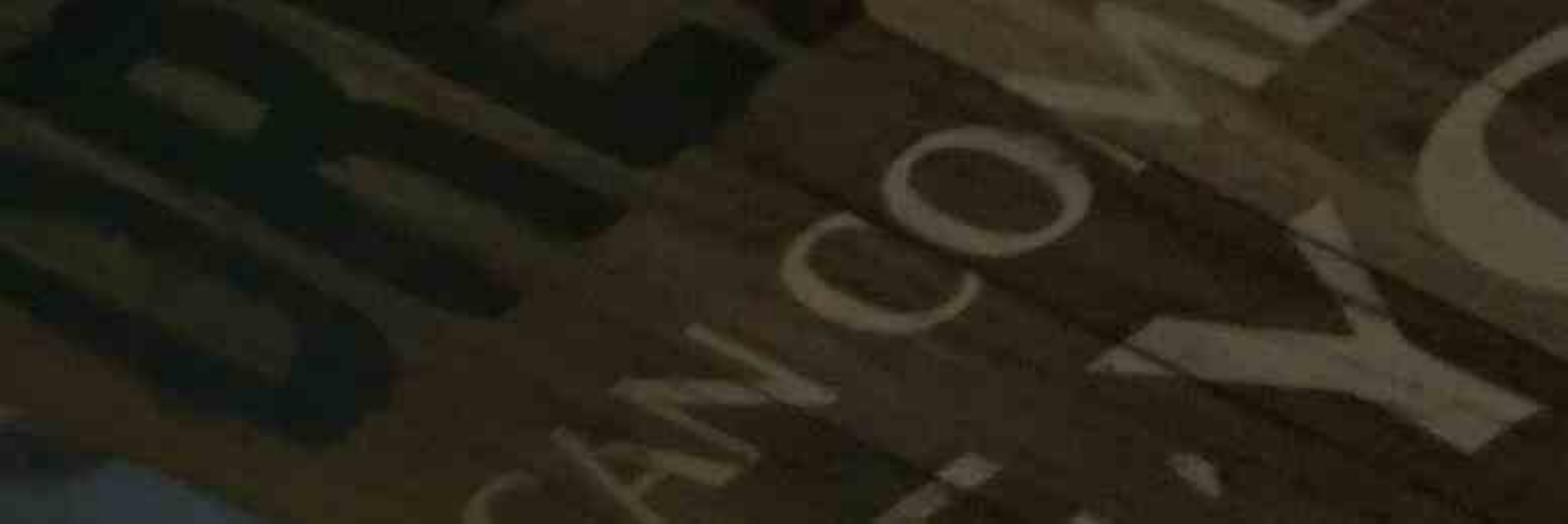}                & 
			\includegraphics[width = 0.16\textwidth]{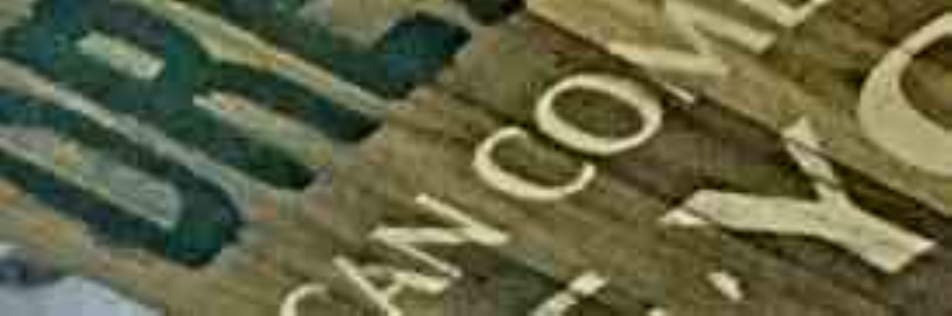}              &
			\includegraphics[width = 0.16\textwidth]{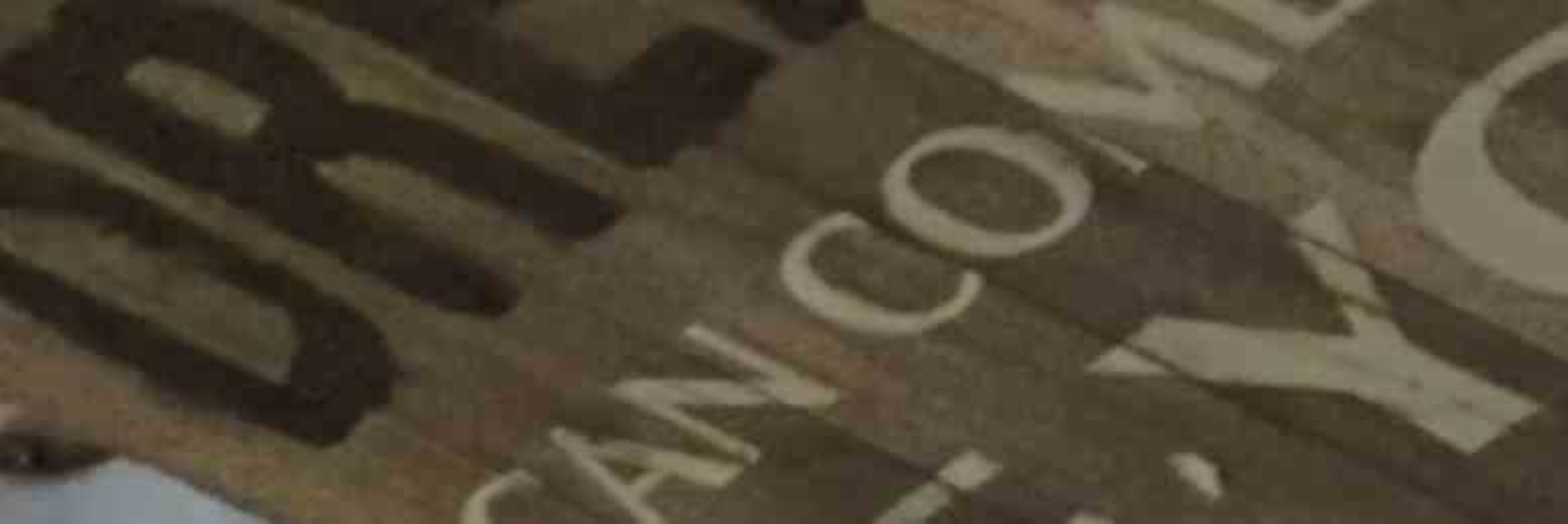}                 & 
			\includegraphics[width = 0.16\textwidth]{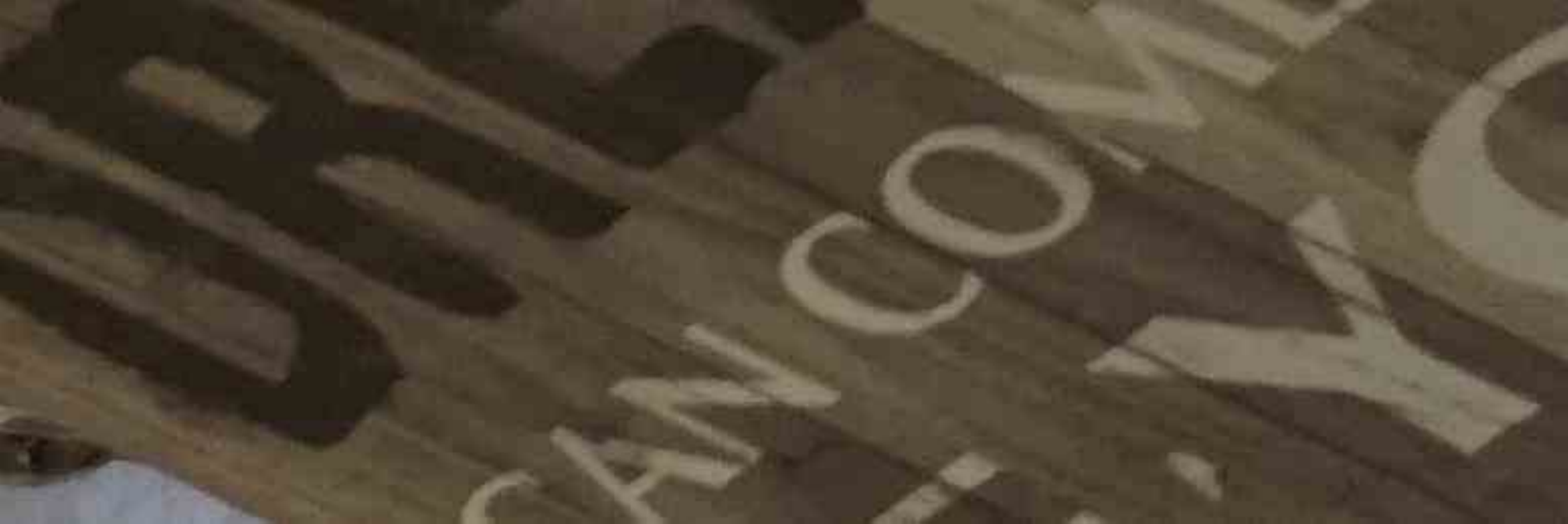}                     \\   
			
			NLD~\cite{berman2016non} (15.98/0.75)&
			PFF~\cite{mei2018progressive} (15.99/0.7377)& 
			DCP~\cite{he2010single} (12.75/0.6452)& 
			PSD~\cite{chen2021psd} (13.47/0.6697)& 
			\textbf{Ours  (25.62/0.8582)}& 
			GT ($+ \infty$/1)
			\\
			
		\end{tabular}
	\end{center}
	
	\caption{Our method obtains better visual quality and recovers more image details compared with other state-of-the-art methods in the I-HAZE dataset.}
	\vspace{-2mm}
	\label{IHAZE}
\end{figure*}

\begin{figure*}[h]\scriptsize
	\begin{center}
		\tabcolsep 1pt
		\begin{tabular}{@{}ccccc@{}}
			\includegraphics[width = 0.2\textwidth]{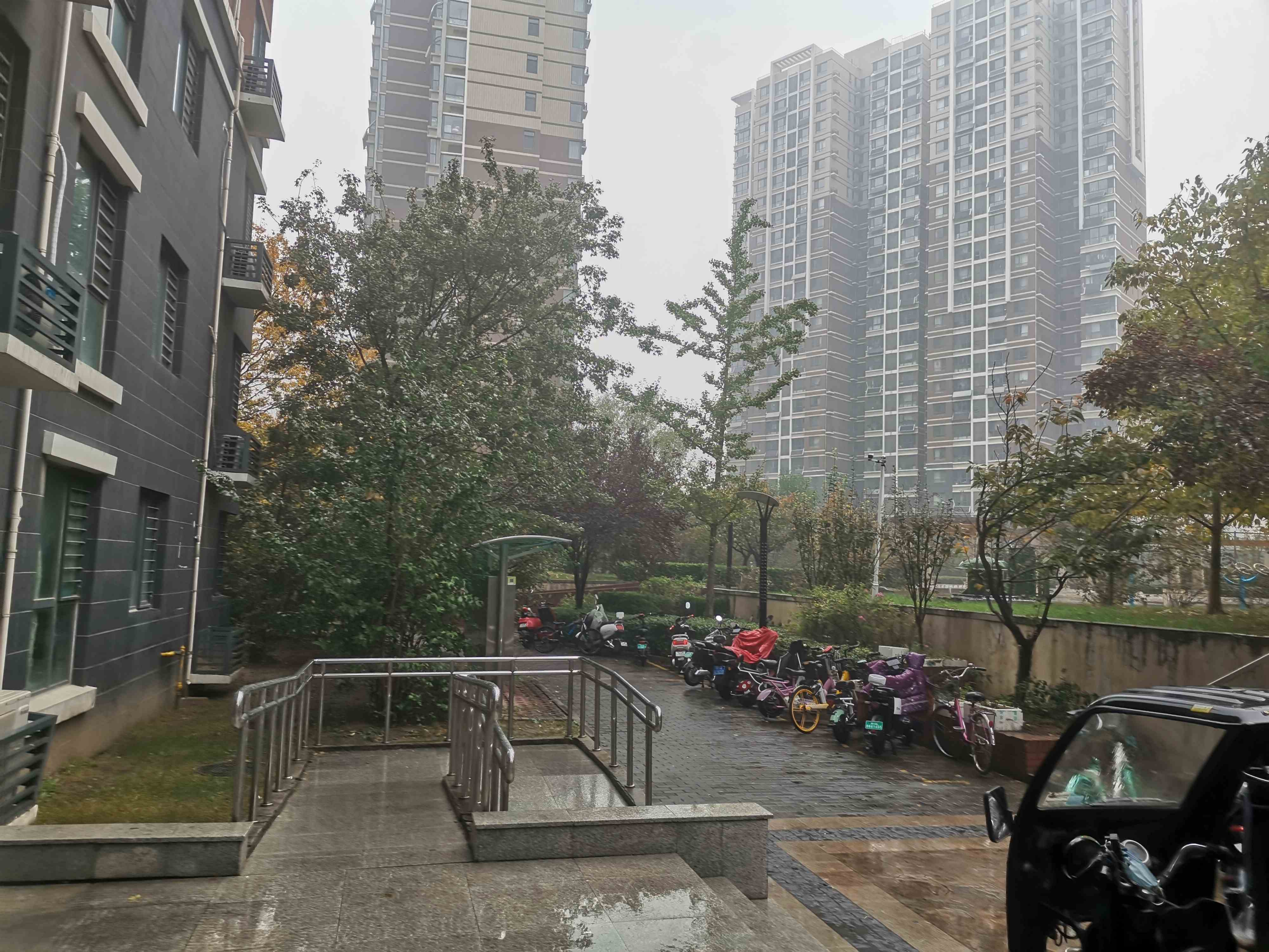}              &
			\includegraphics[width = 0.2\textwidth]{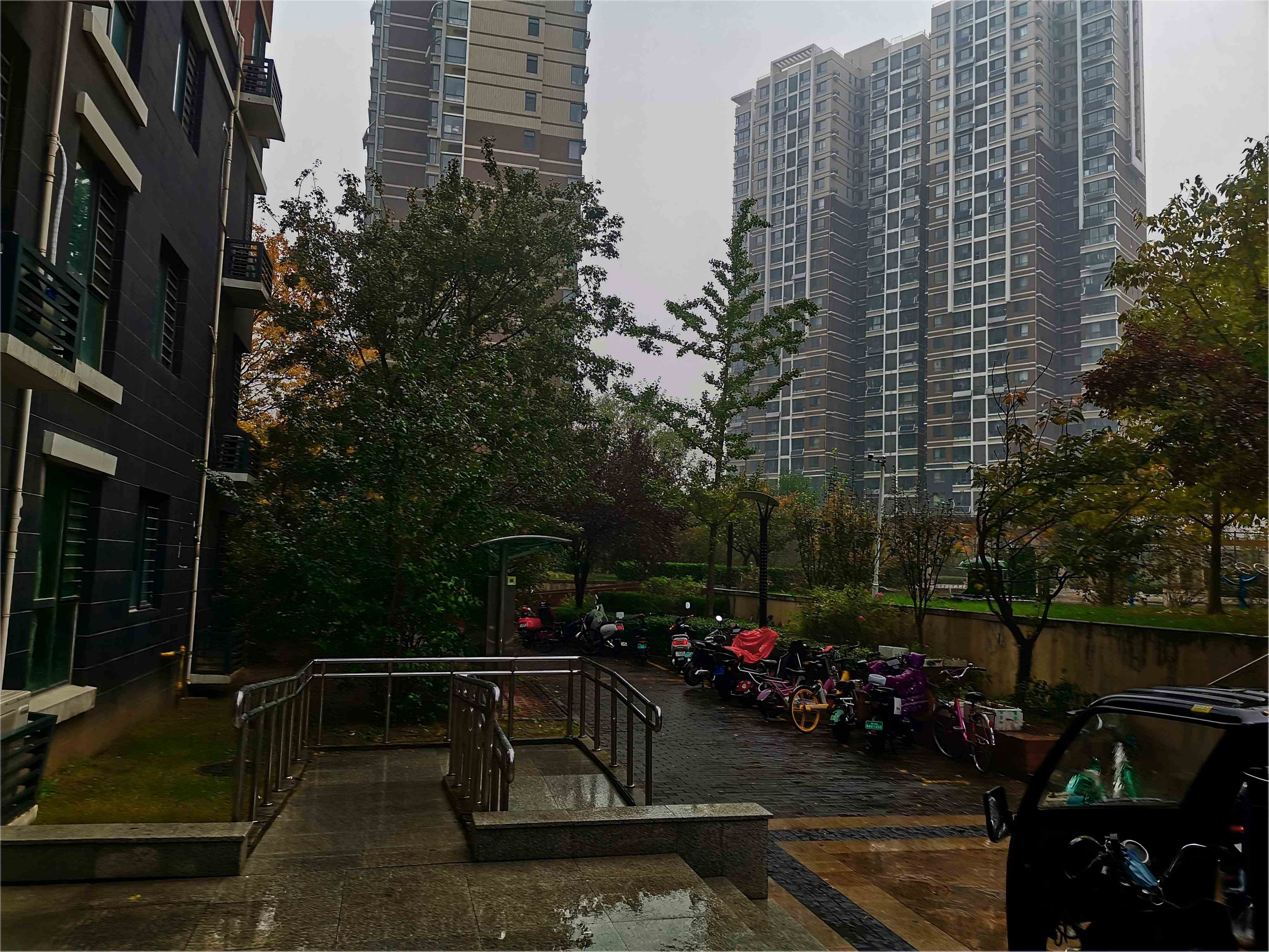}                 &
			\includegraphics[width = 0.2\textwidth]{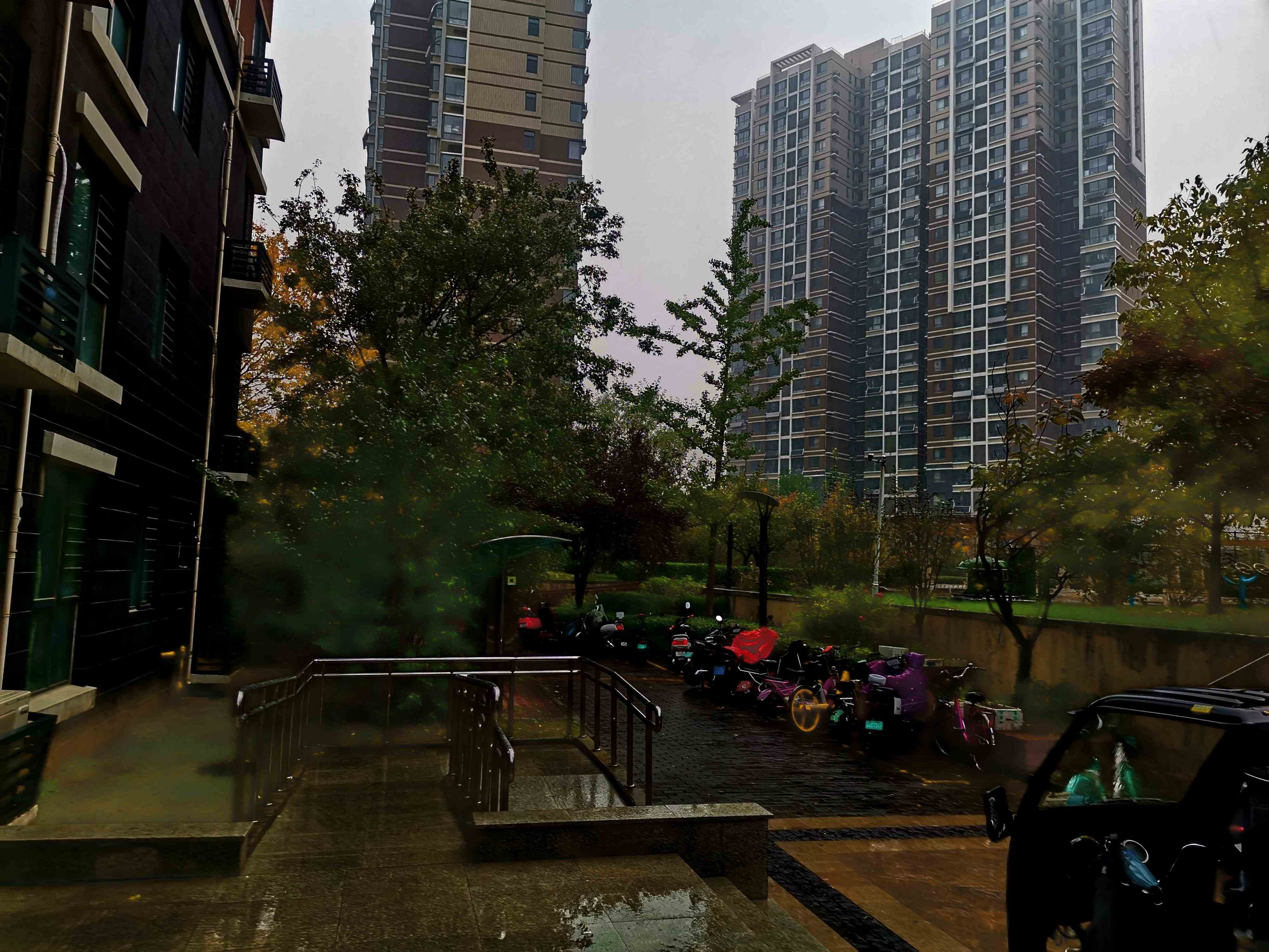}               &
			\includegraphics[width = 0.2\textwidth]{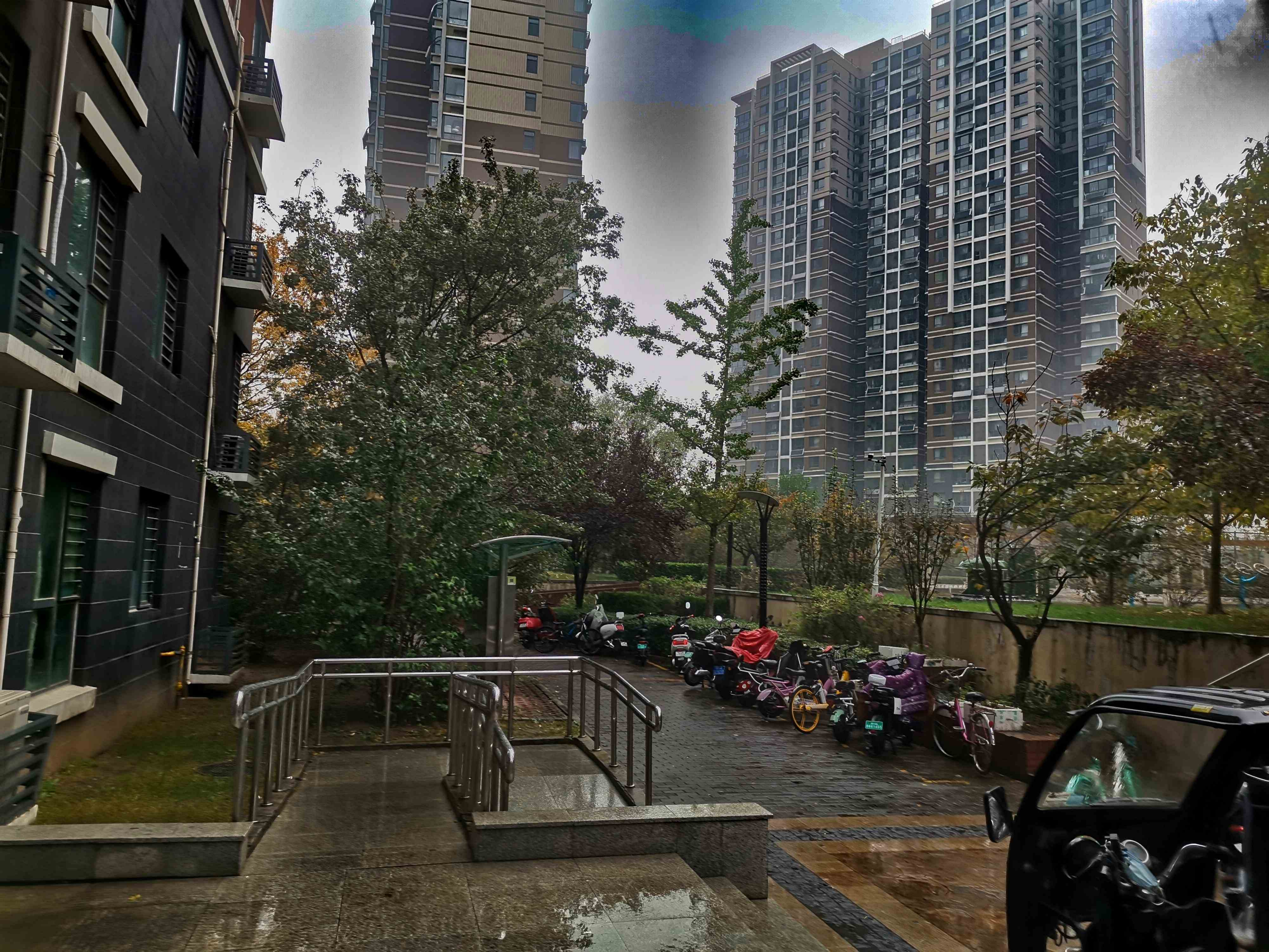}                &
			\includegraphics[width = 0.2\textwidth]{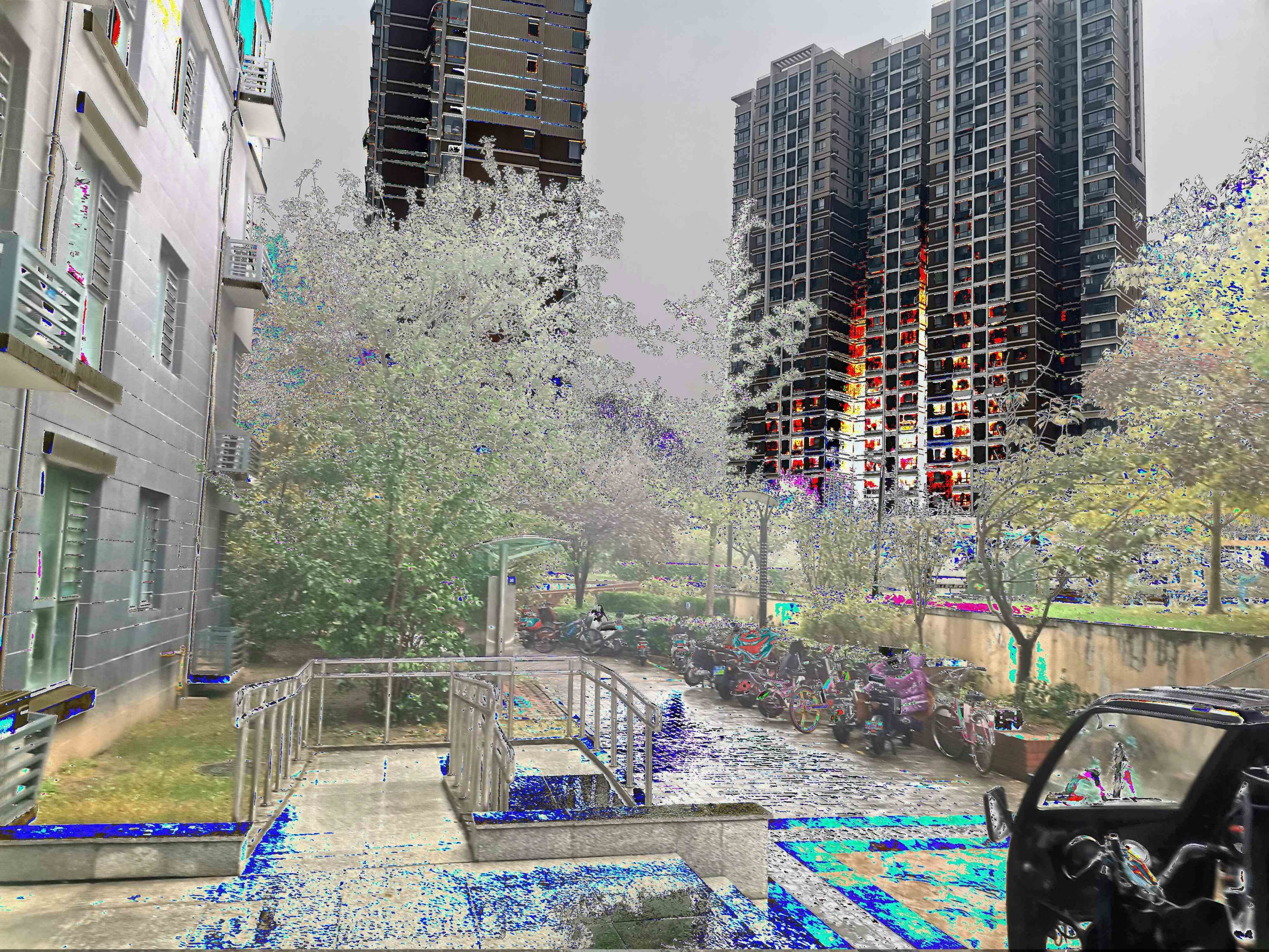}                 \\
			
			\includegraphics[width = 0.2\textwidth]{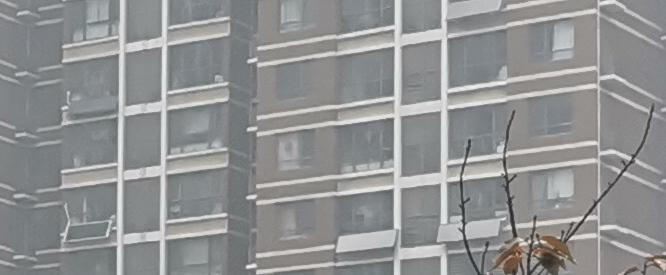}              &
			\includegraphics[width = 0.2\textwidth]{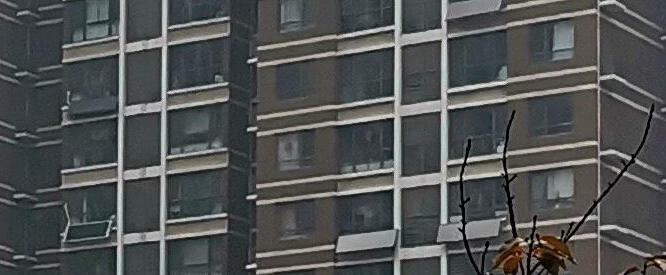}                 &
			\includegraphics[width = 0.2\textwidth]{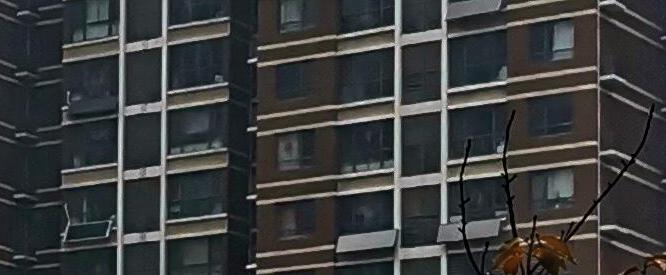}               &
			\includegraphics[width = 0.2\textwidth]{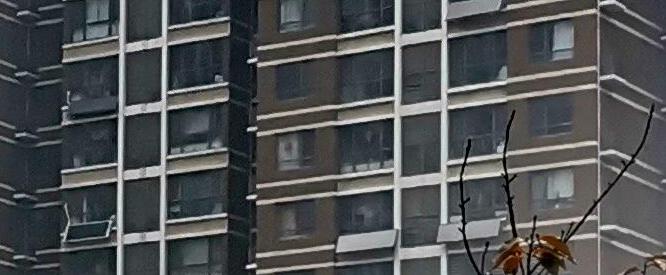}                &
			\includegraphics[width = 0.2\textwidth]{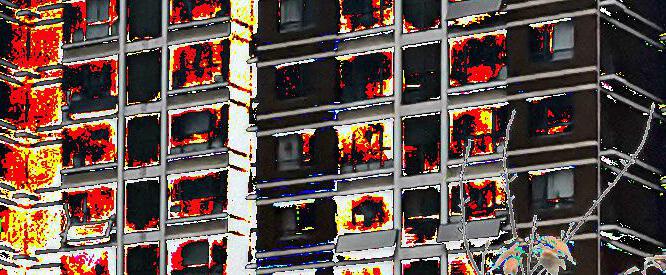}                 \\
			
			Input &
			AOD~\cite{li2017all}    &
			CAP~\cite{zhu2015fast}  &  
			DCP~\cite{he2010single}   & 
			DehazeNet~\cite{cai2016dehazenet}       
			\\ 
			
			\includegraphics[width = 0.2\textwidth]{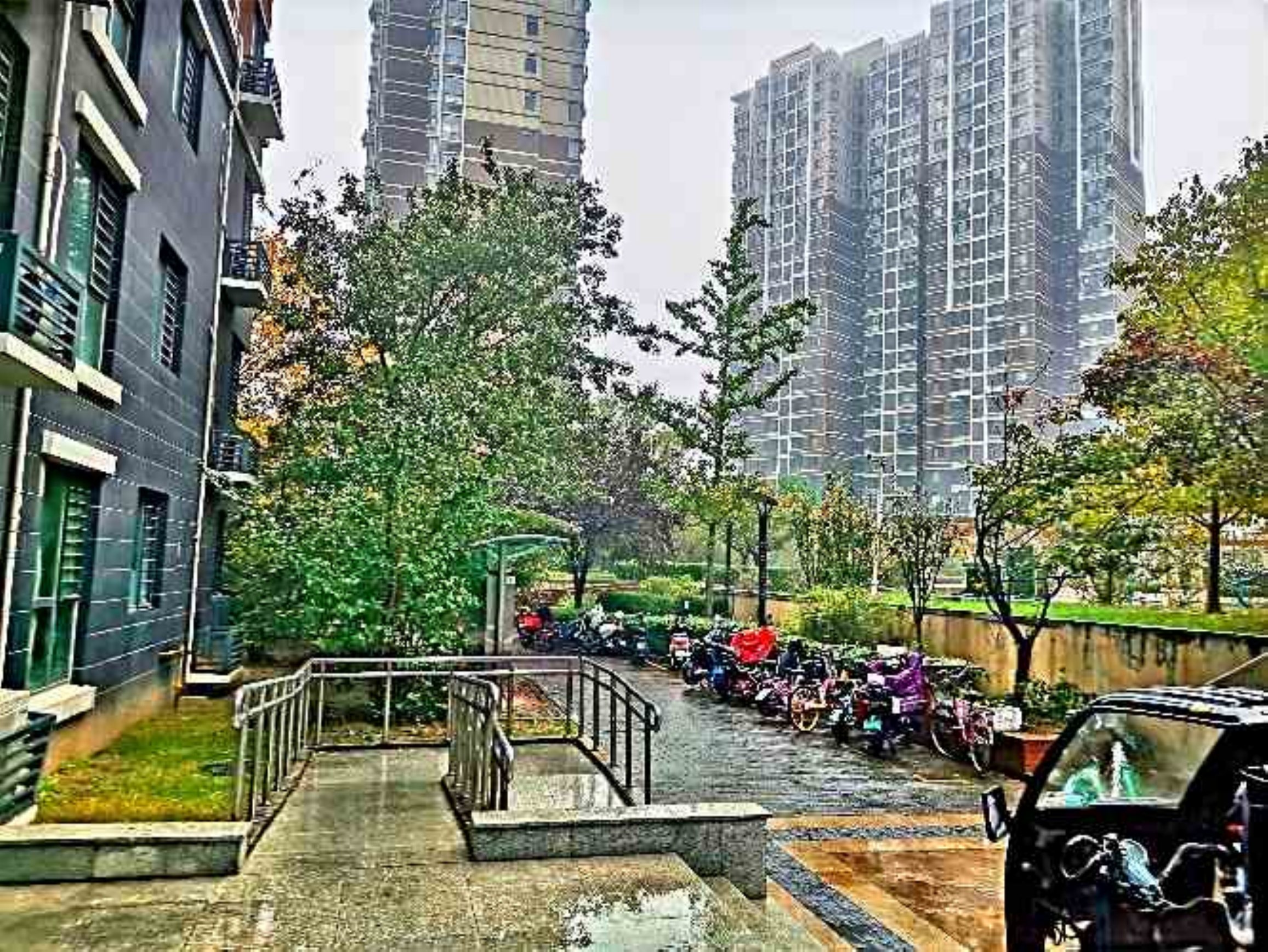}          &
			\includegraphics[width = 0.2\textwidth]{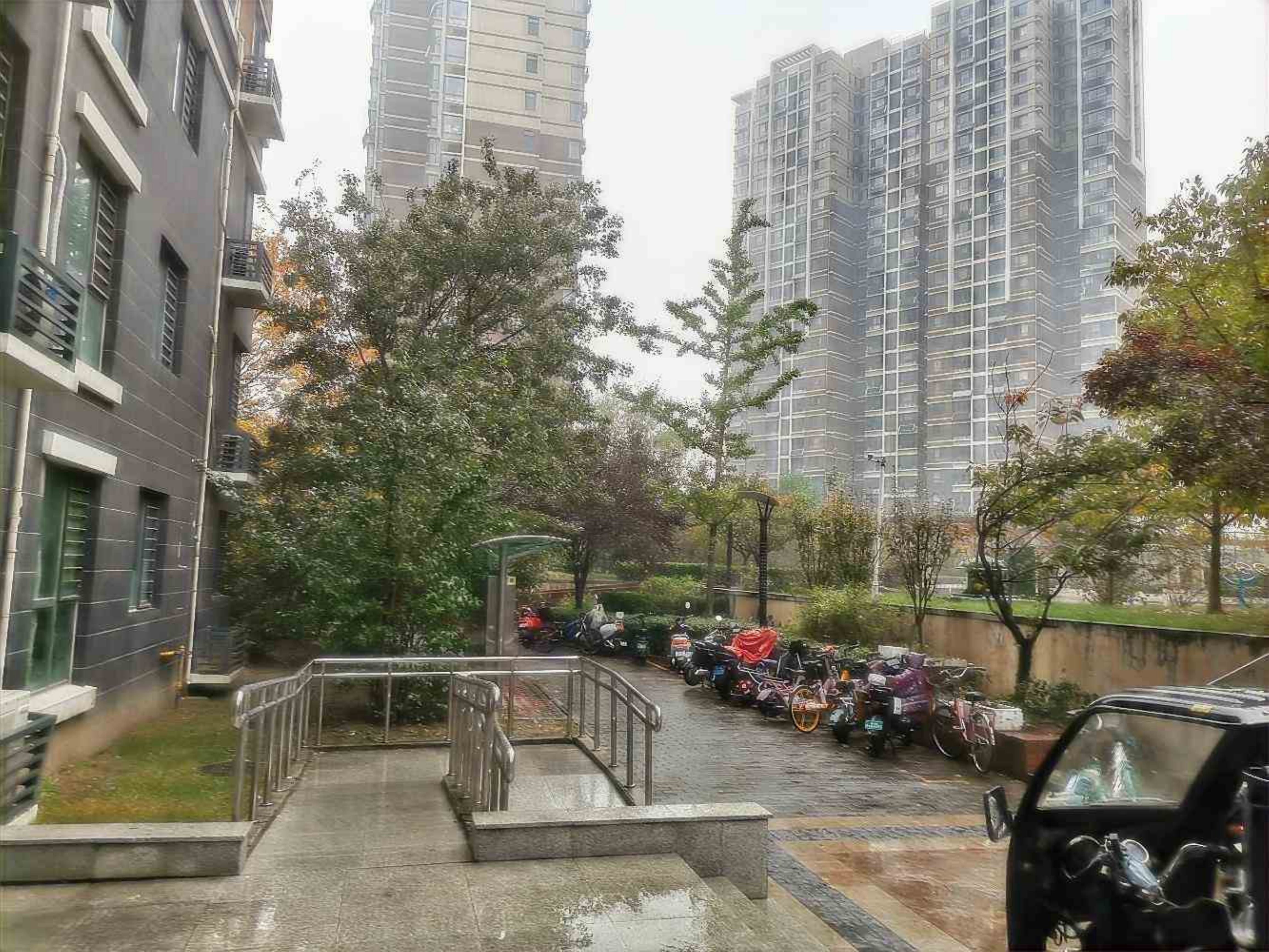}               &
			\includegraphics[width = 0.2\textwidth]{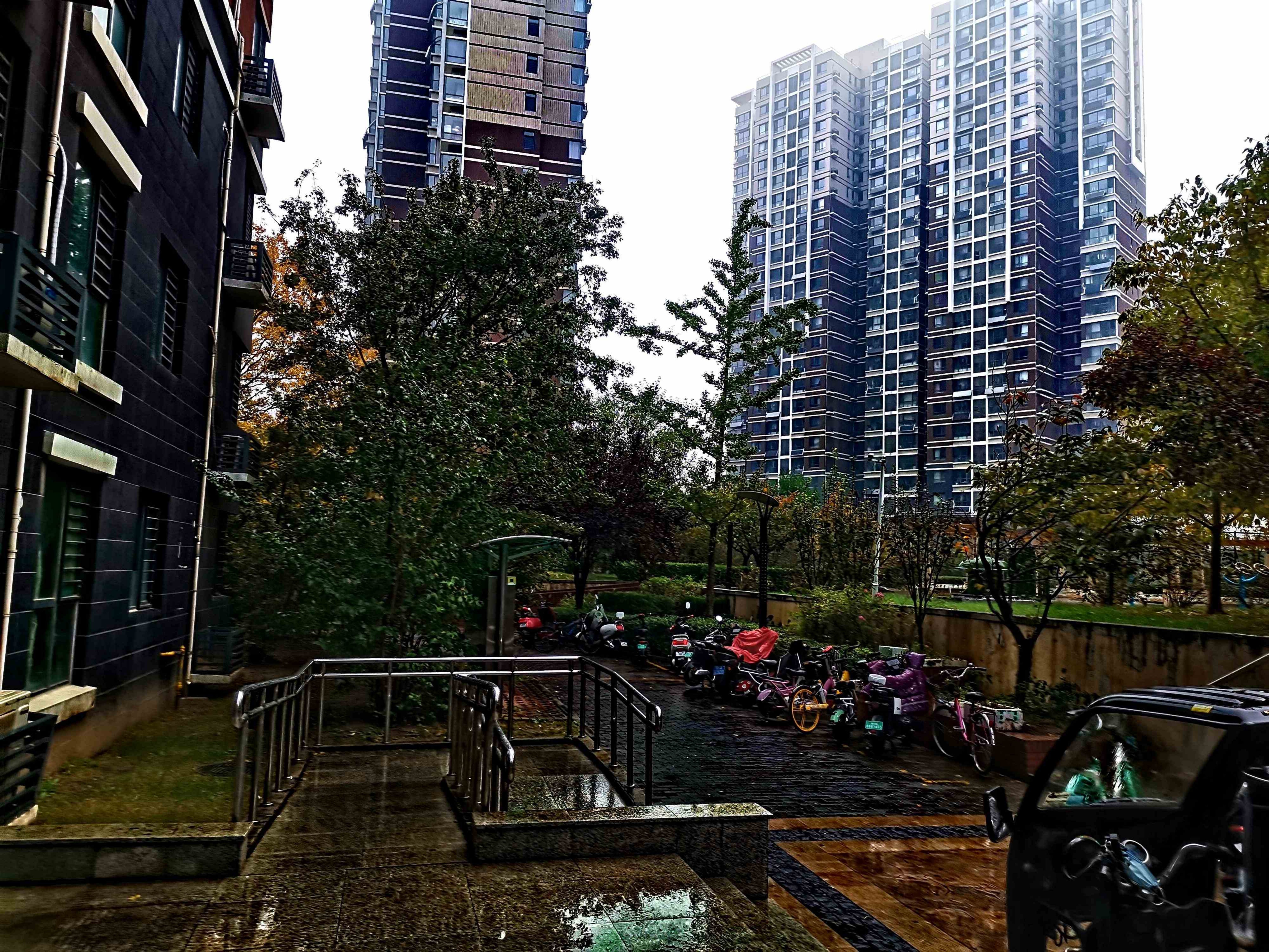}                & 
			\includegraphics[width = 0.2\textwidth]{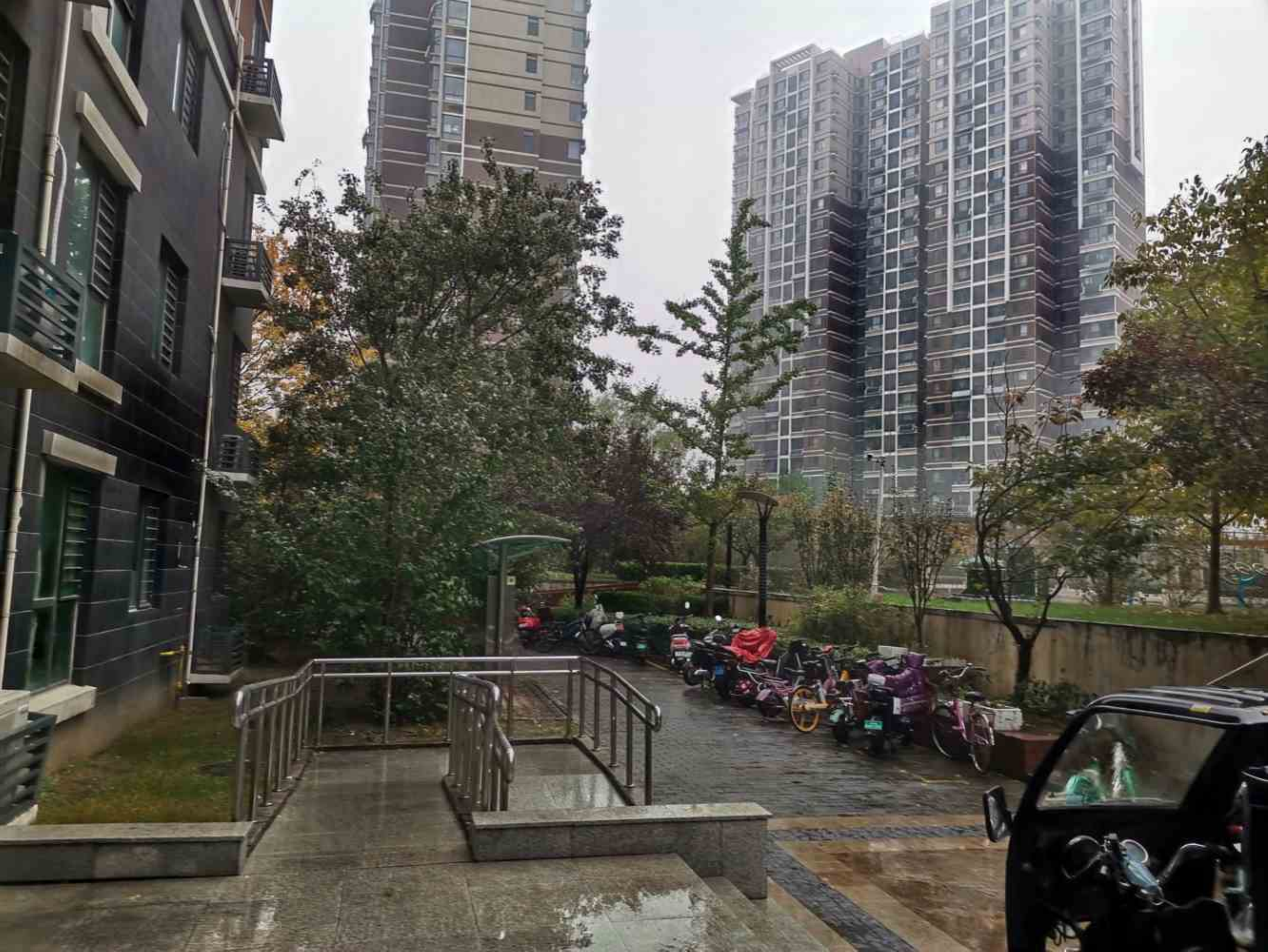}              &
			\includegraphics[width = 0.2\textwidth]{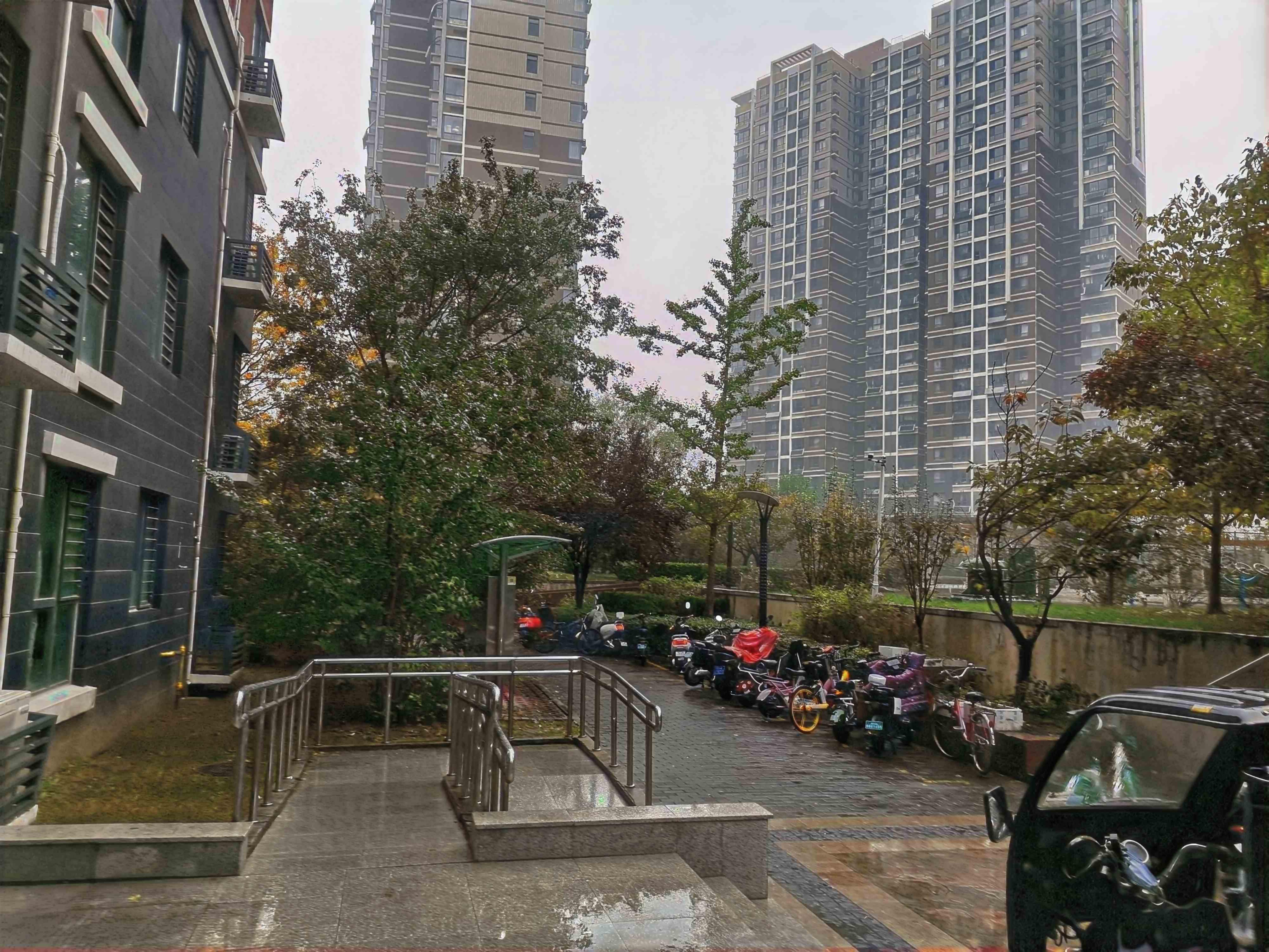}                \\           
			
			\includegraphics[width = 0.2\textwidth]{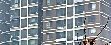}          &
			\includegraphics[width = 0.2\textwidth]{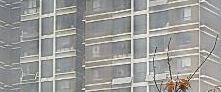}               &
			\includegraphics[width = 0.2\textwidth]{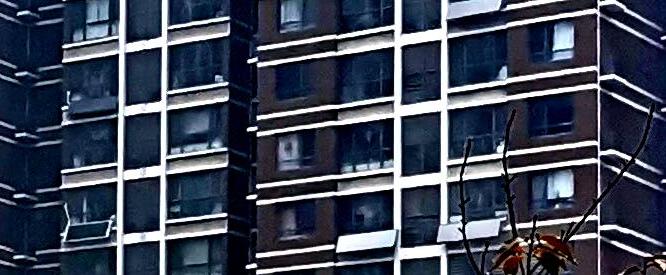}                & 
			\includegraphics[width = 0.2\textwidth]{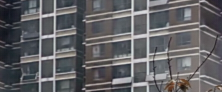}              &
			\includegraphics[width = 0.2\textwidth]{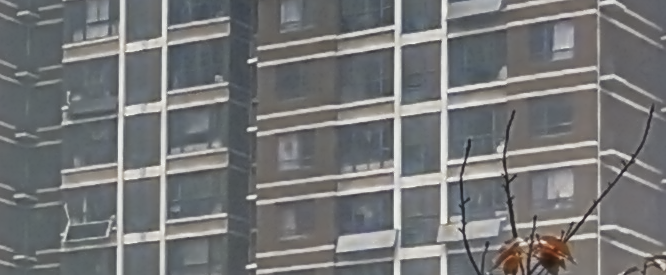}                \\             
			
			PSD~\cite{chen2021psd}     &
			MGLB~\cite{zheng2021ultra} & 
			NLD~\cite{berman2016non}           & 
			GCA~\cite{chen2019gated}         & 
			\textbf{Ours }
			\\
			\includegraphics[width = 0.2\textwidth]{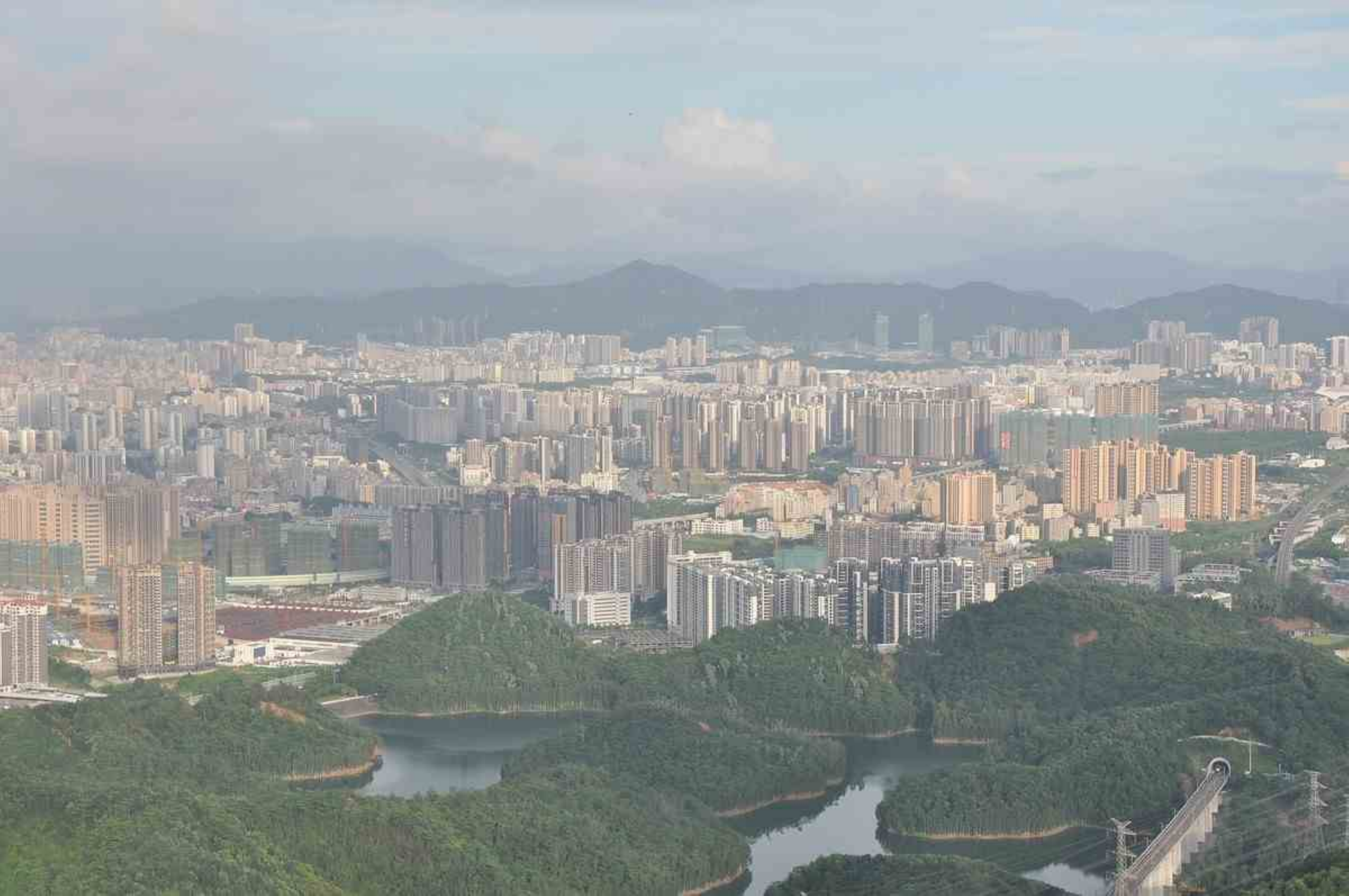}                 &
			\includegraphics[width = 0.2\textwidth]{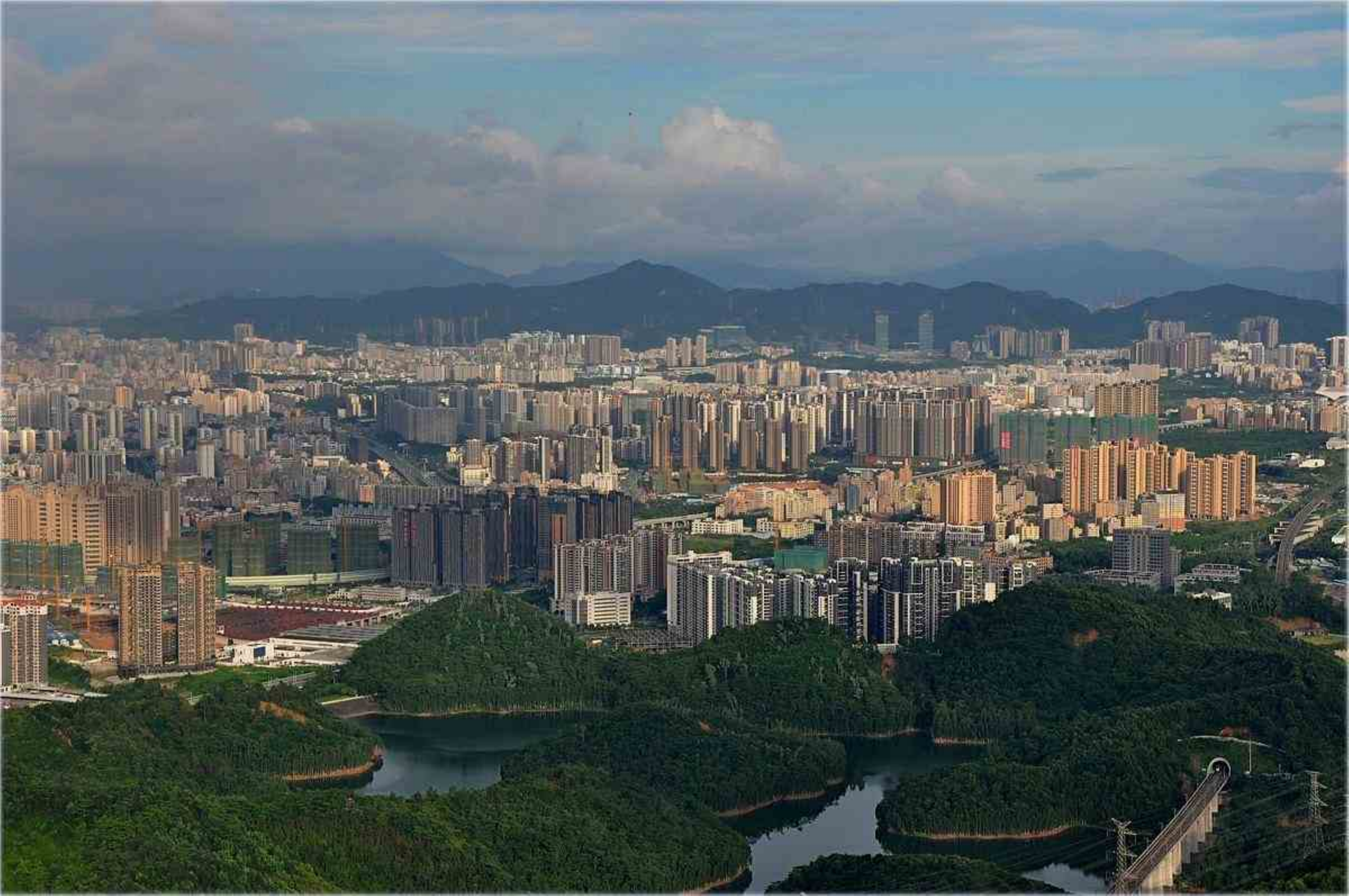}                    &
			\includegraphics[width = 0.2\textwidth]{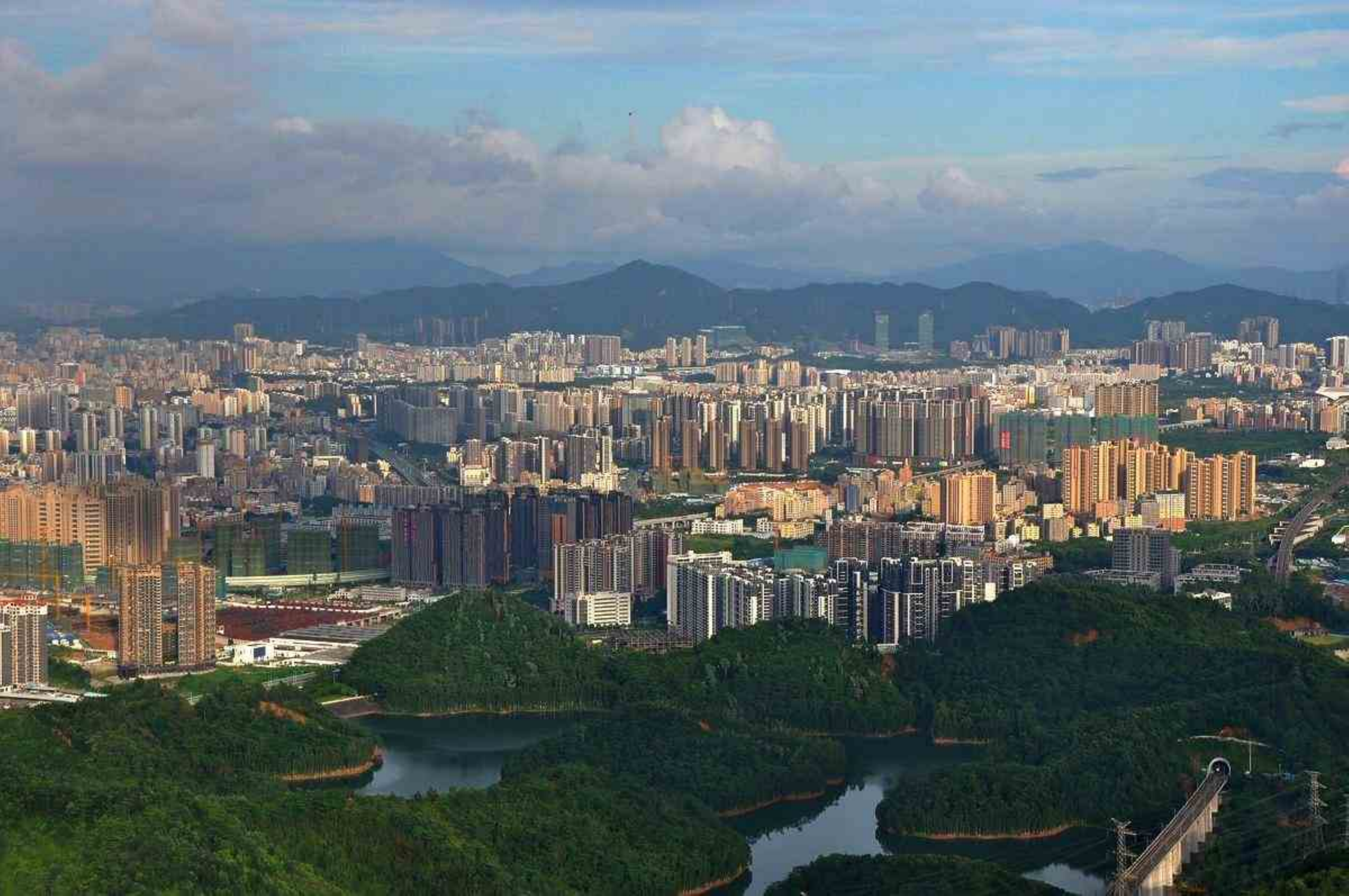}                  &
			\includegraphics[width = 0.2\textwidth]{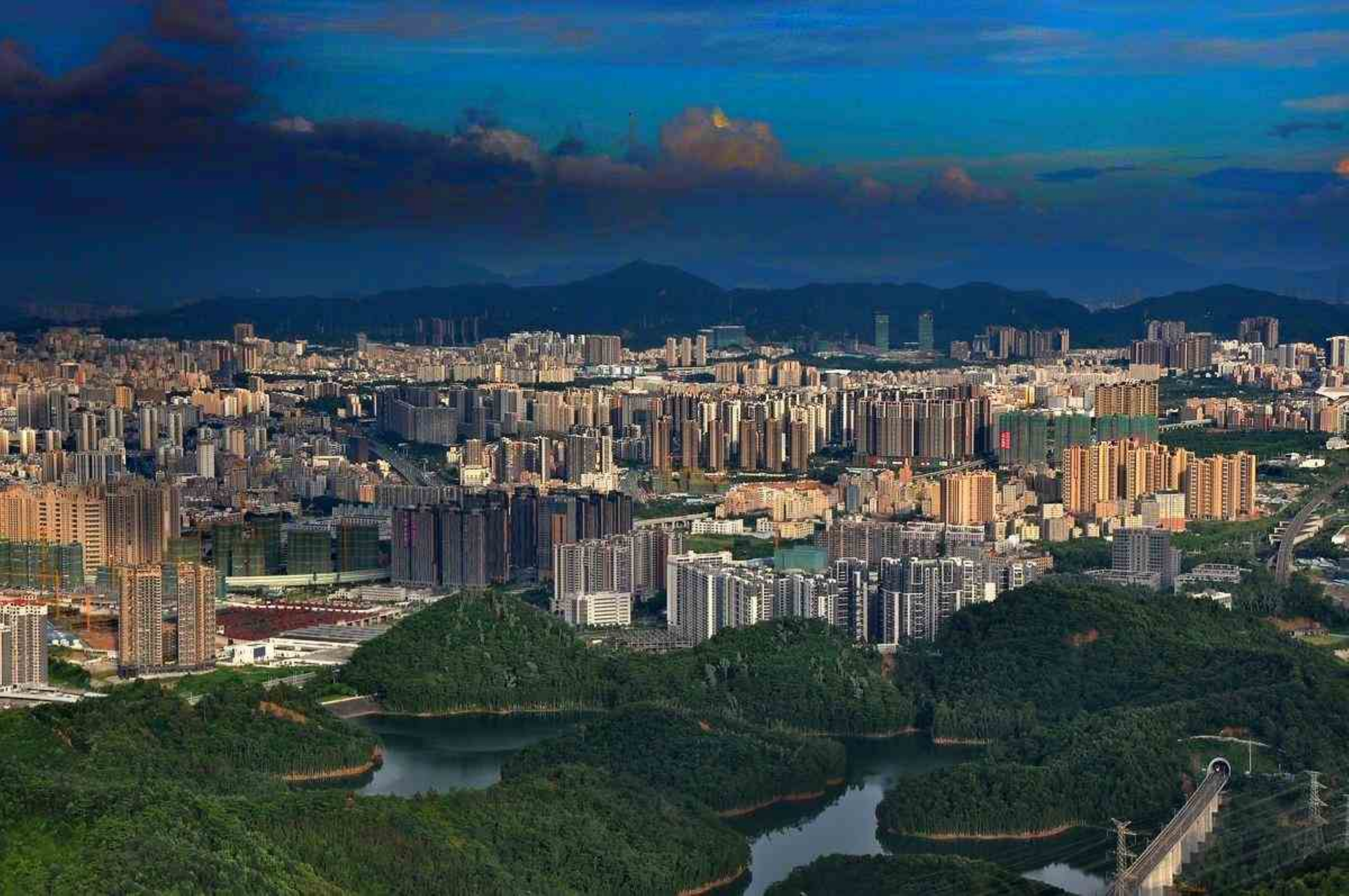}                   &
			\includegraphics[width = 0.2\textwidth]{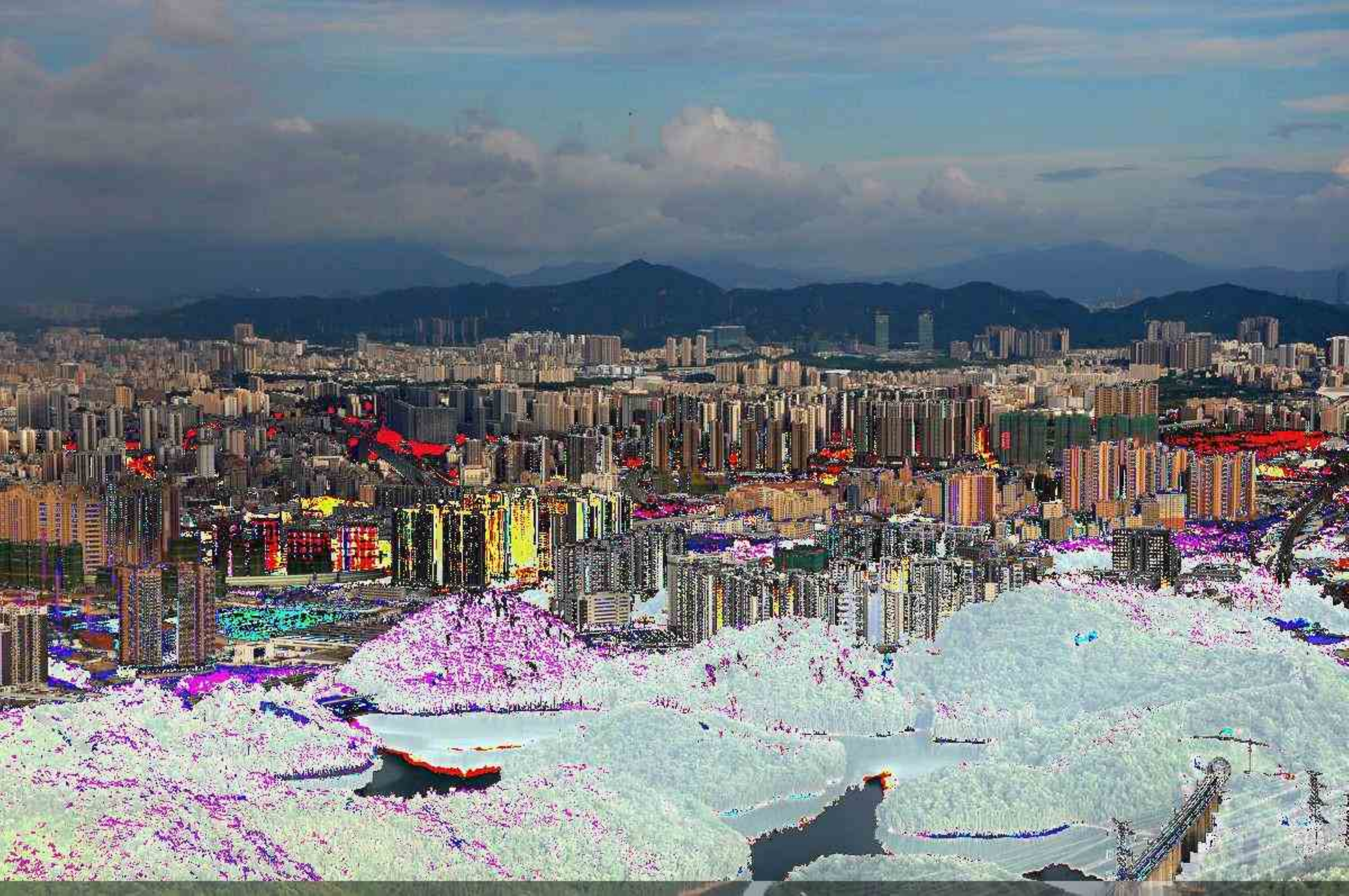}                   \\     
			
			\includegraphics[width = 0.2\textwidth]{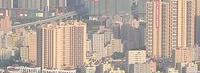}                 &
			\includegraphics[width = 0.2\textwidth]{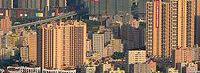}                    &
			\includegraphics[width = 0.2\textwidth]{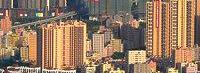}                  &
			\includegraphics[width = 0.2\textwidth]{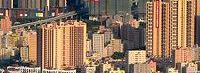}                   &
			\includegraphics[width = 0.2\textwidth]{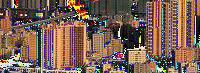}                   \\    
			
			Input &
			AOD~\cite{li2017all}    &
			CAP~\cite{zhu2015fast}  &  
			DCP~\cite{he2010single}   & 
			DehazeNet~\cite{cai2016dehazenet}   
			\\
			
			\includegraphics[width = 0.2\textwidth]{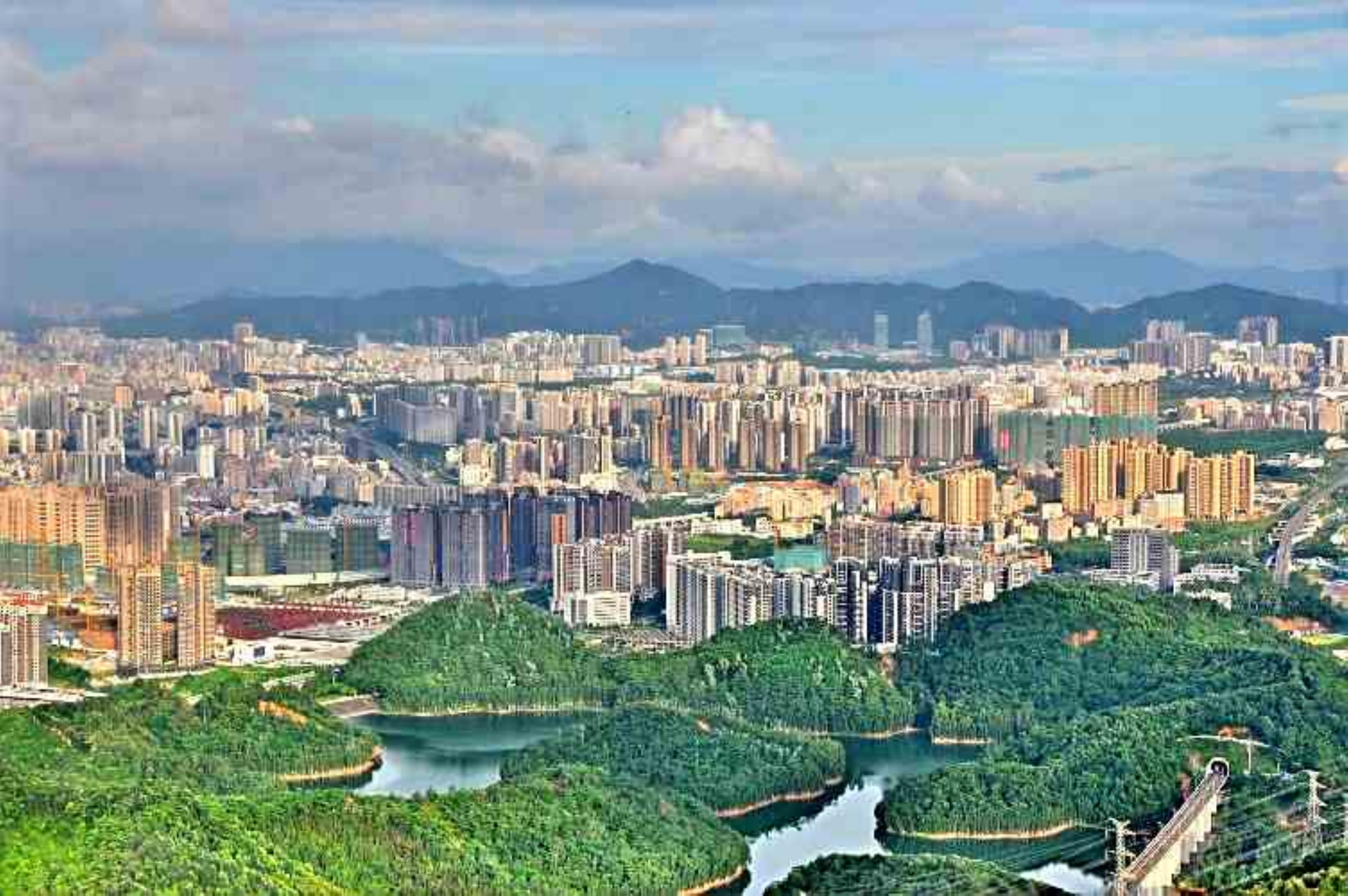}        &
			\includegraphics[width = 0.2\textwidth]{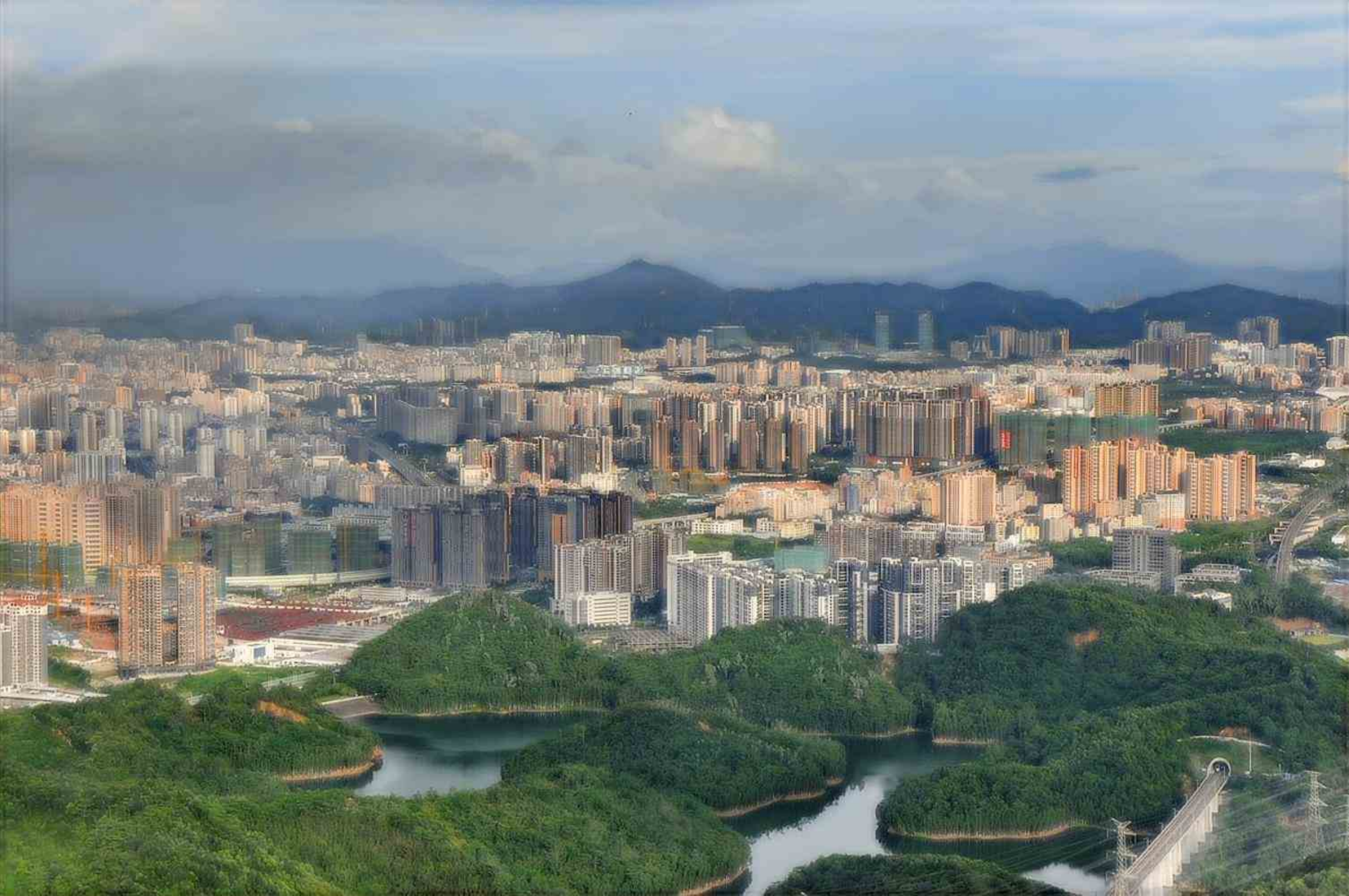}             &
			\includegraphics[width = 0.2\textwidth]{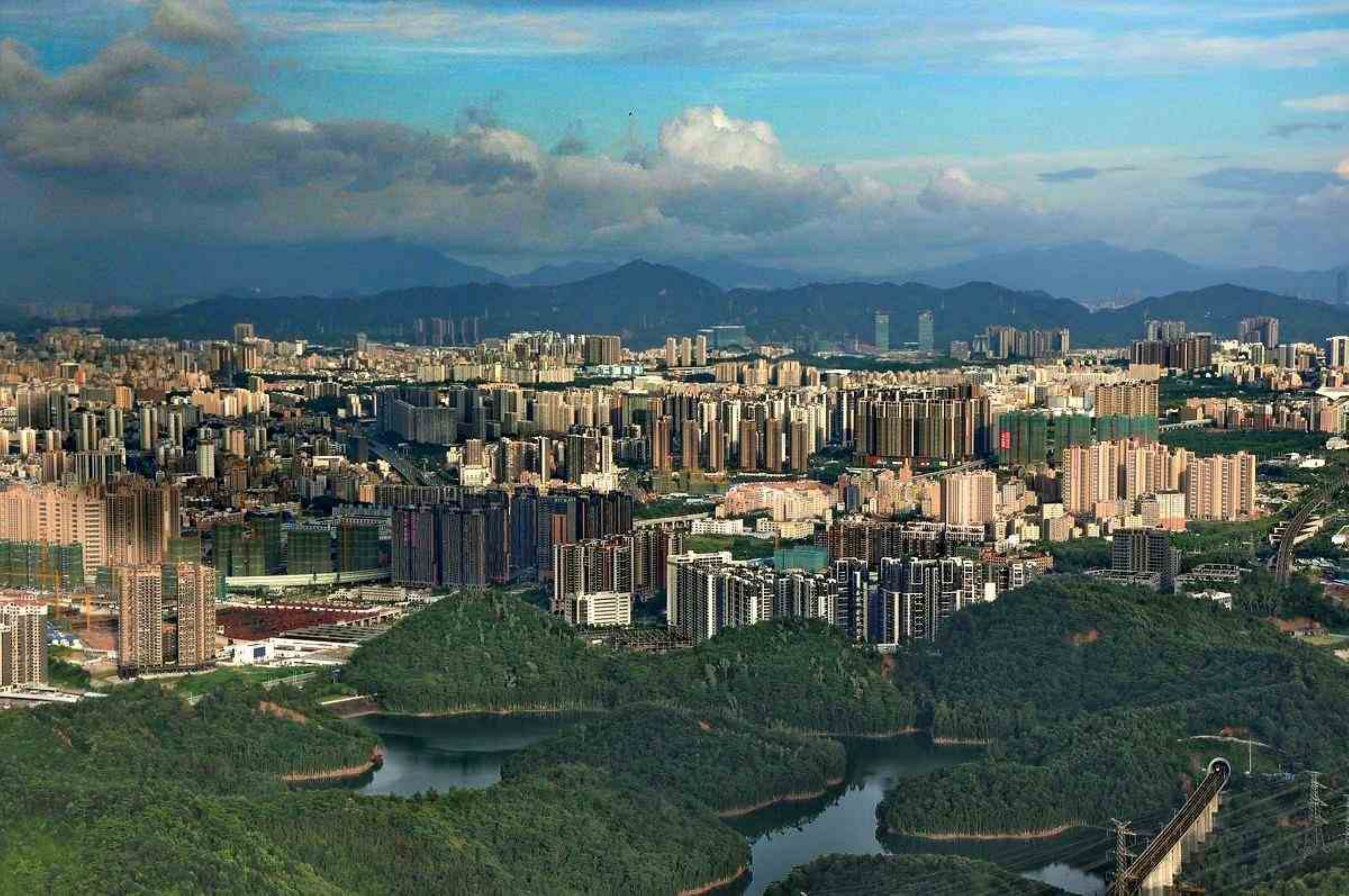}               & 
			\includegraphics[width = 0.2\textwidth]{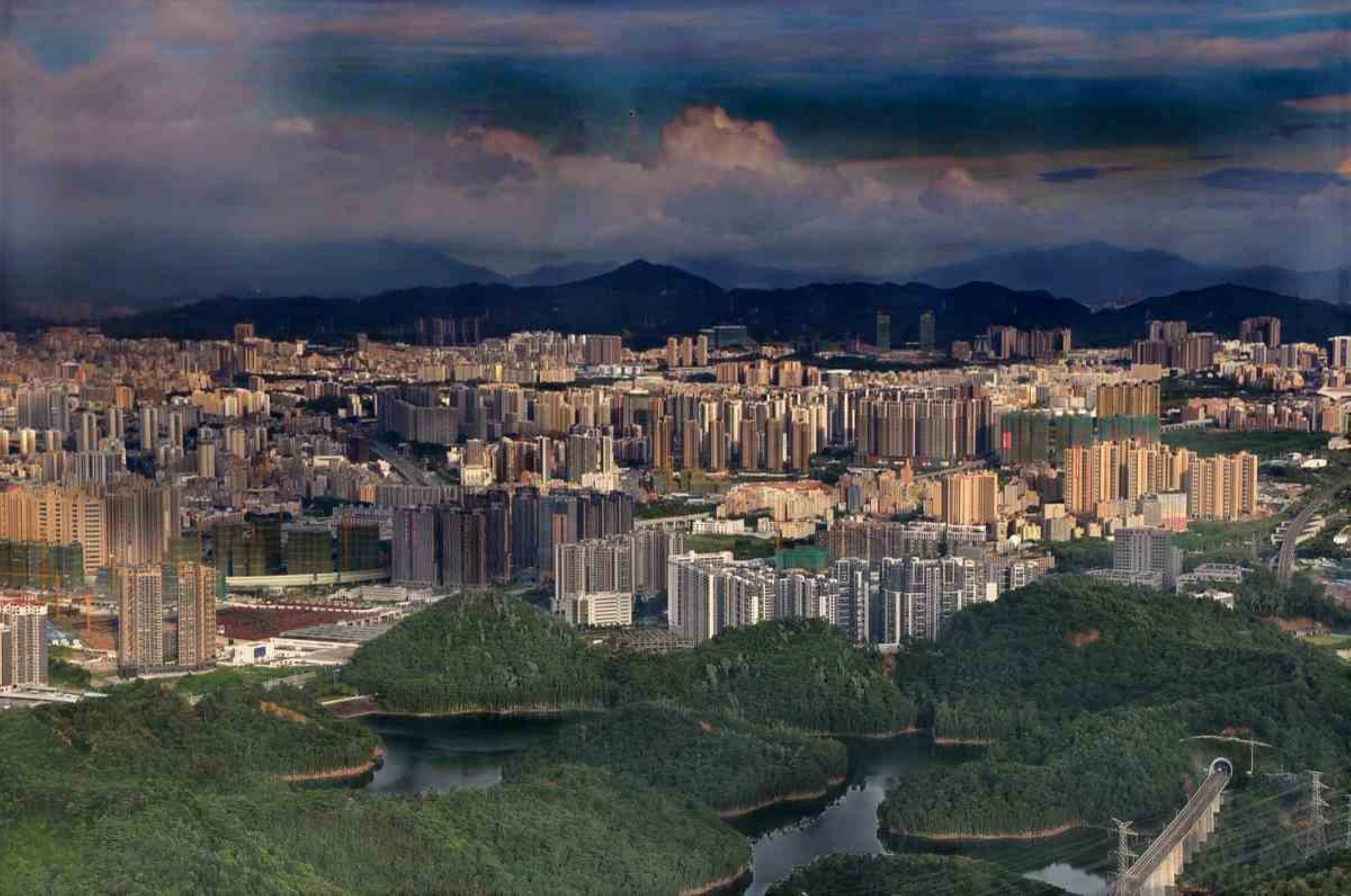}              &
			\includegraphics[width = 0.2\textwidth]{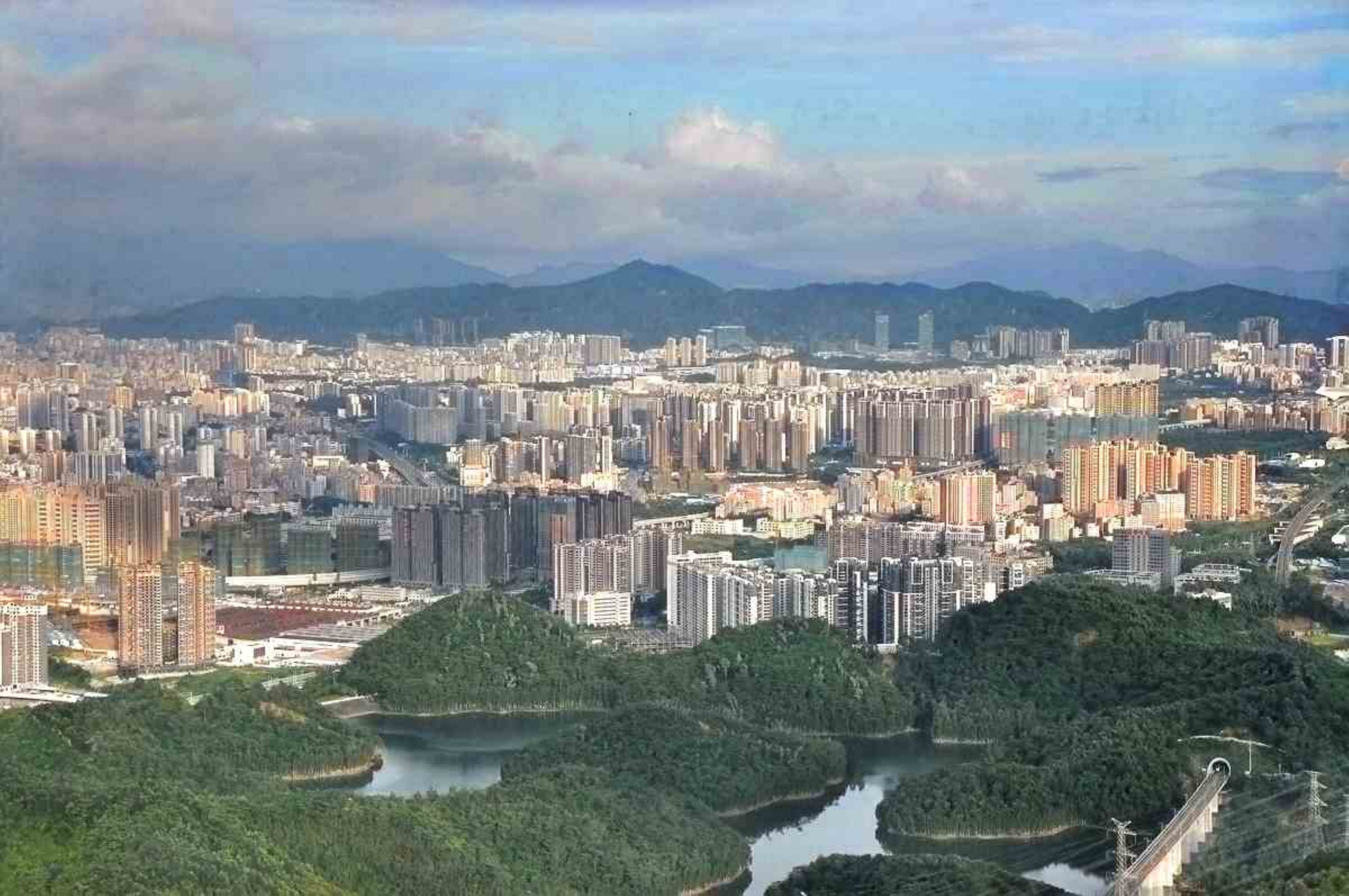}                \\          
			
			\includegraphics[width = 0.2\textwidth]{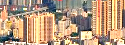}        &
			\includegraphics[width = 0.2\textwidth]{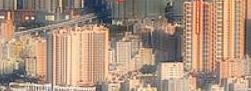}             &
			\includegraphics[width = 0.2\textwidth]{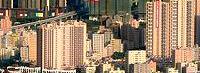}               & 
			\includegraphics[width = 0.2\textwidth]{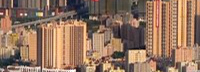}              &
			\includegraphics[width = 0.2\textwidth]{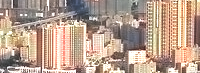}                \\     
			
			PSD~\cite{chen2021psd}     &
			MGBL~\cite{zheng2021ultra} & 
			NLD~\cite{berman2016non}           & 
			GCA~\cite{chen2019gated}         & 
			\textbf{Ours}
			\\
		\end{tabular}
	\end{center}
	\caption{Our method obtains better visual quality and recovers more image details compared with other state-of-the-art methods in the real world.}
	\vspace{-2mm}
	\label{Real_mydata}
\end{figure*}

\subsection{Training Data}
To evaluate our network and others, we choose three public datasets (4KID \cite{zheng2021ultra}, I-HAZE \cite{ancuti2018haze0}, O-HAZE \cite{ancuti2018haze1}). We updated 3000 pairs of hazy and clear images to the 4KID dataset to improve the generalization ability of the model. The newly added 3000 pairs of images were taken by several different mobile phones at 4K resolution and synthesized based on atmospheric scattering model \cite{he2010single}. Additionally, we randomly sampled atmospheric light conditions $A\epsilon[0.8,1]$ and scattering  coefficient $\beta\epsilon[0.4,2.0]$ to generate the corresponding hazy images. We further used the translation module in \cite{shao2020domain} to translate the synthetic hazy images from  synthetic domain to real domain, which can improve the generalization of deep  models in real cases. Finally, we used 10400 images to be training set and randomly selected 200 images from the remaining images as test data.

\subsection{Implementation Details}
The proposed model is implemented in PyTorch 1.7 and Adam optimizer  \cite{kingma2014adam} is used to train the network. We use original resolution images to train the network and the batch size is 1. The initial learning rate is 0.0002. We use charbonnier loss \cite{lai2017deep}. This loss can be restated as follows:

\begin{eqnarray}
\text{CharbonnierLoss} = \sqrt{x^2+\varepsilon^2},
\end{eqnarray}
where $x$ is the difference between the predicted image and the standard image, $\varepsilon$ is a constant.
Instead of using the $L_2$ loss function, it can handle outliers and improve the performance.
In addition, we trained 80,000 steps in total.
Since AOD-NET \cite{li2017all} cannot directly remove haze from 4K images with RTX 3090 GPU and E5-2678 CPU, we use the method proposed by Zheng et al. \cite{zheng2021ultra} to run the program to compare with others. \textbf{Note that all methods were fine-tuned on the 4KID dataset.}

\subsection{Visual Comparisons}
We compared our proposed methods with others on three datasets: 4KID, O-HAZE, and I-HAZE. Figure \ref{4KID} shows two images from the 4KID dataset. Figure \ref{O-HAZE} shows one example from the O-HAZE dataset and Figure \ref{IHAZE} shows one example from the I-HAZE dataset. It is easy to find that the traditional prior-based dehazing models have poor dehazing ability and are prone to color distortion caused by over-exposure. And they need lots of reasoning time. However, the recent deep learning-based dehazing model \cite{zheng2021ultra,li2017all,chen2021psd} have the problem that the dehazing images are fuzzy and texture details are not obvious. In Figure \ref{O-HAZE}, the dehazing images generated by our algorithm not only have small color distortion, but also retain a large number of texture details. Our results have a  realistic effect, and they are close to real dehazing images. The specific experimental data compared with other methods are reported in Table \ref{dataset}.

In addition to the comparison of the dehazing test on the synthetic dataset, we also selected two challenging real hazy images to compare our methods with others. As shown in Figure \ref{Real_mydata}, serious color distortion occurs in DCP and non-local. AOD-NET, CAP, and PSD cause residual haze. The visual perception of MGBL is blurred. At the same time, our algorithm can eliminate haze as much as possible under the condition of minimizing color distortion.

In order to prove the superiority of our method, we also select a group of low-resolution hazy images to remove haze. Please refer to the images in the supplementary material.

\begin{table}[h] \footnotesize
	\begin{center}
		\begin{tabular}{cccc}
			\hline
			& w/o Tucker & single U-Net& Ours            \\ \hline
			PSNR & 27.20     &   26.50    & \textbf{27.79}  \\
			SSIM & 0.89    &    0.90   & \textbf{0.92} \\ \hline
		\end{tabular}%
	\end{center}
	\vspace{-3mm}
	\caption{Our method has achieved the best results on both evaluation indices.}
	\label{tab2}
	\vspace{-5mm}
\end{table}

	\vspace{-1mm}
\subsection{Ablation Study}
To demonstrate the effectiveness of each module introduced in the proposed network, we perform an ablation study on the 4KID dataset involving the following three experiments:

\noindent
\textbf{The effectiveness of the Tucker reconstruction.} We removed the Tucker reconstructions that acted on the pyramid branch network, retrained the model, and tested it.

\noindent
\textbf{The effectiveness of the learning process of $K$.} We concat the low-frequency components before and after processing and the unprocessed high-frequency components directly and use it for the attention sharing tensor $K$ to fit the differential operation in Taylor polynomials. 

\noindent
\textbf{The effectiveness of the number of terms in Taylor's theorem.}  We use different polynomials to fit Taylor's expansion. 

\noindent
Results:
Firstly, the network with Tucker reconstruction regularization term fits the mapping relationship better in the image dehazing (see Table \ref{tab2}). Additionally, after we remove the low-rank network, 
the PSNR only dropped from 27.79dB to PSNR 26.50dB. But the processing time of a single image can reach 7.9ms. It is another way of saying that we have a tradeoff between speed and time.
Secondly, as shown in Table \ref{tab3}, with the increase of Taylor term, the targeted value decreases because the resolution of the first term is low, and part of the effective information is lost, so the first term of Taylor's expansion cannot be fitted well. In the same way, we can find that fewer Taylor terms can also achieve a similar effect, but because of the increase in the minimum resolution of the pyramid, it takes more than twice the time of terms=4, which is obviously not a wise choice.

\begin{table}[h] \footnotesize
	\begin{center}
		\begin{tabular}{ccccc}
			\hline
				 &terms=3	& terms=4 & terms=5 & terms=6            \\ \hline
			PSNR & 27.2	& 27.79    &  26.93    & 25.58			\\
			SSIM & 0.91	&0.92    & 0.92    &  0.91			\\ \hline
		\end{tabular}%
	\end{center}
	\vspace{-6mm}
	\caption{Experimental results of fitting mapping function with different numbers of Taylor terms.}
	\label{tab3}
	\vspace{-8mm}
\end{table}
\subsection{Discussion for UHD image}
 Most UHD processing methods map high-resolution images to low-resolution images for calculation and then restore the image to high-resolution, which can reduce the time but the scale is still single. Additionally, these methods are easy to lose texture detail. For instance, Zheng et al. \cite{zheng2021ultra} fused low-resolution images processed to RGB channels for restoration. It can obtain a better color effect, but the picture appears obvious blur. 
To address the above problems, our method uses a multi-scale pyramid to fuse Taylor items to compensate the texture. It reduces computation while maintaining realistic texture details. Texture details are still visible when the image is enlarged.

\noindent
\textbf{Run time.} We evaluate all the deep models on the same machine with an Intel(R) Xeon(R) CPU and an NVIDIA Titan RTX GPU. The run time is only the processing time of the GPU without considering the I/O operations. The average run times for the 4KID, I-HAZE, and O-HAZE datasets are shown in Table~\ref{dataset}.  The conventional methods solve complex energy functions which inevitably increases the computational cost. Although some lightweight networks can run the 4K images in real-time, their performance is inferior to ours. In addition, while some large networks achieve better performance, they are cannot run a  4K image in real-time.

\noindent
\textbf{Model performance.} 
We test 
through two quantitative indicators PSNR and SSIM, and subjective feelings of images. Conventional methods rely on physical priori and can only deal with thin haze areas, and there will be haze residue in dealing with haze. The generalization ability of deep learning dehazing networks needs to be improved, and some regions are prone to color distortion. In addition, they require a lot of convolution and heavy computation when processing high-resolution images, and the results are not as good as ours. 

\vspace{-3mm}

\section{Conclusion}
\vspace{-1mm}
In this paper, we propose to use Taylor's infinite approximation to fit the mapping function in the process of dehazing and use the Laplace pyramid to fuse the Taylor terms. We also propose a regularization term based on Tucker reconstruction, which acts on low rank U-Net of the pyramid  to further reduce the noise generated during the up-down sampling process. 
Our network performs well on three public datasets (4KID, O-HAZE, IHAZE) and can process 4K images in real-time on a single GPU (80FPS).

\clearpage

{\small
	\bibliographystyle{ieee_fullname}
	\bibliography{egbib}
	
}
\end{document}